\documentclass{amsart}

\usepackage{amsmath}                        
\usepackage{amssymb}                        
\usepackage{dsfont} 	                    
\usepackage{bm}                             
\usepackage{mathtools}                      
\usepackage[hypertexnames=false]{hyperref}  
\usepackage[scientific-notation=true]{siunitx}
\usepackage{blkarray}
\usepackage{enumerate}
\usepackage{centernot}
\usepackage{paralist}
\usepackage{stackrel}
\usepackage[table,xcdraw]{xcolor}
\usepackage{booktabs}
\usepackage[colorinlistoftodos,prependcaption,textsize=tiny]{todonotes} 
\usepackage{pagenote}
\usepackage{setspace}
\usepackage{longtable}
\usepackage{rotating,graphicx}
\usepackage[english]{babel}

\makepagenote

\renewcommand*{\pagenotesubhead}[2]{}
\let\footnote\pagenote

\usepackage[numbers]{natbib}
\usepackage{bibunits}

\usepackage{algorithm}
\usepackage{algpseudocode}

\usepackage{listings}
\usepackage{color}

\usepackage{amsthm}
\theoremstyle{plain}
\theoremstyle{definition}
\newtheorem{example}{Example}

\theoremstyle{remark}

\newcommand*{\QEDE}{\hfill\ensuremath{\blacksquare}}
\def\ci{\perp\!\!\!\perp}

\DeclareMathOperator*{\argmin}{arg\,min}    
\DeclareMathOperator{\tr}{tr}               

\makeatletter
\newcommand{\addresseshere}{%
  \enddoc@text\let\enddoc@text\relax
}

\definecolor{lightgrey}{rgb}{0.9,0.9,0.9}
\definecolor{darkgreen}{rgb}{0,0.6,0}
\calclayout

\begin{document}
\defaultbibliographystyle{plainnat}

\title[Stable Prediction with Radiomics Data]
{Stable Prediction with Radiomics Data}

\author[C.F.W.\ Peeters \emph{et al}.]{Carel F.W.\ Peeters*}
\thanks{*Corresponding author.}
\address[Carel F.W.\ Peeters]{
Dept.\ of Epidemiology \& Biostatistics \\
Amsterdam Public Health research institute \\
Amsterdam University medical centers, location VUmc \\
Amsterdam\\
The Netherlands}
\email{cf.peeters@vumc.nl}

\author[]{Caroline \"{U}belh\"{o}r}
\address[Caroline \"{U}belh\"{o}r]{
Dept.\ of Epidemiology \& Biostatistics \\
Amsterdam Public Health research institute \\
Amsterdam University medical centers, location VUmc \\
Amsterdam\\
The Netherlands}

\author[]{\\Steven W.\ Mes}
\address[Steven W.\ Mes]{
Dept.\ Otolaryngology/Head \& Neck Surgery \\
Amsterdam University medical centers, location VUmc \\
VUmc Cancer Center \\
Amsterdam\\
The Netherlands}

\author[]{Roland Martens}
\address[Roland Martens]{
Dept.\ of Radiology \& Nuclear Medicine \\
Amsterdam University medical centers, location VUmc \\
VUmc Cancer Center \\
Amsterdam\\
The Netherlands}

\author[]{Thomas Koopman}
\address[Thomas Koopman]{
Dept.\ of Radiology \& Nuclear Medicine \\
Amsterdam University medical centers, location VUmc \\
VUmc Cancer Center \\
Amsterdam\\
The Netherlands}

\author[]{Pim de Graaf}
\address[Pim de Graaf]{
Dept.\ of Radiology \& Nuclear Medicine \\
Amsterdam University medical centers, location VUmc \\
VUmc Cancer Center \\
Amsterdam\\
The Netherlands}

\author[]{Floris H.P.\ van Velden}
\address[Floris H.P.\ van Velden]{
Dept.\ of Radiology \\
Leiden University medical center\\
Leiden\\
The Netherlands}

\author[]{Ronald Boellaard}
\address[Ronald Boellaard]{
Dept.\ of Radiology \& Nuclear Medicine \\
Amsterdam University medical centers, location VUmc \\
VUmc Cancer Center \\
Amsterdam\\
The Netherlands}

\author[]{Jonas A.\ Castelijns}
\address[Jonas A.\ Castelijns]{
Dept.\ of Radiology \& Nuclear Medicine \\
Amsterdam University medical centers, location VUmc \\
VUmc Cancer Center \\
Amsterdam\\
The Netherlands}

\author[]{\\Dennis E. te Beest}
\address[Dennis E. te Beest]{
Biometris \\
Wageningen University \& Research \\
Wageningen \\
The Netherlands}

\author[]{Martijn W.\ Heymans}
\address[Martijn W.\ Heymans]{
Dept.\ of Epidemiology \& Biostatistics \\
Amsterdam Public Health research institute \\
Amsterdam University medical centers, location VUmc \\
Amsterdam\\
The Netherlands}

\author[]{Mark A.\ van de Wiel}
\address[Mark A.\ van de Wiel]{
Dept.\ of Epidemiology \& Biostatistics \\
Amsterdam Public Health research institute \\
Amsterdam University medical centers, location VUmc \\
Amsterdam\\
The Netherlands; \and
MRC Biostatistics Unit \\
Cambridge University \\
Cambridge \\
United Kingdom}

\begin{abstract}
\label{abstract}
~\\\noindent
\textbf{Motivation:} Radiomics refers to the high-throughput mining of quantitative features from radiographic images.
It is a promising field in that it may provide a non-invasive solution for screening and classification.
Standard machine learning classification and feature selection techniques, however, tend to display inferior performance in terms of (the stability of) predictive performance.
This is due to the heavy multicollinearity present in radiomic data.
We set out to provide an easy-to-use approach that deals with this problem.

\noindent
\textbf{Results:}
We developed a four-step approach that projects the original high-dimensional feature space onto a lower-dimensional latent-feature space, while retaining most of the covariation in the data.
It consists of (i) penalized maximum likelihood estimation of a redundancy filtered correlation matrix.
The resulting matrix (ii) is the input for a maximum likelihood factor analysis procedure.
This two-stage maximum-likelihood approach can be used to (iii) produce a compact set of stable features that (iv) can be directly used in any (regression-based) classifier or predictor.
It outperforms other classification (and feature selection) techniques in both external and internal validation settings regarding survival in squamous cell cancers.

\noindent
\textbf{Availability:} The \texttt{R} implementation of the pipeline (\texttt{FMradio}) as well as scripts for the data experiments are available at \url{https://github.com/CFWP/FMradio}.

\noindent
\textbf{Contact:} cf.peeters@vumc.nl

\noindent
\textbf{Supplementary information:} Supplementary information is available at XXX online.

\bigskip \noindent \footnotesize {\it Key words}:
Classification;
Factor analysis;
High-dimensional data;
Prediction models;
Radiomics
\end{abstract}

\maketitle

\begin{bibunit}

\section{Introduction}
\label{SEC:Intro}
\begin{sloppypar}

\subsection{Focus}
\label{SSEC:Focus}
Radiomics is a relatively recent addition to the omics-scene.
It encompasses the high-throughput mining of quantitative features from radiographic images.
Such images stem from technology such as computed tomography (CT), positron emission tomography (PET), and magnetic resonance imaging (MRI).
Introduced by \citet{SegalBeginRadio} and matured by \citet{RadiomicsOrigin}, it is rapidly finding widespread application, especially in oncology research \citep{AertsLHN,HuanAppCC,ChenAppLC}.
The promise of radiomic data is that it may be a basis, through the information contained in standard-of-care images, for non-invasive medical decision support at low (additional) cost.
As such, it is often viewed as an addition to or alternative for screening and classification based on molecular omics.
\end{sloppypar}

Irrespective of the medical imaging technology used, radiomics has a delineated workflow, roughly consisting of four steps.
The first is the acquisition of images and the identification of the volumes or regions of interest (VOIs).
The second step consists of image segmentation, i.e., the manual, computer-assisted or automated delineation of the borders of the VOIs.
The third step is feature extraction: the elicitation of features that characterize the segmented VOIs.
While this may be done in an unsupervised manner, the dominant approach is to use tailored extraction algorithms that rely on expert input.
Extracted features can range in the thousands \citep{KolossvaryCP2017} and enrich standard radiological lexicon (e.g., tumor extension, location, cellularity) with descriptors of distributions, textures, and morphology \citep{RadiomicsOverview}.
For an overview of (issues related to) steps one to three, see \citep{KumarOverview12,RadiomicsOverview,Yip16Limitations,RadiomicsReviewLambin,CookLimitations}

We will concern ourselves with the fourth step: downstream analysis.
While there are some committed informatical resources for radiomics, these deal with feature extraction mostly \citep[see, e.g., the overview in][]{Larue17Methods}.
Hence, to date, there has been no dedicated workflow for the downstream analysis of radiomic data.
Here, we will focus on providing such a workflow for classification and prediction problems.
This focus answers recent calls for more robust methods of analysis \citep{Aerts18DataScience} and aids in harnessing radiomics' translational potential.

\subsection{The collinearity problem}
\label{SSEC:Collinearity}
The main problem to overcome in radiomic-based classification and prediction is multicollinearity in the face of high-dimensional data.
Radiomic data are often high-dimensional, in the sense that the number of features, say $p$, exceeds the number of observations $n$.
Moreover, radiomic data typically deal with a heavy multicollinearity burden: (subsets of) features that are highly correlated or linearly dependent on other (subsets of) features.
Hence, there are two sources of ill-conditioning or singularity of the feature space that hamper estimation (stability) and feature selection.
The radiomics literature contains several approaches to deal with this problem.

The first is to use classifiers amended with a penalty on the feature space \citep[see, e.g.,][]{HuanAppCC}.
Arguably the most popular such penalty is the $\ell_1$ (lasso) penalty \citep{lasso} which performs automatic feature selection.
While attractive, $\ell_1$-regularized classifiers will not have a unique minimizer in the case of a heavy multicollinearity burden.
Moreover, its selection capability is constrained by the sample size $n$.
This combination can lead to unstable selection: multiple (non or only partly overlapping) sets of features with similar predictive performance.
The elastic net, which linearly combines an $\ell_1$ and an $\ell_2$ (ridge) penalty \citep{enet}, was developed to deal with lasso's shortcomings.
The effect of the ridge penalty is to force ``the estimated coefficients of highly correlated predictors to be close to each other" \citep{randomLasso}, after which the lasso (de)selects these features alltogether.
However, this behavior introduces heavy bias and inadequate prediction performance when faced with highly correlated features with a negative sign \citep{randomLasso}, a situation common in radiomics data.
Summarizing, the penalized classification approach deals with the problem of high-dimensionality rather than the problem of strong collinearity.

A second approach that is often encountered is to perform heuristic feature selection based on correlation redundancy \citep[see, e.g.,][]{Bala14CT,RadiomicsOverview}.
From a cluster of highly correlated features either a representative feature is chosen or the cluster is collapsed into a representative feature.
This approach often implies a loss of information as more features are removed or collapsed than necessary.
Moreover, it does not guarantee to alleviate the collinearity burden to such a degree that standard classifiers can be expected to give adequate estimates and prediction performance.

A third approach (sometimes used in conjunction with the second approach) is to select features on the basis of bivariate analysis between the individual features and the outcome of interest \citep[see, e.g.,][]{RadiomicsML}.
This approach consists of producing a ranking of the individual features on the basis of some test or measure of their bivariate association with the outcome.
A subset of features is then subsequently selected using a (often arbitrary) cut-off.
This approach has several problems.
As with the second approach, it does not guarantee to alleviate the collinearity burden to an acceptable degree.
Moreover, it does not reflect the behavior of features once they become part of a multi-feature classifier, as the dependencies between features are not considered.
This often results in unwanted feature inclusions and exclusions \citep{Sun96UniMulti}.
Lastly, bivariate filtering will only produce stable and reliable feature-sets when the features are uncorrelated, which is not the situation with radiomics data.

\subsection{Proposed approach}
\label{SSEC:PP}
What we desire is an approach that simultaneously deals with both the high-dimensionality as well as the collinearity burden of radiomics data.
Moreover, we desire an approach that produces stable and robust predictors that transfer well in validation settings.
We provide, through a four-step pipeline, such an approach based on projecting the radiomic feature-space onto a lower-dimensional (near-)orthogonal characteristic feature-space that retains most of the information contained in the full data set.
It consists of (i) penalized maximum likelihood (ML) estimation of a redundancy filtered correlation matrix.
The resulting matrix (ii) is the input for projection by a ML factor analysis procedure.
This two-stage maximum-likelihood (2SML) approach can be used to (iii) produce a compact set of stable characteristic features that (iv) can be directly used in any classifier or survival model.
The benefits of the approach are (a) computational speed and efficiency, (b) the ability to use standard downstream modeling and performance evaluation, (c) robustness of predictors, and (d) a potentially substantial increase in predictive performance.

The approach bears some flavor of principal component regression, in which first a principal component analysis (PCA) is performed after which a selection of retained components serve as regressors \citep{PCR}.
However, PCA differs conceptually from common factor analysis (FA) \citep{PeetersAEfa}:
PCA provides a deterministic transformation of the data by identifying weighted linear combinations of observed features that maximize the explanation of (solely the) observed variance.
FA, in contrast, seeks to explain the \emph{co}variation of and between the observed features through a small number of explanatory latent features (known as common factors).
Hence, FA provides a model-based transformation that includes measurement error.
As such, we favor FA over PCA.
Our approach to FA for high-dimensional data also differs from the mainstream.
Instead of inducing sparsity constraints on the (latent) parameter space \citep[e.g.,][]{Carvalho08SFA,Trend17SFA}, we regularize the sufficient statistic for usage in the (regular) maximization of the model-based likelihood.
As such, the proposed approach resembles the road taken by \citet{Yuan08SEMNS}.

\subsection{Overview}
\label{SSEC:Overview}
Section \ref{SEC:MandM} details the individual steps of the proposed approach.
Section \ref{SEC:Results} applies this approach to several real data sets.
It outperforms other classification (and feature selection) techniques in an MRI-imaging based external validation setting regarding survival in oropharyngeal cancer.
In addition, the approach gives the most stable performance in a PET/CT-imaging based internal validation setting regarding survival in head and neck cancer.
Section \ref{SEC:Discuss} concludes with a discussion.

\section{Materials and Methods}
\label{SEC:MandM}
\subsection{Setting}
\label{SSEC:Bground}
Let $\mathbf{Z}^{\mathrm{T}}\equiv[\boldsymbol{\mathrm{z}}_{1},\ldots,\boldsymbol{\mathrm{z}}_{n}]$ define $p$-variate vectors of radiomic scores on $i = 1, \ldots, n$ subjects with $\boldsymbol{\mathrm{z}}_{i}^{\mathrm{T}}\equiv[\mathrm{z}_{i1},\ldots,\mathrm{z}_{ip}]\in\mathbb{R}^{p}$ as a realization of the random vector $Z_{i}^{\mathrm{T}}\equiv[Z_{i1},\ldots,Z_{ip}]\in\mathbb{R}^{p}$.
We will assume that the data-matrix $\mathbf{Z}$ is column-wise centered and scaled.
The dependent variable of interest is contained in the vector $\boldsymbol{\mathrm{y}}$ and may designate either a continuous, a categorical (binary, nominal, or ordinal), a count, or a survival-type outcome.
Interest then lies in predicting or classifying $\boldsymbol{\mathrm{y}}$ using information contained in $\mathbf{Z}$.
The proposed pipeline is given in the next section.

\subsection{Pipeline overview}
\label{SSEC:Pipeline}
In general, the radiomic features will be highly collinear.
Moreover, in many clinical situations the data setting will be high-dimensional in the sense that $p > n$.
Hence, any standard matrix summarization of the independent data, such as the correlation matrix, will be either ill-behaved or singular.
The first step (i) is then to obtain a regularized estimator of the correlation matrix that is well-behaved.
Next, (ii) this regularized correlation matrix is the basis for factor-analytic data compression.
That is, the original feature-space is projected onto a (much) lower-dimensional latent feature-space.
Subsequently, (iii) factor-scores are obtained, expressing the score each person obtains on each latent feature.
These latent features can be understood as radiomic meta-features.
These factor-scores can then (iv) be used as predictors in standard (generalized linear) predictive modeling.
Below, each step is explained in more detail.

\subsubsection{Regularized correlation matrix estimation}
\label{SSSEC:CorM}
\begin{sloppypar}
The sample correlation matrix $\mathbf{R} = \mathbf{Z}^{\mathrm{T}}\mathbf{Z}/(n-1)$ will likely contain redundant information, in the sense that some entries will approach perfect (negative) correlation.
This implies that the information contained in one feature is almost completely represented by another feature.
Hence, only one of the redundant features needs to be retained.
Algorithm 1 from Section \ref{SMSSEC:Filtering} of the Supplementary Material (SM) contains a procedure for removing the minimal number of redundant features under absolute marginal correlation threshold $\tau$.
In practice, we recommend setting $\tau \in [.9,.95]$.
\end{sloppypar}

Denote the correlation matrix after redundancy filtering by $\dot{\mathbf{R}} \in \mathbb{R}^{p^{*} \times p^{*}}$, where $p^{*}$ is the number of features that passes filtering.
This matrix will still be ill-behaved or singular.
Hence, we will employ a penalized ML representation of $\dot{\mathbf{R}}$ \citep{Warton08}:
\begin{equation}\label{EQ:RR}
    \dot{\mathbf{R}}(\vartheta) = (1 - \vartheta)\dot{\mathbf{R}} + \vartheta\mathbf{I}_{p^{*}}
\end{equation}
where $\vartheta$ denotes a strictly positive penalty parameter.
This estimator is always positive definite, has well-defined limiting behavior, and is asymptotically consistent w.r.t.\ $\dot{\mathbf{R}}$ (see Section \ref{SMSSEC:RCM} of the SM).
For appropriate choices of $\vartheta$, this estimator is also well-conditioned.

Choosing the optimal value of $\vartheta$ is done in a data-driven way using $K$-fold cross-validation (CV).
The $K$-CV procedure is computationally efficient when noticing that (i) the estimator in (\ref{EQ:RR}) is rotation equivariant and (ii) one may employ a root-finding procedure.
This implies that an optimal value $\vartheta^{\ddagger}$ of $\vartheta$ can be found in the (worst-case) computational complexity of a single spectral decomposition: $\mathcal{O}(p^{*3})$.
See Section \ref{SMSSEC:CPP} of the SM for details.
Then, $\dot{\mathbf{R}}(\vartheta^{\ddagger})$ is used as the basis for the next step.

\subsubsection{Factor analytic data compression}
\label{SSSEC:FA}
FA assumes that observed variables can be grouped on the basis of their covariation or correlation into a lower-dimensional linear combination of latent features.
Let $\boldsymbol{\mathrm{z}}_{i}^{*}$ denote the radiomic observation vector that remains after removing the features implicated in redundancy filtering.
Then this linear combination can be expressed as:
\begin{equation}\nonumber
    \boldsymbol{\mathrm{z}}_{i}^{*} = \mathbf{\Lambda}\boldsymbol{\xi}_i + \boldsymbol{\epsilon}_i,
\end{equation}
where $\mathbf{\Lambda}$ is a $(p^{*}\times m)$-dimensional matrix of factor loadings in which each element $\lambda_{jk}$ is the loading of the $j$th variable on the $k$th factor, $j=1,\ldots,p^{*}$, $k=1,\ldots,m$, and where $\boldsymbol{\epsilon}_{i}$ denotes the error measurements for person $i$.
The $\boldsymbol{\xi}_{i}$ then represent realizations of a latent variable of dimension $m$, with $m < p^{*}$, whose elements are referred to as common factors.
Under the assumptions on the model (see Section \ref{SMSSEC:AFA} of the SM) $\boldsymbol{\mathrm{z}}_{i}^{*} \sim \mathcal{N}_{p^{*}}(\boldsymbol{0}, \mathbf{\Lambda}\mathbf{\Lambda}^{\mathrm{T}} + \mathbf{\Psi})$.
Hence, the correlation matrix among the (remaining) radiomic features is implied to be decomposable into common components ($\mathbf{\Lambda}\mathbf{\Lambda}^{\mathrm{T}}$) and unique components ($\mathbb{E}(\boldsymbol{\epsilon}_{i}\boldsymbol{\epsilon}_{i}^{\mathrm{T}}) = \boldsymbol{\Psi} \equiv \mbox{diag}(\psi_{11},\ldots, \psi_{pp})$).
The overarching assumption is thus that the population correlation matrix $\mathbf{\Sigma}_{\mathrm{R}}$ equals the model-implied population correlation $\mathbf{\Sigma}(\mathbf{\Theta}) = \mathbf{\Lambda}\mathbf{\Lambda}^{\mathrm{T}} + \mathbf{\Psi}$, where $\mathbf{\Theta} = \{\mathbf{\Lambda},\mathbf{\Psi}\}$.
From a likelihood perspective, the sample correlation matrix $\dot{\mathbf{R}}$ is a sufficient statistic (see Section \ref{SMSSEC:AFA} of the SM).
This insight has led to the covariance structure modeling approach that would use the following discrepancy function to find estimate $\hat{\mathbf{\Theta}}$ \citep[][and Section \ref{SMSSEC:AFA} of the SM]{Joreskog67}:
\begin{equation}\label{EQ:MinFunc}
    F[\mathbf{\Sigma}(\mathbf{\Theta});\dot{\mathbf{R}}] = \ln|\mathbf{\Sigma}(\mathbf{\Theta})| + \mathrm{tr}\left[\dot{\mathbf{R}}\mathbf{\Sigma}(\mathbf{\Theta})^{-1}\right] - \ln|\dot{\mathbf{R}}| - p^{*}.
\end{equation}
Minimizing (\ref{EQ:MinFunc}) requires a positive definite and, ideally, well-behaved sample correlation matrix.
Hence, we propose to replace $\dot{\mathbf{R}}$ by $\dot{\mathbf{R}}(\vartheta^{\ddagger})$ in (\ref{EQ:MinFunc}) in order to obtain estimate $\hat{\mathbf{\Theta}} = \{\hat{\mathbf{\Lambda}},\hat{\mathbf{\Psi}}\}$.
The algorithm for minimizing (\ref{EQ:MinFunc}) based on the ML normal equations $\partial F[\mathbf{\Sigma}(\mathbf{\Theta});\dot{\mathbf{R}}(\vartheta^{\ddagger})]/\partial\mathbf{\Lambda} = \boldsymbol{0}$ and $\partial F[\mathbf{\Sigma}(\mathbf{\Theta});\dot{\mathbf{R}}(\vartheta^{\ddagger})]/\partial\mathbf{\Psi} = \boldsymbol{0}$ is the Fletcher-Powell algorithm \citep{FPalgo63} introduced into FA by \citet{Joreskog67}.
This algorithm is widely used in many standard platforms performing FA.

Imperative is the selection of a value for $m$.
From the model-implied population correlation matrix and the assumptions on the factor model it follows that:
\begin{equation}
    \mathbf{\Sigma}(\mathbf{\Theta}) - \mathbf{\Psi} = \mathbf{\Lambda}\mathbf{\Lambda}^{\mathrm{T}},
\end{equation}
meaning that the reduced population correlation matrix is Gramian and of rank $m$ \citep[see, e.g., Chapter 8 of][]{Mulaik2010}.
The number of common factors can then be determined by assessing the rank of $\mathbf{\Sigma}(\mathbf{\Theta}) - \mathbf{\Psi}$ when replacing its constituents with appropriate estimates.
We use $\dot{\mathbf{R}}(\vartheta^{\ddagger})$ as the sampling counterpart to the population correlation matrix and $\mathbf{I}_p$ as a conservative estimate of the unique variance matrix.
Let $d(\mathbf{A})_j$ denote the $j$th eigenvalue of the matrix $\mathbf{A}$.
Some ready algebra on $\dot{\mathbf{R}}(\vartheta^{\ddagger}) - \mathbf{I}_p$ will then show that its eigenvalues are of the form $(1 - \vartheta^{\ddagger})[d(\dot{\mathbf{R}})_j - 1]$.
The basic decision rule is then to choose an optimal value of $m$, say $\tilde{m}$, by:
\begin{equation}\label{DRule}
    \tilde{m} := \mbox{card}(\mathrm{A}), \,\,\,\ \mbox{with}~ \mathrm{A} \equiv \Big\{j:(1 - \vartheta^{\ddagger})[d(\dot{\mathbf{R}})_j - 1] > 0\Big\}.
\end{equation}
Hence, we are simply determining the cardinality of the set of positive eigenvalues of $\dot{\mathbf{R}}(\vartheta^{\ddagger}) - \mathbf{I}_p$.
From the perspective of $\dot{\mathbf{R}}(\vartheta^{\ddagger}) - \mathbf{I}_p$, all variance is considered unique or error variance, implying that the average eigenvalue is $0$.
Thus, a positive eigenvalue $(1 - \vartheta^{\ddagger})[d(\dot{\mathbf{R}})_j - 1]$ indicates a latent factor dominating information content, as its contribution to variance-explanation is above and beyond mere unique variance.
We then retain all such factors.
This approach concurs with the Guttman-Kaiser rule \citep{Kaiser70}.
Section \ref{SMSSEC:DS} of the SM contains additional information on the approach to dimensionality selection.
Section \ref{SMSEC:SDS} of the SM contains an extensive comparative simulation indicating that the decision rule in (\ref{DRule}) can reliably function as an upper-bound estimate to the true generating latent dimension $m$.
We suggest that this procedure always be accompanied by substantive considerations (see Section \ref{SMSSEC:ADS} of the SM).
Now let $\hat{\mathbf{\Theta}}_{\tilde{m}} = \{\hat{\mathbf{\Lambda}},\hat{\mathbf{\Psi}}\}_{\tilde{m}}$ denote the ML estimate of the parameters under $\tilde{m}$ factors
In the remainder we will refer to the elements of $\hat{\mathbf{\Theta}}_{\tilde{m}}$ as $\hat{\mathbf{\Lambda}}$ and $\hat{\mathbf{\Psi}}$ in order to avoid notational clutter.

The orthogonal factor model considered copes with an inherent indeterminacy.
Consider an arbitrary orthogonal matrix $\mathbf{H} \in \mathbb{R}^{m \times m}$.
Then $\mathbf{\Sigma}(\mathbf{\Theta}_{m}) = \mathbf{\Lambda}\mathbf{\Lambda}^{\mathrm{T}} + \mathbf{\Psi} = (\mathbf{\Lambda H})(\mathbf{\Lambda H})^{\mathrm{T}} + \mathbf{\Psi}$.
Any method of estimation under $m$ factors thus requires $m(m - 1)/2$ restrictions on $\mathbf{\Lambda}$.
In the ML-procedure this is usually achieved by requiring that $\mathbf{\Lambda^{\mathrm{T}}}\mathbf{\Psi}^{-1}\mathbf{\Lambda}$ is a diagonal matrix and that its diagonal elements are ordered.
These restrictions are convenient from an estimation perspective, but do not necessarily carry substantive meaning.
Hence, after estimation, rotational mappings may be employed that satisfy certain criteria for interpretation.
We opt, for our solution under $\tilde{m}$ factors, for a rotation to orthogonal simple structure \citep{Thurstone47,Thurstone54}.
We do so by maximizing the normalized Varimax criterion (see Section \ref{SMSSEC:AFA} of the SM).
The Varimax-rotated solution, $\hat{\mathbf{\Lambda}}_{V}$, is conducive in the next step.

\subsubsection{Obtaining factor scores}
\label{SSSEC:Scores}
After projection of the original variable-space onto the lower-dimensional factor-space, we desire factor scores:
the score each individual obtains on each of the latent factors.
There are several methods for obtaining such scores.
Here, we regress $\mathbf{\Xi} \in \mathbb{R}^{n \times \tilde{m}}$ on $\mathbf{Z}^{*}$ to obtain \citep[][and Section \ref{SMSSEC:FS} of the SM]{Thomson}:
\begin{equation}\nonumber
    \hat{\mathbf{\Xi}} = \mathbf{Z}^{*}\hat{\mathbf{\Psi}}^{-1}\hat{\mathbf{\Lambda}}_{V}\left( \mathbf{I}_{\tilde{m}} + \hat{\mathbf{\Lambda}}_{V}^{\mathrm{T}}\hat{\mathbf{\Psi}}^{-1}\hat{\mathbf{\Lambda}}_{V} \right)^{-1}.
\end{equation}
Note that these scores are orthogonal by construction.
They will be used as predictors in a prediction rule.

\subsubsection{Prediction rules}
\label{SSSEC:Predict}
\begin{sloppypar}
The original variable-dimension is now projected onto a lower-dimensional space.
In most situations $\tilde{m} < n$.
Hence, we can estimate and evaluate any low-dimensional prediction rule.
We focus on regression-based prediction such as Cox regression for survival responses and the general linear model for categorical and linear responses, with:
\begin{equation}\nonumber
    g\left\{\mathbb{E}(\boldsymbol{\mathrm{y}})\right\} = \hat{\mathbf{\Xi}}\boldsymbol{\beta},
\end{equation}
where $\hat{\mathbf{\Xi}}\boldsymbol{\beta}$ denotes the predictor and where $g$ denotes a link function.
Note that our approach allows the usage of standard estimation (such as ML) and evaluation (such as ROC) methods.
\end{sloppypar}

\section{Results}
\label{SEC:Results}
\subsection{Data}
\label{SSEC:Data}
We have two data settings, both with a right-censored survival (time to event) outcome.
The response concerns realizations $T_{i}$ of a positive random variable $\mathcal{T}$ representing the time from starting point $t = 0$ to an event of interest.
That is, the response for person $i$ is $(\tilde{T}_{i},\delta_i)$.
In this response $\tilde{T}_{i}$ represents $\min(T_{i},C_{i})$, the minimum of event time $T_{i}$ and censoring time $C_{i}$.
Then $\delta_i = \mathrm{I}\left\{T_{i}\leq C_{i}\right\}$, a status indicator that is $1$ when the event of interest has occurred during the study-time and $0$ when the event of interest did not occur during the study-time or when the subject was lost to follow-up (right-censoring).
At $t = 0$ predictor (prognostic) features $\boldsymbol{\mathrm{z}}_{i}$ are available which, in our case, represent $p$ radiomic features.
The general aim would then be to predict survival probabilities based on the prognostic features available at baseline.

The first data setting concerns oropharyngeal squamous cell carcinoma (OPSCC), a common malignancy affecting the oropharynx.
The event of interest is death.
It is a validation setting in the sense that independent test and validation sets are available.
The test set comprises $n = 89$ malignancies and $47$ events and originated at the Amsterdam University medical centers, location VUmc, Amsterdam, the Netherlands.
The validation set comprises $n = 56$ malignancies and $28$ events and originated at the University Medical Center Utrecht, the Netherlands.
The feature-dimension concerns $p = 89$ radiomic features extracted from axial T1-weighted MRI scans.
Additional information on data acquisition and feature extraction can be found in \citet{Mes19} as well as Section \ref{SMSEC:ADIoscc} of the SM.

The second data setting concerns primary (hypopharyngeal, oropharyngeal, and laryngeal) tumors in head and neck squamous cell carcinoma (HNSCC) and the event of interest is again death.
In this case only one data set is available for model building and model evaluation.
We have $n = 174$ primary tumors and counted $55$ events.
The feature-dimension concerns $p = 432$ radiomic features extracted from low-dose $^{18}$F-FDG-PET/CT based images.
Additional information on data acquisition and radiomic feature extraction for this setting can be found in \citet{Martens19} as well as Section \ref{SMSEC:ADIhnscc} of the SM.

\subsection{Model comparison}
\label{SSEC:ModCompare}
In both settings we take interest in the predictive prowess of our pipeline.
The (training) data is used to obtain the projected factor scores.
These can then be directly used as the predictors in the Cox proportional hazards model \citep{Cox72}.
Note that the survival times are not used to guide the construction of the factor scores.
Hence, the projection approach is fully separated from the fitting of the model.
We take interest in comparing the pipeline approach with several other state-of-the-art ensemble and regularized variable selection methods.

For ensemble methods we consider survival forests.
These are nonparametric learning methods where the predictor is an ensemble formed by combining many survival decision trees.
There are, in general, 2 different types of survival forest and we will consider both: random survival forests \citep[RSFs;][]{RSF} and conditional inference forests for survival analysis or conditional survival forests for short \citep[CSFs;][]{CSF}.
They differ in how the ensemble is constructed.
RSFs construct the ensemble ``by aggregating tree-based Nelson-Aalen estimators" \citep{PEcurves}.
CSFs, on the other hand, use weighted Kaplan-Meier (KM) estimates, which tend to put more weight on terminal nodes with large numbers of subjects at risk \citep{PEcurves}.
\citet{RadiomicsML} has previously found that, in a radiomic prediction setting, RFs perform well in comparison to other methods, including PCA and partial least squares regression (which bear some of the flavor of our pipeline).
The main tuning parameter for RFs is the number of trees to grow, which we set to $1,000$ for both the RSF and CSF.
We set no restriction on the depth of the trees.
The remaining tuning parameters are used `factory fresh' (i.e., default settings) as implemented in the \texttt{party} \citep{party} and \texttt{randomForestSRC} \citep{randomForestSRC} packages in \texttt{R} \citep{Rman}.

For the variable selection methods we consider a componentwise likelihood-based offset boosting approach for the Cox proportional hazards model \citep{TB07}.
This can be seen as a regularized regression technique.
It starts with the null model and repeatedly fits (boosting) a base-learner in order to arrive at an adequate model.
At each boosting step only a single regression coefficient is updated.
The optimal updates in each step are determined by an approximation to (the maximization of) the penalized partial log-likelihood for Cox models.
As the non-updated regression coefficients are left at $0$ the approach effectively results in variable selection and a sparse final model.
More information can be found in \citep{Bind08} and \citep{Bind09}.
This Cox boosting approach was previously shown to outperform the Lasso in survival prediction settings by \citet{BSB14} and \citet{MHS16}.
The method relies on two tuning parameters: the shrinkage parameter in the penalized partial likelihood and the number of boosting steps.
The former can be chosen rather coarsely while the latter must not be too small or too large.
Both parameters are set according to the `factory fresh' settings as implemented in the \texttt{CoxBoost} package \citep{CoxBoost} (implying that the optimal number of boosting steps is chosen by 10-fold CV).

Hence, we are considering $a = 1, \ldots, A = 4$ models.
A cox model where the predictors are the projected latent metafeatures $\boldsymbol{\xi}_{i}$ (the pipeline approach), two types of RF and a Cox boosting model.
The latter three use the $\boldsymbol{\mathrm{z}}_{i}$ as predictors.
Let $\boldsymbol{\mathrm{p}}_{i}$ be a generic designation for either $\boldsymbol{\xi}_{i}$ or $\boldsymbol{\mathrm{z}}_{i}$.
Now, let $\hat{\pi}^{a}(t|\boldsymbol{\mathrm{p}}_{i})$ denote the predicted survival probability for individual $i$ at time $t$ under model $a$.
The (ensemble) survival functions $\pi^{a}(t|\boldsymbol{\mathrm{p}}_{i})$ for Cox regression and both forest-type models can be found, e.g., in \citet{PEcurves}.
They are central to the model evaluation approach discussed next.

\subsection{Model evaluation}
\label{SSEC:ModEvaluate}
For model evaluation we focus on prediction error through the time-dependent Brier score \citep{Brier,BrierModern}.
Let $\tilde{\mathrm{y}}_i(t) = \mathrm{I}\{\tilde{T}_{i} \geq t\}$ denote the observed survival status of subject $i$ at time $t$.
The empirical Brier score may then be seen as a mean square error of prediction when $\hat{\pi}^{a}(t|\boldsymbol{\mathrm{p}}_{i}) \in [0,1]$ is taken to be a prediction of the survival status $\tilde{\mathrm{y}}_i(t) \in \{0,1\}$ \citep{BrierModern}.
Hence, it is a measure of inaccuracy.
See \citet{BrierModern} for reasoning regarding the appropriateness of the Brier score over the $c$-index in survival settings.
Section \ref{SMSEC:ModEval} of the SM contains details on the calculation of the (integrated) Brier score in both our external (data setting 1) and internal (data setting 2) validation settings.

The models under consideration will then be compared w.r.t.\ (integrated) Brier scores in the following manner.
First, they will be compared to each other noting that, as a measure of inaccuracy, lower Brier scores are better.
Second, they will be compared w.r.t.\ their respective apparent errors.
The apparent error is the prediction error obtained when the data used from training are re-used for validation, which will result in optimistic prediction errors.
The smaller the gap between apparent and (cross-)validated error, the more stable a model (approach) can be considered to be.
Third, they will be compared to the benchmark value of $.25$, which corresponds to predicting $50\%$ risk constantly over time \citep{SBG07}.
One desires a model to do better than chance over the lion's share of $t \in [0, \tau]$, where $\tau \leq \max(T_{i})$.
We take $\tau$ to be the median follow-up time \citep{BrierModern}.
Fourth, they will be compared to the prediction error based on the KM prediction rule, which ignores all feature information and as such acts as a null model \citep{BrierModern,SBG07,PEcurves}.
One desires a model to do considerably better than the null model.
These considerations are all incorporated in prediction error curves, where the time-dependent Brier scores are plotted against time.
Section \ref{SMSSEC:R2} shows how the (integrated) Brier score can be used to calculate a measure of (overall) explained residual variation: $R^{2}$.
The higher the $R^{2}$ measure, the better, giving a fifth ground for comparison of the models under consideration.

\subsection{External validation performance}
\label{SSEC:PerformanceExternal}
The redundancy filtering algorithm was run on the training data with $\tau$ set to $.95$ and it retained $p^* = 51$ out of the original $p = 89$ features.
Subsequently, using 5-fold CV of the log-likelihood function, the optimal penalty-value for the regularized correlation matrix was determined (Step 1).
For the factor-analytic data compression, $\tilde{m}$ was set to $7$.
Subsequently, the (rotated) ML estimates of the factor loadings and error covariance matrices were determined for the training data (Step 2), say $\hat{\mathbf{\Lambda}}_{V}$ and $\hat{\mathbf{\Psi}}$.
Factor scores were then determined (Step 3) as $\hat{\boldsymbol{\xi}}_i = \left(\mathbf{I}_{7} + \hat{\mathbf{\Lambda}}^{\mathrm{T}}_{V}\hat{\mathbf{\Psi}}^{-1}\hat{\mathbf{\Lambda}}_{V}\right)^{-1}\hat{\mathbf{\Lambda}}^{\mathrm{T}}_{V}\hat{\mathbf{\Psi}}^{-1}\boldsymbol{\mathrm{z}}_{i}^{*}$.
Let $D$ be a data (subset) indicator and let $D_T$ indicate the training set while $D_V$ indicates the external validation set.
Using $i \in D_T$ will then give the factor scores for the training data with which apparent performance can be assessed, while using $i \in D_V$ gives the factor scores for the validation data (based on the factor solution of the training data) with which validation performance is assessed.
Section \ref{SMSSEC:OFP} of the SM gives additional information on Steps 1--3.

Table \ref{TAB:ResultsEVD} gives an overview of the apparent and validated integrated Brier scores and explained residual variations for the proposed approach as well as the models mentioned in Section \ref{SSEC:ModCompare}.
All methods perform better than chance and the KM reference model.
The \texttt{FMradio} approach has the lowest validation prediction error.
This translates into the \texttt{FMradio} approach having the highest integrated $R^{2}$.
Its performance is also amongst the most stable performances as indicated by the relatively small difference between its apparent and validation prediction errors.
Figure \ref{SMFIG:ResultsEVD} in Section \ref{SMSSEC:ORES} of the SM visualizes these results.
Especially the RSF behaves unstably, with great apparent performance but severely diminished validation performance.

\begin{table}[h!]
\caption{Integrated apparent and validated Brier scores and explained residual variations in the external validation setting.}
\label{TAB:ResultsEVD}
\begin{tabular}{lcclcc}
\toprule
                            & \multicolumn{2}{c}{$\mathcal{B}^{I}$} &  & \multicolumn{2}{c}{$R^{2}$} \\ \cline{2-3} \cline{5-6}
                            & Apparent      & Validated          &  & Apparent         &  Validated            \\ \midrule
Reference model             & .160          & .168               &  & --               & --                    \\
\texttt{FMradio}            & .129          & \textbf{.154}      &  & .197             & \textbf{.080}         \\
Conditional survival forest & .107          & .156               &  & .333             & .071                  \\
Random survival forest      & .089          & .159               &  & .441             & .053                  \\
Cox boosting                & .140          & .161               &  & .129             & .042                  \\
\bottomrule
\end{tabular}
\end{table}

The \texttt{FMradio} approach is subject to the strictest evaluation of validation performance: not only are the validation data evaluated at the Cox parameter-estimates stemming from the training sample, also the validation data projection is informed by the training-based factor solution.
In determining the new factor scores through the training factor-solution, we are recalibrating the positioning of the validation observations w.r.t.\ the latent dimensions determined in the training.
In this respect one could argue that, with shifts in latent-trait positioning, one should also recalibrate the Cox parameter-estimates.
If we would do so the recalibrated integrated prediction error of the \texttt{FMradio} approach would drop to $.134$ while its integrated $R^{2}$ would raise to $.2$ (Table \ref{TAB:ResultsEVDrecal}).
For comparison purposes we also recalibrated the one other approach that allows for ready recalibration: Cox boosting.
The validation data were subsetted to retain those features selected by the boosting approach in the training phase.
Subsequently a Cox proportional hazards model was fitted on these data.
Its performance was then assessed on again the validation data.
Note that the feature-subsetting in the Cox Boosting recalibration makes use of the outcome information from the training set, whereas the factor-projection is fully independent from the outcome measurements.
Hence, this gives the Cox boosting recalibration a certain advantage.
Nevertheless, the recalibrated \texttt{FMradio} approach retains its superior performance over the recalibrated boosting approach (Table \ref{TAB:ResultsEVDrecal}).

\begin{table}[h!]
\caption{Recalibrated integrated validated Brier scores and explained residual variations in the external validation setting.}
\label{TAB:ResultsEVDrecal}
\begin{tabular}{lccc}
\toprule
                            & Recalibrated $\mathcal{B}^{I}$        &  & Recalibrated $R^{2}$          \\ \cmidrule{2-2} \cmidrule{4-4}
\texttt{FMradio}            & \textbf{.134}                         &  & \textbf{.200}                \\
Cox boosting                & .149                                  &  & .109                          \\
\bottomrule
\end{tabular}
\end{table}

\subsection{Internal validation performance}
\label{SSEC:PerformanceInternal}
The redundancy filtering algorithm in the second data setting was also run with $\tau$ set to $.95$.
It retained $p^* = 124$ out of the original $p = 432$ features.
For the factor-analytic data compression, $\tilde{m}$ was set to $8$.
Subsequently, the factor scores were determined based on the ML factor solution.
Section \ref{SMSSEC:HNFP} of the SM gives additional information on Steps 1--3 in this second data setting.

\begin{figure}[h!]
\centering
  \includegraphics[width=\textwidth]{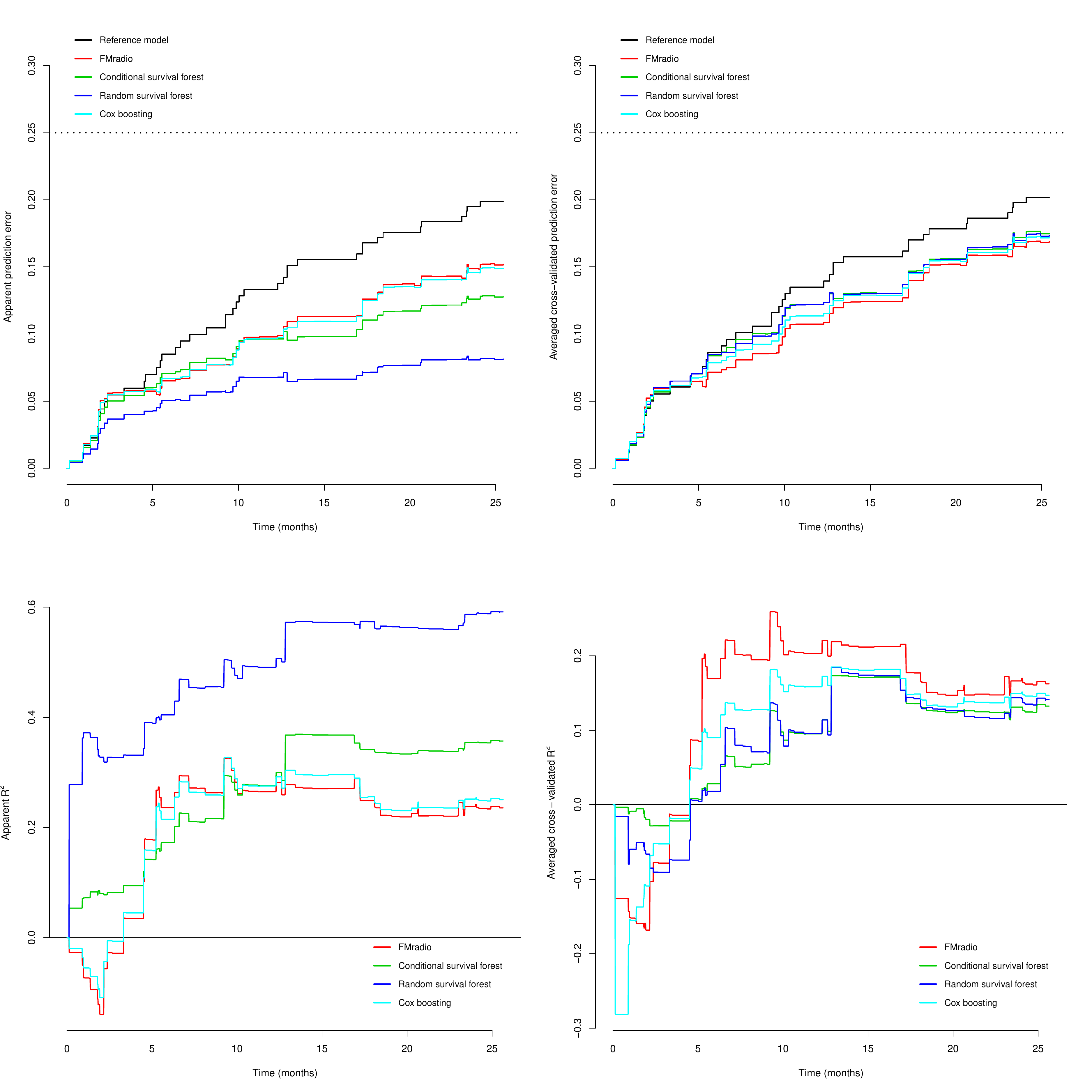}
    \caption{Visualizations for the internal validation setting.
    The upper-panels contain prediction error curves.
    The bottom-panels contain $R^{2}$ plots.
    The left-hand panels pertain to apparent performance while the right-hand panels visualize the averaged cross-validation performance.}
  \label{FIG:ResultsIVD}
\end{figure}

Table \ref{TAB:ResultsIVD} gives an overview of the apparent and cross-validated integrated Brier scores and explained residual variations.
These results were obtained with $K = 5$ and $B = 500$ (see Section \ref{SMSSEC:PredErr} of the SM).
Again, all methods perform better than chance and the KM reference model.
Also, again the \texttt{FMradio} approach has the lowest validation prediction error and the most stable performance (smallest difference between its apparent and validation prediction errors) which again translates into the highest (and most stable) integrated $R^{2}$.
Figure \ref{FIG:ResultsIVD} visualizes these results.

\begin{table}[h!]
\caption{Integrated apparent and averaged cross-validated Brier scores and explained residual variations in the internal validation setting.}
\label{TAB:ResultsIVD}
\begin{tabular}{lcclcc}
\toprule
                            & \multicolumn{2}{c}{$\mathcal{B}^{I}$} &  & \multicolumn{2}{c}{$R^{2}$} \\ \cline{2-3} \cline{5-6}
                            & Apparent      & Cross-validated    &  & Apparent         &  Cross-validated      \\ \midrule
Reference model             & .128          & .130               &  & --               & --                    \\
\texttt{FMradio}            & .098          & \textbf{.108}      &  & .236             & \textbf{.169}         \\
Conditional survival forest & .089          & .115               &  & .306             & .114                  \\
Random survival forest      & .060          & .115               &  & .529             & .114                  \\
Cox boosting                & .096          & .112               &  & .247             & .138                  \\
\bottomrule
\end{tabular}
\end{table}

One could argue that the performance of \texttt{FMradio} vis-\`{a}-vis the other approaches is due to the redundancy filtering, which removed many features.
To assess this the analysis was repeated with the RSF, CSF, and Cox boosting approaches based on the same redundancy-filtered feature-set as the \texttt{FMradio} approach.
Section \ref{SMSSEC:HNFALL} of the SM contains the results which are qualitatively the same as above: The most stable approach resulting in the lowest prediction error remains the proposed \texttt{FMradio} approach.

\section{Discussion}
\label{SEC:Discuss}
We present a new approach for classification and prediction with radiomic data.
A 2-stage ML factor-analytic projection is proposed that summarizes the high-dimensional collinear feature-space in a low-dimensional orthogonal meta-feature space.
This enables standard down-stream estimation and evaluation of radiomic-feature-based classification and prediction models.
Moreover, it results in stable prediction models by directly dealing with the strong collinearity problem characteristic of radiomic data.
This approach then outperforms, in terms of (the stability of) prediction error, the most popular current approaches in radiomic data analysis.
While we have focussed on survival settings in our examples, we stress that any prediction model can use the meta-features obtained by our projection.

We also stress the distinct advantages of our approach.
First, as the meta-features are projections of amalgams of individual features, they are more generalizable within the radiomic context.
This generalizability will---for a given set of latent projections---improve as the feature-dimension grows.
Second, the projection enables the prediction itself to be performed in a low-dimensional context.
Besides stability, this ensures low computation times and allows, in contrast to the other methods, for the quantification of uncertainty (such as through confidence intervals on the parameter estimates).
Third, as the projection is fully independent from the outcome measurements it is possible to recalibrate the parameter estimates without overfitting.
This may lead to considerable further improvements in performance.
Fourth, the meta-features are often interpretable.
This gives the proposed approach an advantage over ensemble-type methods, as interpretability in the clinical setting is often desired.

We see several inroads for further research and improvements.
First, we have worked under the assumption that the (standardized) radiomic features are normally distributed.
In case this assumption is untenable one could base $\mathbf{R}$, and subsequently $\dot{\mathbf{R}}$, on either a rank-type correlation or on a semiparametric Gaussian copula model \citep{NonPara}.
Second, the penalized correlation matrix may be extended with substantive prior information.
In a sense, the estimator $\dot{\mathbf{R}}(\vartheta)$ is a Bayesian estimator, balancing the ill-behaved $\dot{\mathbf{R}}$ with the well-behaved $\mathbf{I}_{p^{*}}$.
Instead of $\mathbf{I}_{p^{*}}$ one could also use more informative matrices (on the correlation scale) representing more elaborate prior information, possibly stemming from previous radiomic studies.
These considerations might further improve the predictive performance of the proposed approach.

We also note several immediate extensions to the approach.
While we focussed on using all extracted meta-features in the prediction process one could indeed also embark on feature-selection amongst these meta-features.
Moreover, we have limited ourselves to using only radiomic meta-features for prediction.
However, one can immediately combine these features with other clinical variables to possibly further improve predictive performance.
One would then be evaluating
\begin{equation}\nonumber
    g\left\{\mathbb{E}(\boldsymbol{\mathrm{y}})\right\} = \hat{\mathbf{\Xi}}\boldsymbol{\beta} + \mathbf{C}\boldsymbol{\beta}'.
\end{equation}
Lastly, the approach could also be of interest for other data settings in which the features are subject to a natural grouping, such as, for example, in metabolomics.


\bigskip
\subsection*{Acknowledgements}
We thank Ruud Brakenhoff and Ren\'{e} Leemans of the department of Otolaryngology/Head \& Neck Surgery of the Amsterdam University medical centers as well as Remco de Bree of the department of Head and Neck Surgical Oncology of the University Medical Center Utrecht, for sharing their data with us.
We also thank the study participants.
The data cohorts in the external validation study have been obtained in accordance with the Declaration of Helsinki and were approved by the Medical Ethics Committees of the associated medical centers.
All patients provided a signed informed consent.
The trials for the internal validation study have also been performed in accordance with the Declaration of Helsinki, approved by the Medical Ethics Committee of the VU University Medical Center.
A written informed consent was waived for the retrospective cohort (103 patients, reference: 2016.498), whereas for the prospective cohort (72 patients, reference: 2013.191) a written informed consent was obtained from all patients.

\subsection*{Funding}
This work was financially supported by the Netherlands Organisation for Health Research and Development (ZonMw), grant 10-10400-98-14002.

\subsubsection*{Conflict of interest:}
None declared.
\bigskip

\putbib[Radiomics_Main]
\end{bibunit}

\vspace{1cm}
\addresseshere

\cleardoublepage

\renewcommand{\theequation}{S.\arabic{equation}}
\renewcommand{\thefigure}{S.\arabic{figure}}
\renewcommand{\thetable}{S.\arabic{table}}
\renewcommand{\thealgorithm}{S.\arabic{algorithm}}
\renewcommand{\bibnumfmt}[1]{[S.#1]}
\renewcommand{\citenumfont}[1]{S.#1}
\renewcommand{\thesection}{S.\arabic{section}}
\renewcommand{\thepage}{S.\arabic{page}}

\setcounter{section}{0}
\setcounter{subsection}{0}
\setcounter{equation}{0}
\setcounter{figure}{0}
\setcounter{table}{0}
\setcounter{algorithm}{0}
\setcounter{page}{1}

\phantomsection
\addcontentsline{toc}{section}{Supplementary Material}
\begin{center}
{\huge SUPPLEMENTARY MATERIAL\\~\\
Stable Prediction with Radiomics Data}
\end{center}

\begin{bibunit}
\vspace{2cm}
This supplement contains text, figures, and \textsf{R} code in support of the Main Text.
Section \ref{SMSEC:Pdetails} contains additional technical information on various steps of the proposed pipeline.
Section \ref{SMSEC:ModEval} contains details on the evaluation of predictive performance.
Section \ref{SMSEC:SDS} contains a description of simulation efforts regarding the selection of an appropriate latent-dimensionality.
Sections \ref{SMSEC:ADIoscc} and \ref{SMSEC:ADIhnscc} contain additional information on the data used in the Main Text.
\textsf{R} code with which all analyzes and simulations can be repeated are found on \url{https://github.com/CFWP/FMradio}.

\section{Pipeline details}
\label{SMSEC:Pdetails}
Section \ref{SMSSEC:Filtering} contains the redundancy filtering algorithm.
Section \ref{SMSSEC:RCM} gives details on the regularized estimator for the correlation matrix.
Subsequently, Section \ref{SMSSEC:CPP} explains how the regularization parameter is chosen.
Section \ref{SMSSEC:AFA} then delves into the assumptions underlying the factor analysis model.
Sections \ref{SMSSEC:DS} and \ref{SMSSEC:FS} show how to select the dimension of the latent vector and how to obtain factor scores.
Most of the development will be in the general notation $p$ and $m$ for the feature and latent dimensions, respectively.
These may be replaced, without loss of generality, by $p^{*}$ and $\tilde{m}$.

\subsection{Redundancy filtering}
\label{SMSSEC:Filtering}
Algorithm \ref{AlgRF} contains pseudocode for redundancy filtering of a correlation matrix.

\begin{algorithm}[H]
\caption{(Redundancy filter).}\label{AlgRF}
\begin{algorithmic} [1]
\Require $\mathbf{R} \in \mathbb{R}^{p \times p}$ \Comment{Raw correlation matrix}

\Require $\tau$ \Comment{Thresholding value}

\bigskip

\Procedure{RF}{$\mathbf{R}$, $\tau$}
\State Go $= \mathrm{TRUE}$ \Comment{Initializing logical}
\State create $\boldsymbol{v}$ \Comment{Empty vector}

\While{Go}

\For{$j=1$ to row-dimension $\mathbf{R}$}
\State $\boldsymbol{v}[j] \gets \sum_{j'}\mathds{1}_{\left\{|\mathbf{R}[j,j']| \geq \tau\right\}}$
\EndFor
\State $c \gets \mathrm{which}(\boldsymbol{v} = \max(\boldsymbol{v}))[1]$ \Comment{Retain index}
\If{$\max(\boldsymbol{v}) < 2$}
\State Go $= \mathrm{FALSE}$
\Else
\State $\mathbf{R} \gets \mathbf{R}[-c,-c]$ \Comment{Removing corresponding row and column}
\State empty $\boldsymbol{v}$
\EndIf

\EndWhile

\Return{$\dot{\mathbf{R}} \in \mathbb{R}^{p^{*} \times p^{*}}$}
\EndProcedure

\end{algorithmic}
\end{algorithm}

Consider the following toy example explaining the workings of Algorithm \ref{AlgRF}.

\begin{example}
\label{SMEX:AlgoExample}
Say we have the following $4 \times 4$ correlation matrix on variables $A$, $B$, $C$, and $D$:
\begin{equation}\nonumber
\begin{blockarray}{ccccc}
& A & B & C & D\\
\begin{block}{c[cccc]}
  A & ~1 & .95 & .95 & .30~ \\
  B & ~.95 & 1 & .30 & .30~ \\
  C & ~.95 & .30 & 1 & .95~ \\
  D & ~.30 & .30 & .95 & 1~ \\
\end{block}
\end{blockarray}.
\end{equation}
Assume that we consider absolute correlations $\geq .95$ to indicate redundancy.
It is clear that, if we would blindly remove all columns and corresponding rows in which an (absolute) correlation appears that abides the redundancy criterion, we would be left with no matrix at all.
Also, if we would treat this matrix as representing a collinear block, then choosing one representative feature would also imply a loss of information as no feature represents all other features in terms of information.
For example, if we would choose feature $A$ as the representative feature then we would loose much information regarding feature $D$: $A$ is highly collinear with features $B$ and $C$, but not with $D$.

Hence, we apply the RF procedure of Algorithm \ref{AlgRF}.
It starts by evaluating, for each row, the number of times the corresponding feature exhibits an absolute correlation $\geq .95$.
This information is collected in the vector $\boldsymbol{v}$.
For our correlation matrix this evaluation would amount to:
\begin{equation}\nonumber
\begin{blockarray}{ccccccc}
& A & B & C & D & & \boldsymbol{v}[j] \gets \sum_{j'}\mathds{1}_{\left\{|\mathbf{R}[j,j']| \geq \tau\right\}}\\
\begin{block}{c[cccc]cc}
  A & ~1   & .95 & .95 & .30\,\, ~& ~& \boldsymbol{v}[1] = 3\\
  B & ~.95 & 1   & .30 & .30\,\, ~& ~& \boldsymbol{v}[2] = 2\\
  C & ~.95 & .30 & 1   & .95\,\, ~& ~& \boldsymbol{v}[3] = 3\\
  D & ~.30 & .30 & .95 &   1\,\, ~& ~& \boldsymbol{v}[4] = 2\\
\end{block}
\end{blockarray}.
\end{equation}
The feature(s) that have the most absolute correlations exceeding the threshold can be thought of a having their information represented by the most alternative features.
Hence, such features are natural candidates for removal.
The algorithm then evaluates for which indices (corresponding to row and column numbers) $\boldsymbol{v}$ attains a maximum: $\mathrm{\texttt{which}}(\boldsymbol{v} = \max(\boldsymbol{v}))$.
In this case $\mathrm{\texttt{which}}(\boldsymbol{v} = \max(\boldsymbol{v})) = [1,3]$, as both feature A and feature C have $3$ (absolute) correlations exceeding our threshold.
In the situation where $\mathrm{\texttt{which}}(\boldsymbol{v} = \max(\boldsymbol{v}))$ is a vector of length greater than $1$, the first index in this vector is retained, i.e.: $c \gets \mathrm{\texttt{which}}(\boldsymbol{v} = \max(\boldsymbol{v}))[1]$.
Thus, at current, $c = 1$.
Then, as at this point, $\max(\boldsymbol{v}) = 3 > 2$ (i) an updated $\mathbf{R}$ is formed by removing the first row and column ($\mathbf{R} \gets \mathbf{R}[-1,-1]$), corresponding to feature $A$, (ii) $\boldsymbol{v}$ is emptied, and (iii) a new round of evaluation is started.

The new round of evaluation starts with the updated correlation matrix.
The vector $\boldsymbol{v}$ is now evaluated as:
\begin{equation}\nonumber
\begin{blockarray}{cccccc}
& B & C & D & & \boldsymbol{v}[j] \gets \sum_{j'}\mathds{1}_{\left\{|\mathbf{R}[j,j']| \geq \tau\right\}}\\
\begin{block}{c[ccc]cc}
  B & 1   & .30 & .30\,\, ~& ~& \boldsymbol{v}[1] = 1\\
  C & .30 & 1   & .95\,\, ~& ~& \boldsymbol{v}[2] = 2\\
  D & .30 & .95 &   1\,\, ~& ~& \boldsymbol{v}[3] = 2\\
\end{block}
\end{blockarray}.
\end{equation}
Now, $\mathrm{\texttt{which}}(\boldsymbol{v} = \max(\boldsymbol{v})) = [2,3]$ and $c \gets \mathrm{\texttt{which}}(\boldsymbol{v} = \max(\boldsymbol{v}))[1] = 2$.
As $\max(\boldsymbol{v}) = 2$ (i) an updated $\mathbf{R}$ is formed by removing the second row and column ($\mathbf{R} \gets \mathbf{R}[-2,-2]$), corresponding to feature $C$, (ii) $\boldsymbol{v}$ is emptied, and (iii) a new round of evaluation is started.
Again, the subsequent round of evaluation starts with the updated correlation matrix.
This matrix and the corresponding vector $\boldsymbol{v}$ are now given as:
\begin{equation}\label{SMEQ:FinalCor}
\begin{blockarray}{ccccc}
& B & D & & \boldsymbol{v}[j] \gets \sum_{j'}\mathds{1}_{\left\{|\mathbf{R}[j,j']| \geq \tau\right\}}\\
\begin{block}{c[cc]cc}
  B & 1   &  .30\,\, ~& ~& \boldsymbol{v}[1] = 1\\
  D & .30 &    1\,\, ~& ~& \boldsymbol{v}[2] = 1\\
\end{block}
\end{blockarray}.
\end{equation}
As now $\max(\boldsymbol{v}) < 2$, we have that, logically, only the diagonal elements of the remaining matrix exceed the threshold.
Hence, the evaluation stops and the correlation matrix in (\ref{SMEQ:FinalCor}) is returned as the final, redundancy filtered correlation matrix.

In this final matrix retained feature $B$ also represents most of the information contained in removed feature $A$, and retained feature $D$ also represents most of the information contained in removed feature $C$.
Thus, in a sense, the final matrix represents all original features.
Note that, while $c \gets \mathrm{\texttt{which}}(\boldsymbol{v} = \max(\boldsymbol{v}))[1]$ may be termed arbitrary in case $\mathrm{\texttt{which}}(\boldsymbol{v} = \max(\boldsymbol{v}))$ is a vector of length greater than $1$, alternative choices would lead to retainment of roughly the same total information.
For example, say we would have used $c \gets \mathrm{\texttt{which}}(\boldsymbol{v} = \max(\boldsymbol{v}))[2]$ in each case where $\mathrm{\texttt{length}}(\mathrm{\texttt{which}}(\boldsymbol{v} = \max(\boldsymbol{v}))) > 1$.
This would have led to the removal of feature $C$ in the first step and feature $B$ in the second, giving the final correlation matrix:
\begin{equation}\nonumber
\begin{blockarray}{ccc}
& A & D\\
\begin{block}{c[cc]}
  A & ~1 &  .30~ \\
  D & ~.30  & 1~ \\
\end{block}
\end{blockarray}.
\end{equation}
In this alternative final matrix removed feature $B$ is represented by retained feature $A$ and removed feature $C$ is represented by retained features $A$ and $D$.
\QEDE
\end{example}

The algorithm can be expanded by including the sum of absolute correlations exceeding the threshold in the assignment of index $c$ whenever $\mathrm{\texttt{length}}(\mathrm{\texttt{which}}(\boldsymbol{v} = \max(\boldsymbol{v}))) > 1$.
This would imply bookkeeping of an extra vector.

\subsection{Regularized correlation matrix estimation}
\label{SMSSEC:RCM}
Here we give some details on the regularized correlation matrix estimation used in the Main Text.
Section \ref{SMSSSEC:PML} explains how the estimator may be understood as a penalized maximum likelihood estimator.
Section \ref{SMSSEC:CPP} details on selecting a value for the penalty parameter.

\subsubsection{Penalized maximum likelihood}
\label{SMSSSEC:PML}
We are considering sample correlation matrices.
Let $\boldsymbol{\mathrm{x}}_{i}^{\mathrm{T}}\equiv[\mathrm{x}_{i1},\ldots,\mathrm{x}_{ip}]\in\mathbb{R}^{p}$ define a realization of the random vector $X_{i}^{\mathrm{T}}\equiv[X_{i1},\ldots,X_{ip}]\in\mathbb{R}^{p}$.
Then the sample covariance matrix can be obtained as
\begin{equation*}
    \mathbf{S} = (n - 1)^{-1}\sum_{i=1}^{n}(\boldsymbol{\mathrm{x}}_i - \bar{\boldsymbol{\mathrm{x}}})(\boldsymbol{\mathrm{x}}_i - \bar{\boldsymbol{\mathrm{x}}})^{\mathrm{T}},
\end{equation*}
where $\bar{\boldsymbol{\mathrm{x}}} = n^{-1}\sum_{i=1}^{n} \boldsymbol{\mathrm{x}}_i$.
The correlation matrix can then be obtained directly from $\mathbf{S}$ or from the standardized sample realizations $(\mathbf{S} \circ \mathbf{I}_p)^{-1/2}(\boldsymbol{\mathrm{x}}_i - \bar{\boldsymbol{\mathrm{x}}}) \equiv \boldsymbol{\mathrm{z}}_i$ as:
\begin{equation*}
    \mathbf{R} = (\mathbf{S} \circ \mathbf{I}_p)^{-1/2}\mathbf{S}(\mathbf{S} \circ \mathbf{I}_p)^{-1/2} =
    (n-1)^{-1}\sum_{i=1}^{n}\boldsymbol{\mathrm{z}}_i\boldsymbol{\mathrm{z}}_i^{\mathrm{T}}.
\end{equation*}
If $\boldsymbol{\mathrm{x}}_i \sim \mathcal{N}_p(\boldsymbol{\mu}, \mathbf{\Sigma})$ then, under the stated data transformation, $\boldsymbol{\mathrm{z}}_i \sim \mathcal{N}_p(\boldsymbol{0}, \mathbf{\Sigma}_{\mathrm{R}})$, where $\boldsymbol{\mu}, \mathbf{\Sigma}$, and $\mathbf{\Sigma}_{\mathrm{R}}$ respectively refer to the population mean, the population covariance matrix, and the population correlation matrix.
It is well-known that, if all realizations $\boldsymbol{\mathrm{x}}_i \sim \mathcal{N}_p(\boldsymbol{\mu}, \mathbf{\Sigma})$, then $\mathbf{S}$ has a Wishart distribution with scale matrix $\mathbf{\Sigma}/(n-1)$ and $n-1$ degrees of freedom, i.e., $\mathbf{S} \sim \mathcal{W}_p\left(n-1,\mathbf{\Sigma}/(n-1)\right)$.
While the exact distribution of $\mathbf{R}$ is unknown, it can be well-approximated by the Wishart distribution \citep{KR03_SM}.

Hence, we assume $\mathbf{R} \sim \mathcal{W}_p\left(n-1,\mathbf{\Sigma}_{\mathrm{R}} /(n-1)\right)$.
The negative log-likelihood of $\mathbf{R}$ (up to proportionality) is then:
\begin{equation}\label{SMEQ:LLR}
    \mathcal{L}(\mathbf{\Sigma}_{\mathrm{R}}; \mathbf{R}) \propto \ln|\mathbf{\Sigma}_{\mathrm{R}}| + \tr\Big[\mathbf{R}\mathbf{\Sigma}_{\mathrm{R}}^{-1}\Big].
\end{equation}
If we would minimize (\ref{SMEQ:LLR}) w.r.t.\ $\mathbf{\Sigma}_{\mathrm{R}}$ we would find that $\mathbf{R}$ is the maximum likelihood (ML) estimator of $\hat{\mathbf{\Sigma}}_{\mathrm{R}}$.
This estimator will be ill-conditioned when $p \rightarrow n - 1$ or in the presence of strong multicollinearity and it will be singular when $p > n-1$.
These issues can be resolved if we adapt (\ref{SMEQ:LLR}) by (a) rescaling $\mathbf{R}$ with $(1 - \vartheta)$ where $\vartheta \in [0,1]$, and (b) adding the penalty $\vartheta\tr(\mathbf{\Sigma}_{\mathrm{R}})^{-1}$.
We then have a penalized log-likelihood
\begin{equation}\nonumber
    \mathcal{L}^{p}(\mathbf{\Sigma}_{\mathrm{R}}; \mathbf{R}, \vartheta) \propto \ln|\mathbf{\Sigma}_{\mathrm{R}}| + \tr\left[(1-\vartheta)\mathbf{R}\mathbf{\Sigma}_{\mathrm{R}}^{-1}\right] + \vartheta\tr\left(\mathbf{\Sigma}_{\mathrm{R}}^{-1}\right),
\end{equation}
whose minimization w.r.t.\ $\mathbf{\Sigma}_{\mathrm{R}}$ conditional on $\vartheta$ results in:
\begin{equation}\label{SMEQ:MINarg}
    \hat{\mathbf{\Sigma}}_{\mathrm{R}}(\vartheta) :=
    (1 - \vartheta)\mathbf{R} + \vartheta\mathbf{I}_p \equiv \mathbf{R}(\vartheta),
\end{equation}
a convex combination of $\mathbf{R}$ and $\mathbf{I}_p$.

The estimator (\ref{SMEQ:MINarg}) has certain appreciable properties.
First, it is on the correlation scale (i.e., its diagonal elements are always unity).
Second, whenever the penalty parameter is strictly positive the resulting estimate is positive definite (p.d.).
It can also be seen that the right-hand and left-hand limits are the sample correlation matrix and the identity matrix, respectively.
This latter observation enables the understanding of the estimator: It balances the unbiased but highly variable matrix $\mathbf{R}$ with the biased but perfectly stable identity matrix.
We note that \citet{Warton08_SM} arrives at the same correlation matrix estimator via a somewhat different road.

\subsubsection{Choosing the penalty parameter}
\label{SMSSEC:CPP}
From the above we get that $\mathbf{R}(\vartheta)$ is always p.d.\ when the penalty parameter is strictly positive, irrespective of the ratio $p/(n - 1)$.
It is important to make a well-informed choice regarding its value.
Choosing it too small can lead to an ill-conditioned estimate when $p \gtrapprox n - 1$, meaning that the estimate is unstable and that its use in matrix operations may lead to large error propagation \citep{PeetersCNplot_SM}.
Choosing it too large, however, may suppress relevant data signal.
Here, we choose a data-driven approach that seeks loss efficiency.
That is, we will use K-fold cross-validation ($k$CV) of the log-likelihood function which, asymptotically, can be explained in terms of minimizing Kullback-Leibler divergence.
The $k$CV score for $\mathbf{R}(\vartheta)$ based on a fixed choice of $\vartheta$ can be stated as:
\begin{equation}\label{SMEQ:kCV}
    \varphi^{K}(\vartheta) := \frac{1}{K}\sum_{k = 1}^{K}n_k \left\{\ln|\mathbf{R}(\vartheta)_{\neg k}| + \tr\left[\mathbf{R}_k\left(\mathbf{R}(\vartheta)_{\neg k}\right)^{-1}\right]\right\},
\end{equation}
where $n_k$ denotes the sample size of subset $k$, for $k = 1, \ldots, K$ disjoint subsets and where $\mathbf{R}_k$ denotes the sample correlation matrix based on subset $k$, while $\mathbf{R}(\vartheta)_{\neg k}$ denotes the regularized correlation matrix based on all samples not in $k$.
Using (\ref{SMEQ:kCV}), we choose an optimal $\vartheta$, denoted $\vartheta^{\ddagger}$ such that:
\begin{equation}\label{SMEQ:kCVchoose}
    \vartheta^{\ddagger} := \argmin_{\vartheta \in (0,1]} \varphi^{K}(\vartheta).
\end{equation}
The conditioning of the resulting estimate $\mathbf{R}(\vartheta^{\ddagger})$ can be evaluated along the lines laid out in \citet{PeetersCNplot_SM}.
The procedure is fast for practically accepted choices for $K$, such as $K \in \{5,10\}$, when realizing that the optimization problem in (\ref{SMEQ:kCVchoose}) can be combined with root-finding procedures such as the Brent algorithm \citep{Brent_SM}.

\subsection{Factor analytic modeling}
\label{SMSSEC:AFA}
The assumptions considered for the factor model are \citep{PeetersThesis_SM,PeetersAEfa_SM}:
(i) $\boldsymbol{\mathrm{z}}_{i}\ci \boldsymbol{\mathrm{z}}_{i'},\forall i\neq i'$; (ii) rank$(\mathbf{\Lambda})=m$; (iii) $\boldsymbol{\epsilon}_{i}\sim \mathcal{N}_{p}(\boldsymbol{0}, \boldsymbol{\Psi})$, with $\boldsymbol{\Psi}\equiv \mbox{diag}[\psi_{11},\ldots, \psi_{pp}]$, and $\psi_{jj}>0,\forall j$; (iv) $\boldsymbol{\xi}_{i}\sim \mathcal{N}_{m}(\boldsymbol{0}, \mathbf{I}_m)$; and (v) $\boldsymbol{\xi}_{i}\ci\boldsymbol{\epsilon}_{i'},\forall i,i'$.
The likelihood for the observations conditional on the realization of $\mathbf{\Xi}$ can then be expressed as:
\begin{equation}\label{SMEQ:likelihood}\nonumber
  L(\mathbf{\Lambda}, \mathbf{\Xi}, \mathbf{\Psi}; \mathbf{Z}) =
  \prod_{i=1}^{n}f(\boldsymbol{\mathrm{z}}_{i}|\mathbf{\Lambda},\boldsymbol{\xi}_{i},\mathbf{\Psi})
  =
  \prod_{i=1}^{n}(2\pi)^{-\frac{p}{2}}|\mathbf{\Psi}|^{-\frac{1}{2}}
  \exp\left\{-\frac{1}{2}\boldsymbol{\epsilon}_{i}^{\mathrm{T}}\mathbf{\Psi}^{-1}\boldsymbol{\epsilon}_{i}\right\},
\end{equation}
where
$\boldsymbol{\epsilon}_{i}=\boldsymbol{\mathrm{z}}_{i}-\mathbf{\Lambda}\boldsymbol{\xi}_{i}$.
The likelihood of the observed data can be obtained by marginalizing over $\boldsymbol{\xi}_{i}$:
\begin{align}\label{likelihoodMarg}\nonumber
  L(\mathbf{\Lambda}, \mathbf{\Psi}; \mathbf{Z})&=
  \prod_{i=1}^{n}\int f(\boldsymbol{\mathrm{z}}_{i}|\mathbf{\Lambda},\boldsymbol{\xi}_{i},\mathbf{\Psi})
  g(\boldsymbol{\xi}_{i}|\mathbf{I}_m)\,\partial\boldsymbol{\xi}_{i}\\\nonumber
  &=\prod_{i=1}^{n}(2\pi)^{-\frac{p}{2}}\left|\mathbf{\Lambda}\mathbf{\Lambda}^{\mathrm{T}}+\mathbf{\Psi}\right|^{-\frac{1}{2}}
  \exp\left\{-\frac{1}{2}\boldsymbol{\mathrm{z}}_{i}^{\mathrm{T}}\left(\mathbf{\Lambda}\mathbf{\Lambda}^{\mathrm{T}}+\mathbf{\Psi}\right)^{-1}
  \boldsymbol{\mathrm{z}}_{i}\right\}\\
  &=(2\pi)^{-\frac{np}{2}}\left|\mathbf{\Lambda}\mathbf{\Lambda}^{\mathrm{T}}+\mathbf{\Psi}\right|^{-\frac{n}{2}}
  \exp\left\{-\frac{n}{2}\tr\left[\mathbf{R}\left(\mathbf{\Lambda}\mathbf{\Lambda}^{\mathrm{T}}+\mathbf{\Psi}\right)^{-1}\right]
  \right\}
  ,
\end{align}
giving that the factor decomposition constrains the correlation
structure of the $\boldsymbol{\mathrm{z}}_{i}$ to be a function of $\mathbf{\Theta} = \{\mathbf{\Lambda},\mathbf{\Psi}\}$, i.e.:
\begin{equation}\label{Fundamental}
    \mathbf{\Sigma}(\mathbf{\Theta})=\mathbf{\Lambda}\mathbf{\Lambda}^{\mathrm{T}}+\mathbf{\Psi}.
\end{equation}
Then, for existence (vi), generally $(p-m)^{2}-p-m\geqslant0$, simply stating that the number of nonredundant elements in the sample correlation matrix $\mathbf{R}$ must be greater than or equal
to the number of freely estimable parameters in $\boldsymbol{\Sigma}$, which places an upper bound on $m$.

Hence, under the assumptions of the factor model $\boldsymbol{\mathrm{z}}_{i} \sim \mathcal{N}_p(\boldsymbol{0}, \mathbf{\Sigma}(\mathbf{\Theta}))$.
Moreover, from (\ref{likelihoodMarg}) we understand that $\mathbf{R}$ is a sufficient statistic.
Thus, we may assume that under the model $\mathbf{R} \sim \mathcal{W}_p\left(n-1,\mathbf{\Sigma}(\mathbf{\Theta})/(n-1)\right)$.
The negative log-likelihood of $\mathbf{R}$ (up to proportionality) under this correlation-structure model is then:
\begin{equation}\label{SMEQ:LLRmod}
    \mathcal{L}(\mathbf{\Sigma}(\mathbf{\Theta}); \mathbf{R}) \propto \ln|\mathbf{\Sigma}(\mathbf{\Theta})| + \tr\Big[\mathbf{R}\mathbf{\Sigma}(\mathbf{\Theta})^{-1}\Big].
\end{equation}
\citet{Joreskog67_SM} then proposed to use the following discrepancy function to find an estimate $\hat{\mathbf{\Theta}} = \{\hat{\mathbf{\Lambda}},\hat{\mathbf{\Psi}}\}$ of $\mathbf{\Theta}$:
\begin{equation}\label{SMEQ:MinFunc}
    F[\mathbf{\Sigma}(\mathbf{\Theta});\mathbf{R}] = \ln|\mathbf{\Sigma}(\mathbf{\Theta})| + \mathrm{tr}\left[{\mathbf{R}}\mathbf{\Sigma}(\mathbf{\Theta})^{-1}\right] - \ln|{\mathbf{R}}| - p.
\end{equation}
Minimizing (\ref{SMEQ:MinFunc}) is equivalent to minimizing (\ref{SMEQ:LLRmod}), but computationally more convenient.
Minimizing (\ref{SMEQ:MinFunc}) does require a p.d.\ and, ideally, well-conditioned sample correlation matrix for the estimate $\hat{\mathbf{\Theta}}$ to be meaningful and stable.
Hence, we propose to replace the ill-behaved ${\mathbf{R}}$ by the well-behaved ${\mathbf{R}}(\vartheta^{\ddagger})$ in (\ref{SMEQ:MinFunc}) in order to obtain estimate $\hat{\mathbf{\Theta}}$:
\begin{equation}\label{SMEQ:MinFuncProb}
    \hat{\mathbf{\Theta}} := \argmin_{\mathbf{\Lambda},\mathbf{\Psi}} F\left[\mathbf{\Sigma}(\mathbf{\Theta});{\mathbf{R}}(\vartheta^{\ddagger})\right].
\end{equation}
The algorithm for minimizing (\ref{SMEQ:MinFuncProb}) based on the ML normal equations $\partial F[\mathbf{\Sigma}(\mathbf{\Theta});{\mathbf{R}}(\vartheta^{\ddagger})]/\partial\mathbf{\Lambda} = \boldsymbol{0}$ and $\partial F[\mathbf{\Sigma}(\mathbf{\Theta});{\mathbf{R}}(\vartheta^{\ddagger})]/\partial\mathbf{\Psi} = \boldsymbol{0}$ is the Fletcher-Powell algorithm \citep{FPalgo63_SM} introduced into FA by \citet{Joreskog67_SM}.
This algorithm is widely used in many standard platforms performing FA.

For model determinacy (see Section 2.2.2.\ of the Main Text), the minimization is performed under the requirement that $\mathbf{\Lambda^{\mathrm{T}}}\mathbf{\Psi}^{-1}\mathbf{\Lambda}$ be diagonal with ordered diagonal elements.
This gives the canonical solution to the minimization problem which is convenient and efficient from an estimation perspective.
For interpretation purposes the resulting estimate $\hat{\mathbf{\Lambda}}$ may be subjected to a post-hoc rotations.
In our case this will be the normalized Varimax rotation \citep{Kaiser58_SM,Horst65_SM,Mulaik2010_SM} giving a rotation to orthogonal simple structure:
\begin{equation}\nonumber
     \mathbf{\Gamma} := \arg\max_{\mathbf{H}}\left\{ \sum_{k = 1}^{\tilde{m}}\sum_{j = 1}^{p^{*}}
     \frac{(\hat{\mathbf{\Lambda}}\mathbf{H})_{jk}^{4}}{\hat{c}_{j}^{2}} -
     \frac{1}{p^{*}} \sum_{k = 1}^{\tilde{m}} \left[ \sum_{j = 1}^{p^{*}}
     \frac{(\hat{\mathbf{\Lambda}}\mathbf{H})_{jk}^{2}}{\hat{c}_{j}} \right]^{2}
     : \mathbf{H}^{\mathrm{T}}\mathbf{H} = \mathbf{I}_{\tilde{m}} \right\},
\end{equation}
where $\hat{c}_{j} = \sum_{k = 1}^{\tilde{m}}(\hat{\mathbf{\Lambda}})_{jk}^{2}$ denotes the retrieved communality: the amount of variance of item $j$ explained by the latent features (also see Section \ref{SMSSEC:DA}).
Maximizing this argument comes down to solving quartic equations and is well-implemented in most standard packages and platforms.
The Varimax-rotated solution is then $\hat{\mathbf{\Lambda}}\mathbf{\Gamma} \equiv \hat{\mathbf{\Lambda}}_{V}$.
In the remaining text $\hat{\mathbf{\Lambda}}$ may refer to the loadings matrix either before or after post-hoc rotation in order to avoid notational clutter.

\subsection{Latent dimensionality selection}
\label{SMSSEC:DS}
The estimation process detailed in the previous section hinges upon a choice for the latent dimensionality $m$.
Here we give some details on the latent dimensionality selection approach used in the Main Text.
Section \ref{SMSSEC:DA} explains the basic decisional approach.
Section \ref{SMSSEC:ADS} details on how this basic approach may be supplemented for additional decision support.

\subsubsection{Decision approach}
\label{SMSSEC:DA}
From (\ref{Fundamental}) we get that the variance of the $j$th feature can be expressed as:
\begin{equation}\nonumber
    \sum_{k = 1}^{m} \lambda_{jk}^{2} + \psi_{jj},
\end{equation}
where $\psi_{jj}$ represents the unique variance and where $\sum_{k = 1}^{m} \lambda_{jk}^{2} \equiv c_j$ represents the \emph{common} variance for feature $j$.
The common variance $c_j$ can be understood as the amount of variance of observed feature $j$ that is explained by the $m$ latent features.
In the literature $c_j$ is known as the \emph{communality}.
For the population correlation case, we have the consequence: $c_j = 1 - \psi_{jj}$.
This forms the basis of our decisional approach.

From a matrix perspective we have that:
\begin{equation}
    \mathbf{\Sigma}(\mathbf{\Theta}) - \mathbf{\Psi} = \mathbf{\Lambda}\mathbf{\Lambda}^{\mathrm{T}}.
\end{equation}
That is, the correlation matrix with communalities in the diagonal (the reduced correlation matrix) is Gramian and of rank $m$ \citep[see, e.g., Chapter 8 of][]{Mulaik2010_SM}.
The number of common factors can then be determined by assessing the rank of $\mathbf{\Sigma}(\mathbf{\Theta}) - \mathbf{\Psi}$ when replacing its constituents with appropriate estimates.
We use $\mathbf{R}(\vartheta^{\ddagger})$ as the sampling counterpart to the population correlation matrix.
We use $\mathbf{I}_p$ as a conservative estimate of the unique variance matrix.\
Hence, we then obtain $0$ as a lower-bound estimate of the communalities, such that all variance is considered unique or error variance.
This information can be used to understand the decisional approach.

Some ready algebra will show that:
\begin{equation}\label{SMEQ:EIGdc}
    \mathbf{R}(\vartheta^{\ddagger}) - \mathbf{I}_p = (1 - \vartheta^{\ddagger})(\mathbf{R} - \mathbf{I}_p),
\end{equation}
giving the matrix we assume Gramian and whose rank we want to determine.
Now, let $\mathbf{V}\mathbf{D}(\mathbf{R})\mathbf{V}^{\mathrm{T}}$ be the spectral decomposition of $\mathbf{R}$ with $\mathbf{D}(\mathbf{R})$ denoting a diagonal matrix with the eigenvalues of $\mathbf{R}$ on the diagonal and where $\mathbf{V}$ denotes the matrix that contains the corresponding eigenvectors as columns.
Note that $\mathbf{V}\mathbf{V}^{\mathrm{T}} = \mathbf{V}^{\mathrm{T}}\mathbf{V} = \mathbf{I}_p$.
Then (\ref{SMEQ:EIGdc}) can be written as
\begin{equation}\label{SMEQ:EIGdc2}
    \mathbf{V}\Big\{(1 - \vartheta^{\ddagger})\left[\mathbf{D}(\mathbf{R}) - \mathbf{I}_p\right]\Big\}\mathbf{V}^{\mathrm{T}},
\end{equation}
giving that the eigenvalues of interest are of the form $(1 - \vartheta^{\ddagger})[d(\mathbf{R})_j - 1]$.
The basic decisional rule is then to choose an optimal value of $m$, say $\tilde{m}$, by:
\begin{equation}\label{SMEQ:DRule}
    \tilde{m} := \mbox{card}(\mathrm{A}), \,\,\,\ \mbox{with}~ \mathrm{A} \equiv \Big\{j:(1 - \vartheta^{\ddagger})[d(\mathbf{R})_j - 1] > 0\Big\}.
\end{equation}
Our communality estimate concurs with the weakest lower-bound estimate \citep{Guttman56_SM}, which corresponds to all variance being unique or error variance.
From this perspective, a positive eigenvalue $(1 - \vartheta^{\ddagger})[d(\mathbf{R})_j - 1]$ indicates a latent factor whose contribution to variance-explanation is above and beyond mere unique variance.
We then retain all such factors.

The decisional rule in (\ref{SMEQ:DRule}) can be understood as a regularized version of Guttman's weakest lower-bound approach to minimum rank determination \citep{Guttman56_SM}.
Hence, in the remainder we will refer to this rule as the Guttman bound (GB).
It also concurs with a regularized version of what is known as the Kaiser-rule \citep{Kaiser70_SM}.
While especially the Guttman approach has historically been used as a lower-bound estimate of the latent dimensionality, we consider the decisional rule in (\ref{SMEQ:DRule}) to give an upper-bound.
Section \ref{SMSEC:SDS} contains an extensive simulation study that shows that, in challenging situations, the rule (\ref{SMEQ:DRule}) provides a reliable upper-bound.
This section also explains why formal testing (such as likelihood ratio testing) is not available in the situations we consider.
In such situations the approach above gives a computationally inexpensive and reliable probe regarding the number of factors that dominate information content.

\subsubsection{Additional decision support}
\label{SMSSEC:ADS}
Here we describe tools and strategies for assessing the quality of the factor solution and the choice of latent dimensionality.
These revolve around the assessment of (i) factorability, (ii) extracted communalities, (iii) proportion of explained variance, and (iv) interpretability of extracted factors.
These assessments may inform if the result of the decisional rule above should be accepted or be treated as an upper-bound.

\emph{Factorability}.
Factorability refers to the assessment if factor analysis is an appropriate tool for the data at hand.
Where appropriateness is taken to mean the ability to identify coherent common latent factors.
The basic premise of common factor analysis, from a conditional independence perspective, is that the observed features are independent given the common latent features, i.e.: $\boldsymbol{\mathrm{z}}_{j}\ci \boldsymbol{\mathrm{z}}_{j'}|\{\boldsymbol{\mathrm{\xi}}_{k}\}_{k = 1}^{m},\forall j\neq j'$.
Guttman \citep{Gutt55_SM} then proved that the inverse of the marginal (population and sample) correlation matrix should also be near diagonal for factor analysis to be appropriate.
Moreover, it necessarily approaches a diagonal matrix when the number of observed features grows to infinity.
Hence, a basic inspection of the off-diagonal elements of $\mathbf{R}(\vartheta^{\ddagger})^{-1}$ would be a basic evaluation of factorability in our case.
One practical option for such an inspection is to assess the Kaiser-Meyer-Olkin (KMO) measure of feature-sampling adequacy \citep{Kaiser70_SM, KaiserRice74_SM}.
For our setting the KMO index amounts to:
\begin{equation}
    \frac{\sum_{j}\sum_{j'}\left[\mathbf{R}(\vartheta^{\ddagger}) - \mathbf{I}_p\right]^{2}_{jj'}}
    {\sum_{j}\sum_{j'}\left[\mathbf{R}(\vartheta^{\ddagger}) - \mathbf{I}_p\right]^{2}_{jj'} + \sum_{j}\sum_{j'}\left[\mathbf{P}(\vartheta^{\ddagger}) - \mathbf{I}_p\right]^{2}_{jj'}},
\end{equation}
where $\mathbf{P}(\vartheta^{\ddagger}) = [\mathbf{R}(\vartheta^{\ddagger})^{-1} \circ \mathbf{I}_p]^{-1/2}\mathbf{R}(\vartheta^{\ddagger})^{-1}[\mathbf{R}(\vartheta^{\ddagger})^{-1} \circ \mathbf{I}_p]^{-1/2}$, the standardized regularized precision matrix.
The index basically compares the sizes of the off-diagonal entries of the regularized correlation matrix to the sizes of the off-diagonal entries of its scaled inverse (proportional to the partial correlation matrix) and takes values in $[0,1]$.
In general the KMO index will be larger when the marginal correlations get larger and the partial correlations get smaller.
It will indeed equal unity when $\mathbf{P}(\vartheta^{\ddagger}) = \mathbf{I}_p$.
In general a KMO index between $.9$ and $1$ is considered to be indicative of marvelous factorability \citep{Kaiser70_SM}.

\emph{Comparison of communalities}.
The GB tends to be very reliable when the communalities of the generating data mechanism are high, irrespective of the obtained sample size (see Section \ref{SMSEC:SDS} below).
Hence, it is of considerable interest to compare lower-bound estimates of the communalities to the extracted communalities.
The strongest lower-bound estimate of the communality for any feature $j$ then is \citep{Guttman56_SM}:
\begin{equation}
    \mathrm{SMC}_{j} = 1 - \frac{1}{\left[\mathbf{R}(\vartheta^{\ddagger})^{-1} \circ \mathbf{I}_p\right]_{jj}},
\end{equation}
which is essentially the squared multiple correlation for predicting feature $j$ from the remaining $p - 1$ features.
It is the best possible lower-bound estimate for the communality \citep{Guttman56_SM}.
When, overall, these are high, the generating data mechanism (from a factor analytic perspective) most likely has high communalities and, hence, the GB has reliable performance.
It is also of interest to juxtapose the $\mathrm{SMC}_{j}$ with the retrieved estimated communalities $\hat{c}_{j} = \sum_{k = 1}^{\tilde{m}}(\hat{\mathbf{\Lambda}})_{jk}^{2} = (\hat{\mathbf{\Lambda}}\hat{\mathbf{\Lambda}}^{\mathrm{T}})_{jj}$.
When the chosen value $\tilde{m}$ is sufficient then one would expect, for almost all $j$, that $\hat{c}_{j} \gtrapprox \mathrm{SMC}_{j}$.
If this is not the case then one might have extracted too few factors.

\emph{Proportion of variance explained}.
We might also look at the proportion of explained variance.
The proportion of variance explained by any factor $k$ can be obtained as:
\begin{equation}\nonumber
    v_k = \frac{(\hat{\mathbf{\Lambda}}^{\mathrm{T}}\hat{\mathbf{\Lambda}})_{kk}}{p}.
\end{equation}
The proportion of variance explained by all factors is then $\sum_{k} v_k$.
We want the proportion of variance explained by all factors to be appreciable (say, in excess of 70\%).
Moreover, one would want the proportion of variance explained by the $\tilde{m}$th factor in relation to the $(\tilde{m} - 1)$th factor to be appreciable and the proportion of variance of the $(\tilde{m} + 1)$th factor in relation to the $\tilde{m}$th factor to be negligible.
These are all qualitative judgements.
It is often informative to graph $\sum_{k} v_k$ against the (possible) number of factors.
The point at which the graph flattens out is indicative of a formative number of latent factors.

\emph{Substantiveness of factors}.
If a factor does not have at least 3 significant loadings (and thus indicators), than this factor is considered weak or redundant \citep{RJbookFANA,Mulaik2010_SM}.
Such weak factors should be considered for removal.
This can be evaluated by thresholding the loadings matrix:
\begin{equation}\nonumber
\hat{\lambda}_{jk}^{\dag} =
\left\{
\begin{array}{ll}
      0,                  & \mathrm{if} ~|\hat{\lambda}_{jk}| \leq \omega \\
      \hat{\lambda}_{jk}, & \mathrm{if} ~|\hat{\lambda}_{jk}| > \omega \\
\end{array}.
\right.
\end{equation}
In practice $\omega$ is often set to $.3$.
Hence, it is important to assess the interpretability (from a meta-radiomic feature perspective) and practical significance of the extracted latent features.
Again, these are qualitative judgments.

\emph{Factor score determinacy}.
The appropriateness of factor analysis and the quality of the factor extraction can also be probed through the squared multiple correlation between the observed features and the latent common factors. In our setting this squared multiple correlation for latent factor $k$ is given as \citep[see, e.g., p 375 of][]{Mulaik2010_SM}:
\begin{equation}\nonumber
    \mathrm{SMC}_{k}^{\xi}  = \left(\hat{\mathbf{\Lambda}}^{\mathrm{T}}\mathbf{R}(\vartheta^{\ddagger})^{-1}\hat{\mathbf{\Lambda}}\right)_{kk}.
\end{equation}
It indicates how well common factor $k$ can be predicted by the observed features or, from another perspective, to what extent the factor scores (Section \ref{SMSSEC:FS}) are determinate.
The closer $\mathrm{SMC}_{k}^{\xi}$ is to unity, the more determinate the factor scores for latent factor $k$ can be considered to be.
In practice a $\mathrm{SMC}_{k}^{\xi} \geq .9$ would be considered (very) adequate.
Observing $\mathrm{SMC}_{k}^{\xi}$ that are relatively low might indicate that one has retained (a) weak latent factor(s).
We will turn to estimating factor scores next.

\subsection{Factor scores}
\label{SMSSEC:FS}
We desire an estimate of the score each individual would obtain on each of the latent factors, i.e., we desire factor scores.
These will be obtained through the expectation of the conditional distribution $\boldsymbol{\xi}_i|\boldsymbol{\mathrm{z}}_i$.
First, we use standard covariance algebra to obtain the joint distribution of $\boldsymbol{\mathrm{z}}_i$ and $\boldsymbol{\xi}_i$:
\begin{equation}\nonumber
    \begin{bmatrix}
    \boldsymbol{\mathrm{z}}_i\\
    \boldsymbol{\xi}_i
    \end{bmatrix}
    \sim \mathcal{N}_{p + m}\left(\boldsymbol{0}, \mathbf{\Sigma}(\mathbf{\Theta})_{\xi}^{z}\right),
\end{equation}
with
\begin{equation}\nonumber
    \mathbf{\Sigma}(\mathbf{\Theta})_{\xi}^{z}
    =
    \begin{bmatrix}
    \mathbf{\Sigma}(\mathbf{\Theta})_{zz} & \mathbf{\Sigma}(\mathbf{\Theta})_{z\xi} \\
    \mathbf{\Sigma}(\mathbf{\Theta})_{\xi z}                                 & \mathbf{\Sigma}(\mathbf{\Theta})_{\xi\xi}
    \end{bmatrix}
    =
    \begin{bmatrix}
    \mathbf{\Lambda}\mathbf{\Lambda}^{\mathrm{T}} + \mathbf{\Psi} & \mathbf{\Lambda} \\
    \mathbf{\Lambda}^{\mathrm{T}}                                 & \mathbf{I}_m
    \end{bmatrix}.
\end{equation}
Here we indeed recognize the marginal distribution of the observed data as obtained via an alternative route in Section \ref{SMSSEC:AFA}.
As the joint distribution is normal, a standard result states that the conditional expectation $\mathbb{E}(\boldsymbol{\xi}_i|\boldsymbol{\mathrm{z}}_i)$ can be obtained as $\mathbf{\Sigma}(\mathbf{\Theta})_{\xi z}\mathbf{\Sigma}(\mathbf{\Theta})_{zz}^{-1}\boldsymbol{\mathrm{z}}_i$ \citep[see, e.g., Theorem 2.5.1 in][]{AndersonBible_SM}.
The matrix $\mathbf{\Sigma}(\mathbf{\Theta})_{\xi z}\mathbf{\Sigma}(\mathbf{\Theta})_{zz}^{-1}$ is the matrix of (OLS) regression coefficients of $\boldsymbol{\xi}_i$ on $\boldsymbol{\mathrm{z}}_i$ and $\mathbf{\Sigma}(\mathbf{\Theta})_{\xi z}\mathbf{\Sigma}(\mathbf{\Theta})_{zz}^{-1}\boldsymbol{\mathrm{z}}_i$ is the regression function.
Hence, scores obtained through this route can be thought of as regression-type scores.
Now,
\begin{align}\label{SMEQ:FacRegressor}\nonumber
    \mathbf{\Sigma}(\mathbf{\Theta})_{\xi z}\mathbf{\Sigma}(\mathbf{\Theta})_{zz}^{-1}
    &= \mathbf{\Lambda}^{\mathrm{T}}\left(\mathbf{\Lambda}\mathbf{\Lambda}^{\mathrm{T}} + \mathbf{\Psi}\right)^{-1}\\\nonumber
    &= \left[\mathbf{I}_m - \mathbf{\Lambda}^{\mathrm{T}}\mathbf{\Psi}^{-1}\mathbf{\Lambda}\left(\mathbf{I}_m + \mathbf{\Lambda}^{\mathrm{T}}\mathbf{\Psi}^{-1}\mathbf{\Lambda}\right)^{-1}\right]\mathbf{\Lambda}^{\mathrm{T}}\mathbf{\Psi}^{-1}\\
    &= \left(\mathbf{I}_m + \mathbf{\Lambda}^{\mathrm{T}}\mathbf{\Psi}^{-1}\mathbf{\Lambda}\right)^{-1}\mathbf{\Lambda}^{\mathrm{T}}\mathbf{\Psi}^{-1},
\end{align}
by the Woodbury matrix identity \citep{Woodbury_SM}.
Hence, $\mathbb{E}(\boldsymbol{\xi}_i|\boldsymbol{\mathrm{z}}_i) = \left(\mathbf{I}_m + \mathbf{\Lambda}^{\mathrm{T}}\mathbf{\Psi}^{-1}\mathbf{\Lambda}\right)^{-1}\mathbf{\Lambda}^{\mathrm{T}}\mathbf{\Psi}^{-1}\boldsymbol{\mathrm{z}}_i$ such that an estimate of obtained score $\boldsymbol{\xi}_i$ given an observed $\boldsymbol{\mathrm{z}}_i$ amounts to:
\begin{equation}\label{SMEQ:facScores}
    \hat{\boldsymbol{\xi}}_i = \left(\mathbf{I}_m + \hat{\mathbf{\Lambda}}^{\mathrm{T}}\hat{\mathbf{\Psi}}^{-1}\hat{\mathbf{\Lambda}}\right)^{-1}\hat{\mathbf{\Lambda}}^{\mathrm{T}}\hat{\mathbf{\Psi}}^{-1}\boldsymbol{\mathrm{z}}_i,
\end{equation}
where $\hat{\mathbf{\Lambda}}$ and $\hat{\mathbf{\Psi}}$ are (possibly rotated) estimates of $\mathbf{\Lambda}$ and $\mathbf{\Psi}$.
To collect all $i = 1, \ldots, n$ estimated scores from (\ref{SMEQ:facScores}) in the $(n \times m)$-dimensional matrix $\hat{\mathbf{\Xi}}$ we use the observed data matrix $\mathbf{Z}$ and the (properties of the) matrix transpose to arrive at the expression given in Section 2.2.3 of the Main Text.

The scores obtained through (\ref{SMEQ:facScores}) are known as Thomson scores.
\citet{Thomson_SM} obtained factor scores by minimizing $\mathbb{E}(\mathbf{B}\boldsymbol{\mathrm{z}}_i - \boldsymbol{\xi}_i)^{2}$.
This minimization leads to $\mathbf{B}$ equalling (\ref{SMEQ:FacRegressor}).
Hence, this estimator minimizes the mean squared prediction error.
There exist alternative consistent estimators of the factor scores.
\citet{BartlettScores_SM} devised an unbiased estimator that, in practice, is very close to Thomson scores.
Under the orthogonal model the latent factors are orthogonal in the population and, hence, the Thomson and Bartlett-type factor scores will be near orthogonal in the sample.
\citet{AndersonRubinClassic_SM} constructed an alternative estimator for the factor scores that enforces their orthogonality in the sample.
This estimator considers the $\boldsymbol{\xi}_i$ to be fixed rather than random.
For an overview of factor score estimation we confine by referring to \citep{AndersonRubinClassic_SM}.

\section{Model evaluation}
\label{SMSEC:ModEval}
In this section we provide details on the measures of model performance used in the Main Text.
For model evaluation we focus on prediction error through the time-dependent Brier score \citep{Brier_SM,BrierModern_SM}.
Let $\tilde{\mathrm{y}}_i(t) = \mathrm{I}\{\tilde{T}_{i} \geq t\}$ denote the observed survival status of subject $i$ at time $t$.
The empirical Brier score may then be seen as a mean square error of prediction when $\hat{\pi}^{a}(t|\boldsymbol{\mathrm{p}}_{i}) \in [0,1]$ is taken to be a prediction of the survival status $\tilde{\mathrm{y}}_i(t) \in \{0,1\}$ \citep{BrierModern_SM}.
Hence, it is a measure of inaccuracy.
Section \ref{SMSSEC:PredErr} provides details on the Brier-based evaluation of prediction error for both our internal and external validation settings.
Section \ref{SMSSEC:R2} explain how the Brier score may be used to calculate a measure of explained variation.

\subsection{Prediction error}
\label{SMSSEC:PredErr}
Let $D$ be a data (subset) indicator.
And say we have a training set $D_T$ and validation set $D_V$ available (data setting 1) of respective sample sizes $n_T$ and $n_V$.
In this case the estimated Brier score is \citep{BrierModern_SM}:
\begin{equation}\label{EQ:BrierValidate}
    \mathcal{B}_{V}(t, \hat{\pi}^{a}) = \frac{1}{n_{V}}\sum_{i \in D_{V}}\hat{W}_{i}(t)\big[\tilde{\mathrm{y}}_i(t) - \hat{\pi}_{T}^{a}(t|\boldsymbol{\mathrm{p}}_{i})\big]^{2},
\end{equation}
where $\hat{\pi}_{T}^{a}$ denotes the survival function under model $a$ based on training data $D_T$ and where
\begin{equation}\label{EQ:IPCW}
    \hat{W}_{i}(t) = \frac{\left[1-\tilde{\mathrm{y}}_i(t)\right]\delta_{i}}{\hat{G}(\tilde{T}_{i})} + \frac{\tilde{\mathrm{y}}_i(t)}{\hat{G}(t)},
\end{equation}
denotes an inverse probability of censoring weighting scheme to cope with information loss due to censoring.
The $\hat{G}$ in (\ref{EQ:IPCW}) denotes the marginal Kaplan-Meier (KM) estimate of the censoring distribution.

In case only one data set is available (as in data setting 2) we have to harness ourselves form overoptimistic prediction errors.
In such situations we will use a cross-validated prediction error.
Let the available data be randomly split into $K$ roughly equally sized subsets $k = 1, \ldots, K$.
The prediction rule $\hat{\pi}^{a}$ is then trained on $D\backslash D_{k}$ ($\hat{\pi}^{a}_{k}$) and the data in $D_{k}$ are used as an internal validation set.
The cross-validated prediction error is then \citep{PEcurves_SM}:
\begin{equation}\label{EQ:CVbrier}
    \mathcal{B}_{CV}(t, \hat{\pi}^{a}) = \frac{1}{K}\sum_{k = 1}^{K}\frac{1}{n_{D_{k}}}\sum_{i \in D_k} \hat{W}_{i}(t)\big[\tilde{\mathrm{y}}_i(t) - \hat{\pi}_{k}^{a}(t|\boldsymbol{\mathrm{p}}_{i})\big]^{2}.
\end{equation}
Note that (\ref{EQ:CVbrier}) is dependent on the choice of folds.
We mitigate this dependency by repeating the $K$-fold cross-validation ($K$-CV) many times and averaging the corresponding Brier scores over these repeats:
\begin{equation}\label{EQ:CVbrierAveraged}
    \tilde{\mathcal{B}}_{CV}(t, \hat{\pi}^{a}) = \frac{1}{B}\sum_{b=1}^{B} \mathcal{B}_{CV}(t, \hat{\pi}^{a})_{b},
\end{equation}
where $\mathcal{B}_{CV}(t, \hat{\pi}^{a})_{b}$ represents the score (\ref{EQ:CVbrier}) under random $K$-foldage $b$.
In our applications we set $K = 5$ and $B = 500$.
Also, we make use of the facilities provided by the \texttt{pec} package \citep{pec_SM}, ensuring that all models under consideration are evaluated using the exact same folds.

Now, (\ref{EQ:CVbrierAveraged}) can be integrated for $t \in [0, \tau]$, with $\tau \leq \max(T_{i})$, to arrive at the overall prediction error over the chosen time-period \citep{PEcurves_SM}:
\begin{equation*}
    \mathcal{B}_{a|CV}^{I} \equiv \frac{1}{\tau}\int_{0}^{\tau} \tilde{\mathcal{B}}_{CV}(t, \hat{\pi}^{a}) \,\partial t.
\end{equation*}
We take $\tau$ to be the median follow-up time \citep{BrierModern_SM}.
Note that (\ref{EQ:BrierValidate}) can be integrated similarly to arrive at $\mathcal{B}_{a|V}^{I}$

\subsection{Explained residual variation}
\label{SMSSEC:R2}
Let the survival probability under the KM null-model be denoted by $\hat{\pi}^{0}(t|\boldsymbol{\mathrm{p}}_{i})$.
Also, let $\tilde{\mathcal{B}}_{CV}(t, \hat{\pi}^{0})$ be the averaged cross-validated prediction error under this null model.
Then the additional explained residual variation for model $a$ at time $t$ vis-\`{a}-vis the KM reference model is \citep{BrierModern_SM}:
\begin{equation*}
    R^{2}_{a|CV}(t) = 1 - \frac{\tilde{\mathcal{B}}_{CV}(t, \hat{\pi}^{a})}{\tilde{\mathcal{B}}_{CV}(t, \hat{\pi}^{0})}.
\end{equation*}
To arrive at the overall explained residual variation we could use the integrated Brier scores $\mathcal{B}_{a|CV}^{I}$ and $\mathcal{B}_{0|CV}^{I}$.
Note that analogous considerations can be given to the (integrated) Brier scores in the validation setting.

\section{Simulations dimensionality selection}
\label{SMSEC:SDS}
In this section we describe the numerical evaluation of the latent dimensionality selection procedure outlined in Section \ref{SMSSEC:DS}.
Section \ref{SMSSEC:SimSet} contains an overview of the general setup of the simulation study.
Section \ref{SMSSEC:SimComp} described the other methods of dimensionality selection to which the GB will be compared.
The results can then be found in Section \ref{SMSSSEC:SimRes}.

\subsection{Setup}
\label{SMSSEC:SimSet}
We seek to evaluate and compare the performance of the basic decision rule XXX in settings that emulate the complexities of radiomic data.
In particular we take interest in varying (i) the $p/n$ ratio, (ii) the true value for $m$, (iii) the communality-strength, and (iv) the factor-structure complexity.

\begin{figure}[h!]
\centering
  \includegraphics[width=\textwidth]{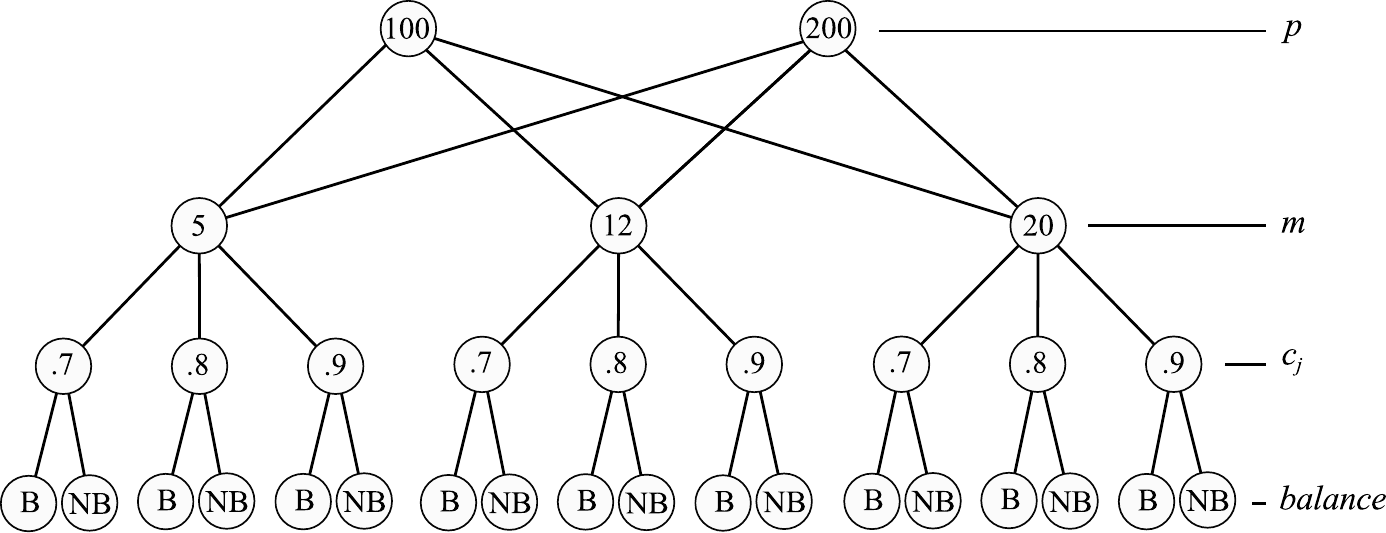}
    \caption{A schematic depiction of the simulation setup.
    Each path from one of the top nodes to a bottom node represents a combination of $p$, $m$, communality-value, and balancedness of structure.
    For each of the possible combinations $100$ data sets are simulated for each of the sample sizes $n \in \{50, 75, 100, 150, 250\}$.
    We then evaluate, for each of the considered dimensionality selection approaches, and for each combination of setup and sample size, the values of $m$ that are deemed optimal.
    In this figure B stands for `balanced' while NB designates `non-balanced'.}
  \label{FIG:SIMSET}
\end{figure}

We will consider $p \in \{100,200\}$.
For each of these feature-dimensions we will consider a true generating $m \in \{5,12,20\}$.
For each of these possible combinations we consider various values for the generating communalities.
As stated previously, the communality $c_j$ for feature $j$ can be understood as the amount of variance of observed feature $j$ explained by the $m$ latent features.
We will consider $c_j \in \{.7,.8,.9\}$, indicating moderate to high communality values.
We will do so by letting each observed feature be an indicator feature for a single latent feature by setting its corresponding loading to $.6$, indicating a moderately high loading.
The remaining $m - 1$ loadings for that feature are then determined as
\begin{equation}\nonumber
    \lambda_m \equiv \sqrt{\frac{c_j - .6^{2}}{m - 1}},
\end{equation}
such that $c_j = .6 + (m-1)\lambda_m^{2}$.
Hence, we are explicitly not considering factor structures that are factorially pure.
For each combination of $p$, $m$, and $c_j$ we will consider both balanced and unbalanced factor structures.
A balanced factor structure is considered to be a structure where each latent feature has (approximately) the same number of indicator features.
An unbalanced factor structure is considered to be a structure where this is not the case.
In the unbalanced setting where $p = 100$ we use the following numbers of indicator features:
\begin{align*}
    & [40 ~20 ~15 ~15 ~10], \\
    & [20 ~10 ~10 ~10 ~10 ~10 ~5 ~5 ~5 ~5 ~5 ~5], \\
    & [10 ~10 ~5 ~5 ~5 ~5 ~5 ~5 ~5 ~5 ~5 ~5 ~5 ~5 ~5 ~3 ~3 ~3 ~3 ~3],
\end{align*}
for the true $m = 5$, $m = 12$, and $m = 20$ solutions, respectively.
For the unbalanced $p = 200$ setting these vectors are multiplied by $2$.
With these settings we can generate for each possible combination a population loadings matrix, say $\mathbf{\Lambda}^{\star}$.
We can then, according to the model, determine the population uniqueness matrix as $\mathbf{\Psi}^{\star} \equiv \mathbf{I}_p - \mathbf{\Lambda}^{\star}(\mathbf{\Lambda}^{\star})^{\mathrm{T}} \circ \mathbf{I}_p$.
For each possible combination of $p$, $m$, $c_j$, and balancedness we then generate $100$ data sets according to $\mathcal{N}_p(\boldsymbol{0}, \mathbf{\Lambda}^{\star}(\mathbf{\Lambda}^{\star})^{\mathrm{T}} + \mathbf{\Psi}^{\star})$ for each of the sample sizes $n \in \{50, 75, 100, 150, 250\}$.
For each individual data set we then determine, after standardization, $\mathbf{R}(\vartheta^{\ddagger})$ based on 5-fold CV (Section \ref{SMSSEC:CPP}).
For each individual $\mathbf{R}(\vartheta^{\ddagger})$ we record, for each considered approach (see Section \ref{SMSSEC:SimComp} below), the value of $m$ deemed or selected as optimal in accordance with the decision rule.
Figure \ref{FIG:SIMSET} contains a schematic depiction of the simulation setup.

\subsection{Comparison}
\label{SMSSEC:SimComp}
The performance of the GB will be compared to testing and model selection strategies that, in the literature, are often applied to the problem of selecting $m$.
In particular, it will be compared to likelihood ratio testing and various information criteria.
These are described below.

\subsubsection{Likelihood ratio test}
\label{SMSSSEC:LRT}
The most formal approach to factor analytic dimensionality assessment is through likelihood ratio (LR) testing.
Let $\hat{\mathbf{\Theta}}_{m} = \{\hat{\mathbf{\Lambda}},\hat{\mathbf{\Psi}}\}_{m}$ denote the ML estimate of the parameters under $m$ factors and let $\mathbf{\Theta}_{m}$ denote its population counterpart.
Testing an $m$-factor model against the saturated model then amounts to assessing:
\begin{align*}
    \mathrm{H}_0:& ~\mathbf{\Sigma} = \mathbf{\Sigma}(\mathbf{\Theta}_m)\\
    \mathrm{H}_a:& ~\mathbf{\Sigma} \succ 0.
\end{align*}
The corresponding LR criterion then converges, under the standard correlation matrix and corresponding parameter estimates under $m$-factors, to $(n - 1)$ times the discrepancy function evaluated at the ML-parameters, and is approximately $\chi^{2}$-distributed under certain regularity conditions \citep{AA_LRT_SM}.
That is,
\begin{equation}\label{SMEQ:LRT}
    \mathcal{T}_m \equiv (n-1)F[\mathbf{\Sigma}(\hat{\mathbf{\Theta}}_{m});\mathbf{R}] \xrightarrow[]{~~d~~} \chi^{2}_{\kappa_{m}},
\end{equation}
with $\kappa_{m} = \left[(p - m)^{2} - (p + m)\right]/2$, the number of parameters in the saturated model (i.e., the unstructured sample correlation) minus the number of freely estimable parameters in the $m$-factor model.
Note that our interest lies with $\mathrm{H}_0$.
The suggested strategy would then be to sequentially test solutions of increasing dimensionality $m = 1, \ldots, m_{\mathrm{max}} \leq \lfloor[2p + 1 - (8p + 1)^{1/2}]/2\rfloor$ until $\mathrm{H}_0$ is \emph{not} rejected at Type-I error level $\alpha$, i.e., until $\mathcal{T}_m < \chi^{2}_{\kappa_{m}}(\alpha)$.
This sequential testing approach is intended to avoid overfactoring (retaining too many factors) due to irregularities in the parameter space \citep[Chapter 2 of][]{PeetersThesis_SM, HBY07_SM}.

Note that (\ref{SMEQ:LRT}) represents an asymptotic result.
In our setting, however, $n$ is usually small relative to $p$.
Moreover, $\mathbf{R}$ might even be unstable when $p > n - 1$ due to multicollinearity.
As $\mathbf{R}(\vartheta)$ is an asymptotically unbiased estimator of the population correlation matrix $\mathbf{R}(\vartheta)$ and $\mathbf{R}$ will converge in the sample limit.
Hence, it is tempting to evaluate the quantity $\mathcal{T}_m$ at $\mathbf{R}(\vartheta)$ and corresponding estimate $\hat{\mathbf{\Theta}}_{m}$.
Although $\kappa_{m}$ will in a sense overestimate the degrees of freedom when $p > n - 1$.

There might be several ways to cope with the last observation.
One might be to scale the likelihood ratio test statistic under $\mathbf{R}(\vartheta)$ such that it is more in compliance with $\chi^{2}_{\kappa_{m}}$.
This road was explored by \citet{Yuan08SEMNS_SM}.
Their method, however, implicates duplication matrices and Kronecker products which pose serious computational bottlenecks for large $p$.
Another simple option would be to adapt the degrees of freedom to incorporate the rank deficiency of $\mathbf{R}$.
One could do this by setting
\begin{equation}\nonumber
   \kappa_{m}' = p'(p' + 1)/2 - [pm + p - m(m - 1)/2],
\end{equation}
where $p' = \mathrm{rank}(\mathbf{R})$.
In general, when $p \geq n$ then $p' = \mathrm{rank}(\mathbf{R}) = n - 1$.
Some ready algebra on $\kappa_{m}' > 0$ under this situation will give that:
\begin{equation}\nonumber
    m_{\mathrm{max}} \leq \left\lfloor-\frac{1}{2}\left[\sqrt{-4n^{2} + 4n + 4p^{2} + 12p + 1} - 2p - 1\right]\right\rfloor,
\end{equation}
which puts serious constraints on the maximum value for $m$ and the $p$ to $n$ ratio.
It is then very well possible that the true factor solution is not represented in the feasible domain for $m$.

Hence, we will refer $\mathcal{T}_{m,\vartheta} \equiv (n-1)F[\mathbf{\Sigma}(\hat{\mathbf{\Theta}}_{m});\mathbf{R}(\vartheta)]$ to $\chi^{2}_{\kappa_{m}}$ in a sequential testing approach as outlined above.
So, while we are aware that we cannot use $\mathcal{T}_{m,\vartheta} \sim \chi^{2}_{\kappa_{m}}$ for inference in general, we are interested to assess its performance in the combinations indicated in Section \ref{SMSSEC:SimSet}.

\subsubsection{Information criteria}
\label{SMSSSEC:ICs}
Information criteria are often used in selecting a value for $m$.
These are model selection criteria that try to balance model fit with model complexity.
Their general setup evaluates (minus 2 times) the maximized value of the (model-dependent) likelihood function offset with a penalty function dependent on the free parameters in the model.
We are interested in evaluating the Akaike Information Criterion \citep[AIC;][]{Aik73_SM} and Bayesian Information Criterion \citep[BIC;][]{BIC_SM}, as these belong to the most oft used criteria \citep[see, e.g.,][]{PreachChoose_SM}.
Now, let $\mathbf{\Sigma}(\hat{\mathbf{\Theta}}_{m}) = \hat{\mathbf{\Lambda}}_{m}\hat{\mathbf{\Lambda}}_{m}^{\mathrm{T}} + \hat{\mathbf{\Psi}}_{m}$ denote the $\mathbf{R}(\vartheta^{\ddagger})$-based MLE solution under $m$ factors and let $\eta = p(m + 1) - m(m - 1)/2$ indicate the number of free parameters in the model.
Then, for our problem at hand, the general IC amounts to:
\begin{equation}\nonumber
    n \left\{ p \ln(2\pi) + \ln|\mathbf{\Sigma}(\hat{\mathbf{\Theta}}_{m})| + \mbox{tr}\left( \mathbf{\Sigma}(\hat{\mathbf{\Theta}}_{m})^{-1}\mathbf{R}(\vartheta^{\ddagger}) \right) \right\} + \varphi,
\end{equation}
where the first term equals $-2\ln[L(\hat{\mathbf{\Lambda}}_{m}, \hat{\mathbf{\Psi}}_{m}; \mathbf{Z})]$ (see equation (\ref{likelihoodMarg})) and where $\varphi$ denotes the penalty.
For the AIC $\varphi = 2\eta$.
For the BIC $\varphi = \ln(n)\eta$.
The strategy would then be to determine the AIC and BIC for a range of consecutive values of $m$.
The solution with the lowest IC score is deemed optimal.

In general ICs are found to overfactor \citep[see, e.g.,][]{LW04_SM}.
This behavior, however, is observed mostly in (numerical studies with) small scale models with relatively large $n$.
Here, we are interested to evaluate their performance in larger scale models with more challenging $p$ to $n$ ratios, such as outlined in Section \ref{SMSSEC:SimSet}.

\subsection{Results}
\label{SMSSSEC:SimRes}
The results can be found in Tables \ref{TABLE:p100m5balanced}--\ref{TABLE:p200m20unbalanced}.
As a reading example, consider Table \ref{TABLE:p100m5balanced}, indicating results for the balanced factor structure situation where $p = 100$ and where the true $m = 5$.
The upper third gives the results when $c_j = .7$, the middle third represents $c_j = .8$ and, lastly, the lower third $c_j = .9$.
We then see that, at $c_j = .7$ and for a sample size of $n = 150$, the GB selects the 5-factor solution 94 times, the 6-factor solution 5 times, and, finally, the 7-factor solution 1 time.
All tables can be read in a similar manner.

\subsubsection{True $m = 5$}
The results for the true $m = 5$ runs can be found in Tables \ref{TABLE:p100m5balanced}--\ref{TABLE:p200m5disbalanced}.
When $p = 100$ and the structure is balanced (Table \ref{TABLE:p100m5balanced}) the AIC performs the best overall, in the sense that it selects the 5-factor solution most often.
The BIC and GB are more on a par in terms of selecting the true factor-dimension.
The AIC and BIC tend to underfactor (select too few factors) when not selecting the true factor-dimension.
The GB tends to (slightly) overfactor when not selecting the true factor-dimension.
The AIC, BIC and GB all give perfect performance, regardless of sample size, when the communality is high.
The LRT consistently underfactors unless both the communality is medium to high and the sample size is (relatively) large.
This picture persists when the factor-structure is unbalanced (see Table \ref{TABLE:p100m5disbalanced}).
This picture also largely persists when $p$ grows to $200$ (Tables \ref{TABLE:p200m5balanced} and \ref{TABLE:p200m5disbalanced}).

\subsubsection{True $m = 12$}
The results for the true $m = 12$ runs can be found in Tables \ref{TABLE:p100m12balanced}--\ref{TABLE:p200m12unbalanced}.
When $p = 100$ and the structure is balanced (Table \ref{TABLE:p100m12balanced}) the GB dominates performance at all combinations of communality-value and sample size.
Again, the GB tends to slightly overfactor when not selecting the true factor-dimension.
Indeed, the AIC and BIC again tend to underfactor when not selecting the true factor-dimension, especially for the lower sample sizes when $c_j \in \{.7,.8\}$.
This behavior is more pronounced for the BIC which generally seems to need higher sample sizes for adequate performance.
The LRT consistently underfactors.
It only selects the true factor-dimensionality adequately at the highest communality-value and sample size.
This picture persists when the factor-structure is unbalanced (see Table \ref{TABLE:p100m12unbalanced}) although all approaches except the GB seem to diminish in performance somewhat.
When $p$ grows to $200$ (Tables \ref{TABLE:p200m12balanced} and \ref{TABLE:p200m12unbalanced}) the AIC fares somewhat better than GB when $c_j \in \{.7,.8\}$.
The GB, however, gives perfect performance, regardless of sample size, when $c_j = .9$.
Again, the GB tends to overfactor under low sample size and communality-value combinations, whereas the ICs tend to underfactor in this situation.
The LRT never selects the true factor-dimensionality when $p = 200$ and seems most affected by unbalancedness of the factor structure.

\subsubsection{True $m = 20$}
The results for the true $m = 20$ runs can be found in Tables \ref{TABLE:p100m20balanced}--\ref{TABLE:p200m20unbalanced}.
When $p = 100$ and the structure is balanced (Table \ref{TABLE:p100m20balanced}) the GB dominates performance at all combinations of communality-value and sample size.
It only slightly overfactors and underfactors at the lower sample sizes.
The AIC, BIC, and LRT all grossly underfactor when $c_j \in \{.7,.8\}$ at all sample sizes, although the AIC fares somewhat better at the highest samples sizes.
The AIC and BIC pick up their performance when $c_j = .9$ and when $n \geq p$.
The LRT gives its best performance when $c_j = .9$ at the highest sample size, but it never selects the true factor-dimensionality as optimal.
This picture persists when the factor-structure is unbalanced (see Table \ref{TABLE:p100m20unbalanced})
This picture also largely persists when $p$ grows to $200$ (Tables \ref{TABLE:p200m20balanced} and \ref{TABLE:p200m20unbalanced}).

\subsubsection{Discussion}
The LRT gives, expectedly, the worst performance.
While the regularization approach gives the opportunity to compute the LRT statistic even when $p > n$, at low sample-sizes the expectation of $\mathcal{T}_{m,\vartheta^{\ddagger}}$ tends to be below that of $\chi^{2}_{\kappa_{m}}$; an observation also made by \citet{Yuan08SEMNS_SM}.
The LRT only performs well at the combination of the highest considered sample size and the lowest considered feature size when model complexity (in terms of the generating latent dimensionality $m$) is low.
For example, the LRT performs well when $p = 100$, $c_j = .9$, and $n \in \{150,250\}$ when the true $m = 5$.
When the true $m = 12$ (and $p$ and $c_j$ as above) the LRT only performs adequate when $n = 250$.
And when the true $m = 20$ (and $p$ and $c_j$ as above) the LRT gives inadequate performance at all sample sizes.
We then make an observation that, to our knowledge, has not been made before: As, \emph{ceteris paribus}, the complexity of the generating factor structure increases, the sample size should increase for the LRT to have adequate performance.

The ICs tend to give good performance when the complexity of the generating factor structure is low.
Their performance diminishes for more complex generating factor structures.
An effect that is more pronounced at the lower communality-values and lower sample sizes.
This effect is also more pronounced for the BIC than for the AIC.
This should be no surprise as the penalty-term for the BIC is dependent on $n$ whereas for the AIC, it is not.
ICs have been observed to overfactor when $n$ gets large \citep[see, e.g.,][]{LW04_SM,PreachChoose_SM}.
We observe, for the more challenging $p$ to $n$ ratios considered here, that the ICs tend to underfactor.
This tendency gets more severe when the complexity of the generating factor structure increases.
This tendency is also more pronounced for the BIC in comparison to the AIC.
Hence, the BIC and especially the AIC seem to balance model fit and complexity well at low generating-model complexities.
As this complexity increases, both the generating communality values and the sample size should be (relatively) high for adequate performance.

The GB is most directly dependent on the regularization of the correlation matrix.
It can be shown that $\mathbf{R}(\vartheta^{\ddagger})$ is rotation equivariant, meaning that it operates on the eigenvalues of $\mathbf{R}$ only.
When $p \gtrapprox n$ the empirical eigenvalues of $\mathbf{R}$ will be distorted, in the sense that the largest eigenvalues are too large whereas the smallest are too small.
The estimator $\mathbf{R}(\vartheta^{\ddagger})$ shrinks the eigenvalues towards the central value of unity (1).
The decision rule regarding the GB is directly based on the eigenvalues of $\mathbf{R}(\vartheta^{\ddagger})$.
We notice that, for generating factor structures of lower factor-dimension, the GB tends to overfactor when the communalities are low to medium in combination with lower sample sizes.
At these combinations some eigenvalues are (still) overestimated after shrinkage (driven by large sampling fluctuation).
When the generating factor-structure gets more complex, however, the shrunken eigenvalues seem better spread over the latent dimensions that matter and the GB performs well.
Even when the communalities are low and the sample size is small relative to $p$.
When it doesn't select the true generating $m$ as optimal it has the (general) tendency to slightly overfactor.

In our performance assessment the AIC seems the only contender for the GB.
Our preference lies with the GB for the following reasons:
\begin{enumerate}[i.]
  \item Its considerable better performance when the generating factor-structure is more complex;
  \item Its tendency to (slightly) overfactor in cases in which it does not select the true generating value for $m$, as opposed to the other methods' tendency for underfactoring. Overfactoring leads to far less error in factor loading estimates and loading-structure representation (so central to the computation of the desired factor scores) than underfactoring \citep{FAV92_SM,WTG96_SM,FWMS99_SM};
  \item Its considerable lower computational complexity. Its computation requires little more than a single spectral decomposition, which is of worst case computational complexity $\mathcal{O}(p^{3})$.
\end{enumerate}
In conclusion, we will use the GB as an upper-bound to the problem of selecting a value for $m$.
The methods for additional decision support mentioned in Section \ref{SMSSEC:ADS} may then be combined with the GB to inform a final decision.

\begin{table}[h!]
\caption{Results for the $p = 100$, true $m = 5$, balanced factor structure setup.
Cells represent the count for the number of times a particular value of $m$ is chosen as optimal under a combination of selection procedure, sample size and communality-value.
Blank cells represent a count of 0.}
\label{TABLE:p100m5balanced}
\begin{tiny}
\scalebox{.75}{
\begin{tabular}{cccccccccccc}
\hline
                       &                                & \multicolumn{1}{c}{$m = 1$} & \multicolumn{1}{c}{$m = 2$} & \multicolumn{1}{c}{$m = 3$} & \multicolumn{1}{c}{$m = 4$} & \multicolumn{1}{c}{\cellcolor[HTML]{C0C0C0}$m = 5$} & \multicolumn{1}{c}{$m = 6$} & \multicolumn{1}{c}{$m = 7$} & \multicolumn{1}{c}{$m = 8$} & \multicolumn{1}{c}{$m = 9$} & \multicolumn{1}{c}{$m \geq 10$} \\ \hline
\multicolumn{12}{c}{\cellcolor[HTML]{C0C0C0}Communality = $.7$}                                                                                                                                                                                                                                                                                                      \\ \hline
                       & \multicolumn{1}{l|}{$n = 50$}  &                             &                             &                             &                             &                             &                             &                             &                 1            &                             &               99               \\
                       & \multicolumn{1}{l|}{$n = 75$}  &                             &                             &                             &                             &                             &                             &             2                &             13                &            23                 &           62                   \\
\multicolumn{1}{c}{GB} & \multicolumn{1}{l|}{$n = 100$} &                             &                             &                             &                             &          3                   &           16                  &            37                 &         31                    &         10                    &          3                    \\
\end{tabular}
}
\end{tiny}
\end{table}

\begin{table}[]
\caption{
Results for the $p = 100$, true $m = 5$, unbalanced factor structure setup.
Cells represent the count for the number of times a particular value of $m$ is chosen as optimal under a combination of selection procedure, sample size and communality-value.
Blank cells represent a count of 0.}
\label{TABLE:p100m5disbalanced}
\begin{tiny}
\begin{tabular}{cccccccccccc}
\hline
                       &                                & \multicolumn{1}{c}{$m = 1$} & \multicolumn{1}{c}{$m = 2$} & \multicolumn{1}{c}{$m = 3$} & \multicolumn{1}{c}{$m = 4$} & \multicolumn{1}{c}{\cellcolor[HTML]{C0C0C0}$m = 5$} & \multicolumn{1}{c}{$m = 6$} & \multicolumn{1}{c}{$m = 7$} & \multicolumn{1}{c}{$m = 8$} & \multicolumn{1}{c}{$m = 9$} & \multicolumn{1}{c}{$m \geq 10$} \\ \hline
\multicolumn{12}{c}{\cellcolor[HTML]{C0C0C0}Communality = $.7$}                                                                                                                                                                                                                                                                                                      \\ \hline
                       & \multicolumn{1}{l|}{$n = 50$}  &                             &                             &                             &                             &                             &                             &                             &                             &              1               &                99              \\
                       & \multicolumn{1}{l|}{$n = 75$}  &                             &                             &                             &                             &                             &                             &            3                 &               12              &           24                  &            61                  \\
\multicolumn{1}{c}{GB} & \multicolumn{1}{l|}{$n = 100$} &                             &                             &                             &                             &           1                  &           17                  &                  35           &        33                     &              12               &                  2            \\
\end{tabular}
\end{tiny}
\end{table}

\begin{table}[]
\caption{
Results for the $p = 200$, true $m = 5$, balanced factor structure setup.
Cells represent the count for the number of times a particular value of $m$ is chosen as optimal under a combination of selection procedure, sample size and communality-value.
Blank cells represent a count of 0.}
\label{TABLE:p200m5balanced}
\begin{tiny}
\begin{tabular}{cccccccccccc}
\hline
                       &                                & \multicolumn{1}{c}{$m = 1$} & \multicolumn{1}{c}{$m = 2$} & \multicolumn{1}{c}{$m = 3$} & \multicolumn{1}{c}{$m = 4$} & \multicolumn{1}{c}{\cellcolor[HTML]{C0C0C0}$m = 5$} & \multicolumn{1}{c}{$m = 6$} & \multicolumn{1}{c}{$m = 7$} & \multicolumn{1}{c}{$m = 8$} & \multicolumn{1}{c}{$m = 9$} & \multicolumn{1}{c}{$m \geq 10$} \\ \hline
\multicolumn{12}{c}{\cellcolor[HTML]{C0C0C0}Communality = $.7$}                                                                                                                                                                                                                                                                                                      \\ \hline
                       & \multicolumn{1}{l|}{$n = 50$}  &                             &                             &                             &                             &                             &                             &                             &                             &                             &             100                 \\
                       & \multicolumn{1}{l|}{$n = 75$}  &                             &                             &                             &                             &                             &                             &                             &                             &                             &             100                 \\
\multicolumn{1}{c}{GB} & \multicolumn{1}{l|}{$n = 100$} &                             &                             &                             &                             &                             &                             &                             &                             &                             &            100                  \\
\end{tabular}
\end{tiny}
\end{table}

\begin{table}[]
\caption{
Results for the $p = 200$, true $m = 5$, unbalanced factor structure setup.
Cells represent the count for the number of times a particular value of $m$ is chosen as optimal under a combination of selection procedure, sample size and communality-value.
Blank cells represent a count of 0.}
\label{TABLE:p200m5disbalanced}
\begin{tiny}
\begin{tabular}{cccccccccccc}
\hline
                       &                                & \multicolumn{1}{c}{$m = 1$} & \multicolumn{1}{c}{$m = 2$} & \multicolumn{1}{c}{$m = 3$} & \multicolumn{1}{c}{$m = 4$} & \multicolumn{1}{c}{\cellcolor[HTML]{C0C0C0}$m = 5$} & \multicolumn{1}{c}{$m = 6$} & \multicolumn{1}{c}{$m = 7$} & \multicolumn{1}{c}{$m = 8$} & \multicolumn{1}{c}{$m = 9$} & \multicolumn{1}{c}{$m \geq 10$} \\ \hline
\multicolumn{12}{c}{\cellcolor[HTML]{C0C0C0}Communality = $.7$}                                                                                                                                                                                                                                                                                                      \\ \hline
                       & \multicolumn{1}{l|}{$n = 50$}  &                             &                             &                             &                             &                             &                             &                             &                             &                             &            100                  \\
                       & \multicolumn{1}{l|}{$n = 75$}  &                             &                             &                             &                             &                             &                             &                             &                             &                             &            100                  \\
\multicolumn{1}{c}{GB} & \multicolumn{1}{l|}{$n = 100$} &                             &                             &                             &                             &                             &                             &                             &                             &                             &            100                  \\
\end{tabular}
\end{tiny}
\end{table}

\begin{table}[]
\caption{
Results for the $p = 100$, true $m = 12$, balanced factor structure setup.
Cells represent the count for the number of times a particular value of $m$ is chosen as optimal under a combination of selection procedure, sample size and communality-value.
Blank cells represent a count of 0.}
\label{TABLE:p100m12balanced}
\begin{tiny}
\begin{tabular}{cccccccccccc}
\hline
                       &                                & \multicolumn{1}{c}{$m \leq 8$} & \multicolumn{1}{c}{$m = 9$} & \multicolumn{1}{c}{$m = 10$} & \multicolumn{1}{c}{$m = 11$} & \multicolumn{1}{c}{\cellcolor[HTML]{C0C0C0}$m = 12$} & \multicolumn{1}{c}{$m = 13$} & \multicolumn{1}{c}{$m = 14$} & \multicolumn{1}{c}{$m = 15$} & \multicolumn{1}{c}{$m = 16$} & \multicolumn{1}{c}{$m \geq 17$} \\ \hline
\multicolumn{12}{c}{\cellcolor[HTML]{C0C0C0}Communality = $.7$}                                                                                                                                                                                                                                                                                                      \\ \hline
                       & \multicolumn{1}{l|}{$n = 50$}  &                             &                             &                             &                             &                             &                             &                             &             4                &          20                   &            76                  \\
                       & \multicolumn{1}{l|}{$n = 75$}  &                             &                             &                             &                             &            1                 &            15                 &            39                 &           31                  &          11                   &             3                 \\
\multicolumn{1}{c}{GB} & \multicolumn{1}{l|}{$n = 100$} &                             &                             &                             &                             &            36                 &             48                &             14                &           2                  &                             &                              \\
\end{tabular}
\end{tiny}
\end{table}

\begin{table}[]
\caption{
Results for the $p = 100$, true $m = 12$, unbalanced factor structure setup.
Cells represent the count for the number of times a particular value of $m$ is chosen as optimal under a combination of selection procedure, sample size and communality-value.
Blank cells represent a count of 0.}
\label{TABLE:p100m12unbalanced}
\begin{tiny}
\begin{tabular}{cccccccccccc}
\hline
                       &                                & \multicolumn{1}{c}{$m \leq 8$} & \multicolumn{1}{c}{$m = 9$} & \multicolumn{1}{c}{$m = 10$} & \multicolumn{1}{c}{$m = 11$} & \multicolumn{1}{c}{\cellcolor[HTML]{C0C0C0}$m = 12$} & \multicolumn{1}{c}{$m = 13$} & \multicolumn{1}{c}{$m = 14$} & \multicolumn{1}{c}{$m = 15$} & \multicolumn{1}{c}{$m = 16$} & \multicolumn{1}{c}{$m \geq 17$} \\ \hline
\multicolumn{12}{c}{\cellcolor[HTML]{C0C0C0}Communality = $.7$}                                                                                                                                                                                                                                                                                                      \\ \hline
                       & \multicolumn{1}{l|}{$n = 50$}  &                             &                             &                             &                             &                             &                             &                             &              4               &             24                &             72                 \\
                       & \multicolumn{1}{l|}{$n = 75$}  &                             &                             &                             &                             &           1                  &              9               &             29                &           43                  &           15                  &              3                \\
\multicolumn{1}{c}{GB} & \multicolumn{1}{l|}{$n = 100$} &                             &                             &                             &                             &           30                  &            58                 &             10                &            2                 &                             &                              \\
\end{tabular}
\end{tiny}
\end{table}

\begin{table}[]
\caption{
Results for the $p = 200$, true $m = 12$, balanced factor structure setup.
Cells represent the count for the number of times a particular value of $m$ is chosen as optimal under a combination of selection procedure, sample size and communality-value.
Blank cells represent a count of 0.}
\label{TABLE:p200m12balanced}
\begin{tiny}
\begin{tabular}{cccccccccccc}
\hline
                       &                                & \multicolumn{1}{c}{$m \leq 8$} & \multicolumn{1}{c}{$m = 9$} & \multicolumn{1}{c}{$m = 10$} & \multicolumn{1}{c}{$m = 11$} & \multicolumn{1}{c}{\cellcolor[HTML]{C0C0C0}$m = 12$} & \multicolumn{1}{c}{$m = 13$} & \multicolumn{1}{c}{$m = 14$} & \multicolumn{1}{c}{$m = 15$} & \multicolumn{1}{c}{$m = 16$} & \multicolumn{1}{c}{$m \geq 17$} \\ \hline
\multicolumn{12}{c}{\cellcolor[HTML]{C0C0C0}Communality = $.7$}                                                                                                                                                                                                                                                                                                      \\ \hline
                       & \multicolumn{1}{l|}{$n = 50$}  &                             &                             &                             &                             &                             &                             &                             &                             &                             &             100                 \\
                       & \multicolumn{1}{l|}{$n = 75$}  &                             &                             &                             &                             &                             &                             &                             &                             &                             &              100                \\
\multicolumn{1}{c}{GB} & \multicolumn{1}{l|}{$n = 100$} &                             &                             &                             &                             &                             &                             &                             &                             &                             &          100                    \\
\end{tabular}
\end{tiny}
\end{table}

\begin{table}[]
\caption{
Results for the $p = 200$, true $m = 12$, unbalanced factor structure setup.
Cells represent the count for the number of times a particular value of $m$ is chosen as optimal under a combination of selection procedure, sample size and communality-value.
Blank cells represent a count of 0.}
\label{TABLE:p200m12unbalanced}
\begin{tiny}
\begin{tabular}{cccccccccccc}
\hline
                       &                                & \multicolumn{1}{c}{$m \leq 8$} & \multicolumn{1}{c}{$m = 9$} & \multicolumn{1}{c}{$m = 10$} & \multicolumn{1}{c}{$m = 11$} & \multicolumn{1}{c}{\cellcolor[HTML]{C0C0C0}$m = 12$} & \multicolumn{1}{c}{$m = 13$} & \multicolumn{1}{c}{$m = 14$} & \multicolumn{1}{c}{$m = 15$} & \multicolumn{1}{c}{$m = 16$} & \multicolumn{1}{c}{$m \geq 17$} \\ \hline
\multicolumn{12}{c}{\cellcolor[HTML]{C0C0C0}Communality = $.7$}                                                                                                                                                                                                                                                                                                      \\ \hline
                       & \multicolumn{1}{l|}{$n = 50$}  &                             &                             &                             &                             &                             &                             &                             &                             &                             &              100                \\
                       & \multicolumn{1}{l|}{$n = 75$}  &                             &                             &                             &                             &                             &                             &                             &                             &                             &              100                \\
\multicolumn{1}{c}{GB} & \multicolumn{1}{l|}{$n = 100$} &                             &                             &                             &                             &                             &                             &                             &                             &                             &           100                   \\
\end{tabular}
\end{tiny}
\end{table}


\begin{table}[]
\caption{
Results for the $p = 100$, true $m = 20$, balanced factor structure setup.
Cells represent the count for the number of times a particular value of $m$ is chosen as optimal under a combination of selection procedure, sample size and communality-value.
Blank cells represent a count of 0.}
\label{TABLE:p100m20balanced}
\begin{tiny}
\begin{tabular}{cccccccccccc}
\hline
                       &                                & \multicolumn{1}{c}{$m \leq 16$} & \multicolumn{1}{c}{$m = 17$} & \multicolumn{1}{c}{$m = 18$} & \multicolumn{1}{c}{$m = 19$} & \multicolumn{1}{c}{\cellcolor[HTML]{C0C0C0}$m = 20$} & \multicolumn{1}{c}{$m = 21$} & \multicolumn{1}{c}{$m = 22$} & \multicolumn{1}{c}{$m = 23$} & \multicolumn{1}{c}{$m = 24$} & \multicolumn{1}{c}{$m \geq 25$} \\ \hline
\multicolumn{12}{c}{\cellcolor[HTML]{C0C0C0}Communality = $.7$}                                                                                                                                                                                                                                                                                                      \\ \hline
                       & \multicolumn{1}{l|}{$n = 50$}  &                             &                             &                             &                             &          9                   &           41                  &            39                 &            11                 &                             &                              \\
                       & \multicolumn{1}{l|}{$n = 75$}  &                             &                             &                             &              2               &           52                  &             42                &              4               &                             &                             &                              \\
\multicolumn{1}{c}{GB} & \multicolumn{1}{l|}{$n = 100$} &                             &                             &                             &                             &           92                  &               8              &                             &                             &                             &                              \\
\end{tabular}
\end{tiny}
\end{table}

\begin{table}[]
\caption{
Results for the $p = 100$, true $m = 20$, unbalanced factor structure setup.
Cells represent the count for the number of times a particular value of $m$ is chosen as optimal under a combination of selection procedure, sample size and communality-value.
Blank cells represent a count of 0.}
\label{TABLE:p100m20unbalanced}
\begin{tiny}
\begin{tabular}{cccccccccccc}
\hline
                       &                                & \multicolumn{1}{c}{$m \leq 16$} & \multicolumn{1}{c}{$m = 17$} & \multicolumn{1}{c}{$m = 18$} & \multicolumn{1}{c}{$m = 19$} & \multicolumn{1}{c}{\cellcolor[HTML]{C0C0C0}$m = 20$} & \multicolumn{1}{c}{$m = 21$} & \multicolumn{1}{c}{$m = 22$} & \multicolumn{1}{c}{$m = 23$} & \multicolumn{1}{c}{$m = 24$} & \multicolumn{1}{c}{$m \geq 25$} \\ \hline
\multicolumn{12}{c}{\cellcolor[HTML]{C0C0C0}Communality = $.7$}                                                                                                                                                                                                                                                                                                      \\ \hline
                       & \multicolumn{1}{l|}{$n = 50$}  &                             &                             &                             &                1             &             5                &             45                &            38                 &           11                  &                             &                              \\
                       & \multicolumn{1}{l|}{$n = 75$}  &                             &                             &                             &                3             &             51                &              42               &             4                &                             &                             &                              \\
\multicolumn{1}{c}{GB} & \multicolumn{1}{l|}{$n = 100$} &                             &                             &                             &                             &             90                &           10                  &                             &                             &                             &                              \\
\end{tabular}
\end{tiny}
\end{table}

\begin{table}[]
\caption{
Results for the $p = 200$, true $m = 20$, balanced factor structure setup.
Cells represent the count for the number of times a particular value of $m$ is chosen as optimal under a combination of selection procedure, sample size and communality-value.
Blank cells represent a count of 0.}
\label{TABLE:p200m20balanced}
\begin{tiny}
\begin{tabular}{cccccccccccc}
\hline
                       &                                & \multicolumn{1}{c}{$m \leq 16$} & \multicolumn{1}{c}{$m = 17$} & \multicolumn{1}{c}{$m = 18$} & \multicolumn{1}{c}{$m = 19$} & \multicolumn{1}{c}{\cellcolor[HTML]{C0C0C0}$m = 20$} & \multicolumn{1}{c}{$m = 21$} & \multicolumn{1}{c}{$m = 22$} & \multicolumn{1}{c}{$m = 23$} & \multicolumn{1}{c}{$m = 24$} & \multicolumn{1}{c}{$m \geq 25$} \\ \hline
\multicolumn{12}{c}{\cellcolor[HTML]{C0C0C0}Communality = $.7$}                                                                                                                                                                                                                                                                                                      \\ \hline
                       & \multicolumn{1}{l|}{$n = 50$}  &                             &                             &                             &                             &                             &                             &                             &                             &                             &             100                 \\
                       & \multicolumn{1}{l|}{$n = 75$}  &                             &                             &                             &                             &                             &                             &                             &                             &                             &            100                  \\
\multicolumn{1}{c}{GB} & \multicolumn{1}{l|}{$n = 100$} &                             &                             &                             &                             &                             &                             &                             &                             &                             &            100                  \\
\end{tabular}
\end{tiny}
\end{table}

\begin{table}[]
\caption{
Results for the $p = 200$, true $m = 20$, unbalanced factor structure setup.
Cells represent the count for the number of times a particular value of $m$ is chosen as optimal under a combination of selection procedure, sample size and communality-value.
Blank cells represent a count of 0.}
\label{TABLE:p200m20unbalanced}
\begin{tiny}
\begin{tabular}{cccccccccccc}
\hline
                       &                                & \multicolumn{1}{c}{$m \leq 16$} & \multicolumn{1}{c}{$m = 17$} & \multicolumn{1}{c}{$m = 18$} & \multicolumn{1}{c}{$m = 19$} & \multicolumn{1}{c}{\cellcolor[HTML]{C0C0C0}$m = 20$} & \multicolumn{1}{c}{$m = 21$} & \multicolumn{1}{c}{$m = 22$} & \multicolumn{1}{c}{$m = 23$} & \multicolumn{1}{c}{$m = 24$} & \multicolumn{1}{c}{$m \geq 25$} \\ \hline
\multicolumn{12}{c}{\cellcolor[HTML]{C0C0C0}Communality = $.7$}                                                                                                                                                                                                                                                                                                      \\ \hline
                       & \multicolumn{1}{l|}{$n = 50$}  &                             &                             &                             &                             &                             &                             &                             &                             &                             &             100                 \\
                       & \multicolumn{1}{l|}{$n = 75$}  &                             &                             &                             &                             &                             &                             &                             &                             &                             &             100                 \\
\multicolumn{1}{c}{GB} & \multicolumn{1}{l|}{$n = 100$} &                             &                             &                             &                             &                             &                             &                             &                             &                             &         100                     \\
\end{tabular}
\end{tiny}
\end{table}

\section{Additional information oral squamous cell carcinoma data}
\label{SMSEC:ADIoscc}
In this section we give additional information on the oral squamous cell carcinoma data as used in the external validation setting.
Section \ref{SMSSEC:OIP} contains an overview of data acquisition.
Section \ref{SMSSEC:OFP} contains information on the factor solution obtained in the training projection.
Lastly, Section \ref{SMSSEC:ORES} contains visualizations of the analysis results.

\subsection{Image processing and radiomic feature extraction}
\label{SMSSEC:OIP}
Two independent cohorts of human papillomavirus-negative oropharyngeal squamous cell carcinoma patients were available.
Data on the Amsterdam University medical centers (location VUmc) training cohort was gathered from 2008 to 2012.
Data on the University Medical Center Utrecht validation cohort was gathered from 2010 to 2013.
The oropharyngeal tumors were imaged with axial T1-weighted MRI scans without gadolinium enhancement.
Different MRI vendors and protocols were used in image acquisition.
The images were segmented using VelocityAI 3.1 (Velocity Medical Solutions, Atlanta GA, USA).
Identification of the volumes of interest was performed manually by a senior radiologist.
The delineated regions were exported as Digital Imaging and Communications in Medicine Radiotherapy Structure Sets \citep{DICOM_SM}.

A total of 545 raw radiomic features were extracted from the MRI-images using in-house build software \citep{MRIextract_SM} relating to, e.g., morphology, texture, and intensity.
Texture features were derived from grey-level co-occurrence matrices or from grey-level run-length matrices, calculated from either the $x$-$y$ plane (2D) or volumetrically (3D), either per direction (and then averaged) or after merging over all possible directions (combined).
In addition, images were discretized to bin sizes of 32, 64 and 128.
The texture features were subsequently averaged over the 12 representations (each combination of bin size, dimensionality, and direction).
This left a total of $p = 89$ radiomic \emph{core} features.
Table \ref{SMTAB:FEatMes} gives a listing of all final features.
Additional information on the data acquisition and radiomic features can be found in \citet{Mes19_SM}.
The processed data object is available from the Authors upon reasonable request.

\begin{table}[ht]
\caption{Radiomic features for the oral squamous cell carcinoma images.}
\label{SMTAB:FEatMes}
\scalebox{.65}{
\begin{tabular}{lll}
\toprule
Whole MRI maximum grey level		&  Whole MRI range          			& Whole MRI mean             		\\
Whole MRI median        		&  Whole MRI standard deviation			& Maximum grey level values $> .5$	\\
Mean grey level values $> .5$		&  Median grey level values $> .5$		& Tumor MRI maximum grey level		\\
Tumor MRI minimum grey level		&  Tumor MRI range				& Tumor MRI interquartile range		\\
Tumor MRI mean				&  Tumor MRI median				& Tumor MRI standard deviation		\\
Coefficient of variation		&  Skewness               			& Excess kurtosis			\\
Kurtosis         			&  Integrated intensity    			& Mean absolute deviation of the median	\\
Mean absolute deviation of the mean	&  Median absolute deviation of the median	& Mean Laplacian			\\
Tumor volume          			&  Maximum 3D diameter        			& Surface area				\\
Surface to volume ratio			&  S2S equivolumetric sphere volume ratio	& Radius of an equivolumetric sphere 	\\
Compactness A          			&  Compactness B           			& Spherical disproportion		\\
Sphericity            			&  Asphericity            			& Area under intensity-volume histogram curve	\\
Total energy				&  Variance                 			& Root-mean-square			\\
Mean of the maximum and adjacent voxels	&  Fractal dimension (calculated)		& Fractal dimension (fitted)		\\
Fractal abundance      			&  Fractal lacunarity      			& Moran's I				\\
Geary's C             			&  Uniformity             			& Entropy				\\
Joint maximum          			&  Joint average             			& Joint variance			\\
Joint entropy        			&  Difference average          			& Difference variance			\\
Difference entropy     			&  Sum average               			& Sum variance				\\
Sum entropy          			&  Angular second moment 			& Contrast				\\
Dissimilarity         			&  Inverse difference				& Inverse difference normalized		\\
Inverse difference moment		&  Inverse difference moment normalized		& Inverse variance			\\
Correlation				&  Autocorrelation                		& Cluster tendency			\\
Cluster shade        			&  Cluster prominence        			& Measure of information correlation 1	\\
Measure of information correlation 2	&  Short runs emphasis				& Long runs emphasis			\\
Low grey level run emphasis		&  High grey level run emphasis			& Short run low grey level emphasis	\\
Short run high grey level emphasis	&  Long run low grey level emphasis		& Long run high grey level emphasis	\\
Grey level non-uniformity		&  Grey level non-uniformity normalized		& Run length non-uniformity		\\
Run length non-uniformity normalized	&  Run percentage                     		& Grey level variance			\\
Run length variance                   	&  Run entropy                    		&					\\
\bottomrule
\multicolumn{3}{l}{}\\[-0.75\normalbaselineskip]%
\multicolumn{3}{l}{S2S = Surface area to surface.}%
\end{tabular}
}
\end{table}

\subsection{Factor solution training data}
\label{SMSSEC:OFP}
\begin{figure}[h!]
\centering
  \includegraphics[width=.9\textwidth]{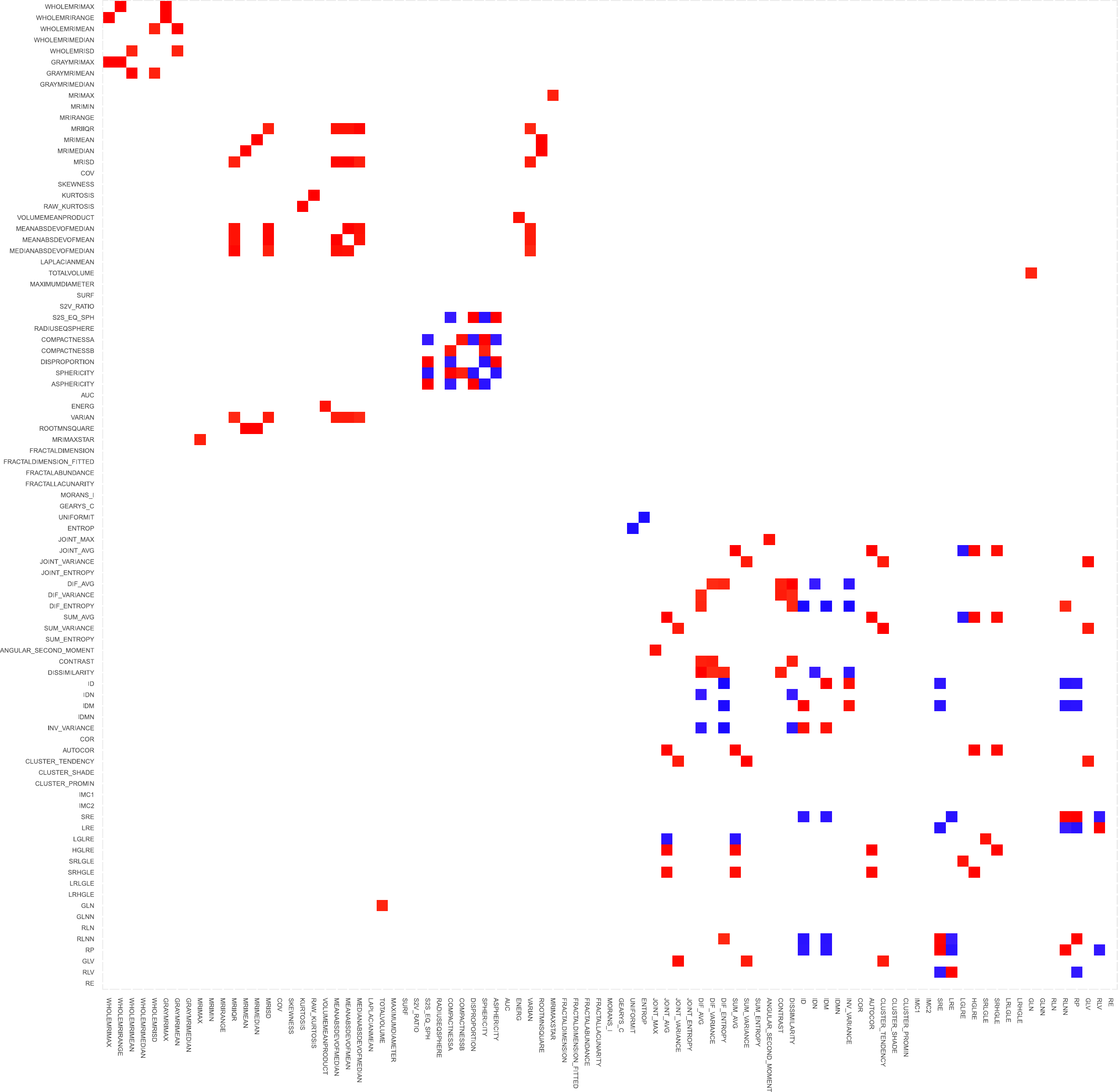}
    \caption{Thresholded correlation heatmap of the radiomic features for the oral squamous cell carcinoma data.
    Absolute correlations $< .95$ were set to zero such that only correlations equalling or exceeding an absolute marginal correlation threshold $\tau$ of $.95$ are visualized.
    Red then indicates a positive correlation $\geq .95$.
    Blue then indicates a negative correlation $\leq -.95$.
    Colored regions then visualize blocks of information redundancy.
    }
  \label{SMFIG:EVDredundant}
\end{figure}

The first step in the factor modeling pipeline is to get a regularized estimate of the correlation matrix on the nonredundant radiomic features.
Figure \ref{SMFIG:EVDredundant} visualizes the redundancy pattern among the full set of 89 radiomic features.
The redundancy filtering algorithm (Algorithm \ref{AlgRF}) was run with $\tau$ set to $.95$ and it retained $p^* = 51$ out of the original $p = 89$ features.
The remaining correlation matrix was then subjected to penalized ML estimation.
The optimal value for the penalty parameter was determined by 5-fold CV of the log-likelihood function and was found to be $.04950291$.
The resulting regularized correlation matrix was well-conditioned, as indicated by a low condition number ($ 292.97$).

The second step consists of factor analytic data compression.
The GB indicates $10$ factors.
In accordance with its purported usage (see Section \ref{SMSEC:SDS}) we treat $10$ as an upper-bound to the optimal dimension of the latent vector.
As the proportion of additional variance explained flattens after 7 common factors and as solutions in excess of 7 factors produce weak factors, we set $\tilde{m} = 7$.
The Varimax-rotated maximum-likelihood estimate of the factor loadings matrix is given in Table \ref{SMTABLE:FacPatMes}.
The 7 common factors explain approximately 76\% of the variance in the observables.

\begin{table}[ht]
\centering
\caption{Factor pattern for the oral squamous cell carcinoma training data.}
\label{SMTABLE:FacPatMes}
\scalebox{.8}{
\begin{tabular}{l r r r r r r r}
\toprule
 & \multicolumn{7}{c}{Factor} \\
 \cline{2-8}
Feature             &1      &2      &3   &4      &5      &6    &7\\
\midrule
Coefficient of variation	        &-0.69 &         &        &  0.57 &         &       &        \\
Kurtosis           				    & 0.58 &         &        &  0.40 &         &       &        \\
Entropy                 			&-0.61 &  -0.48  &        &       &         &       &        \\
Joint entropy          				&-0.91 &         &        &       &  -0.33  &       &        \\
Sum entropy            				&-0.96 &         &        &       &         &       &        \\
Angular second moment  				& 0.85 &         &        &       &   0.32  &       &        \\
Contrast               				&-0.65 &  -0.32  &        &       &         &       &  -0.57 \\
Inverse difference normalized   	& 0.69 &         &        &       &         &       &   0.53 \\
Inverse variance           		    & 0.69 &   0.32  &        &       &         &       &   0.44 \\
Cluster prominence         			&-0.73 &         &        &       &   0.41  &       &        \\
Grey level non-uniformity normalized& 0.90 &         &        &       &         &       &        \\
Run length non-uniformity normalized&-0.68 &  -0.51  &        &       &         &       &  -0.32 \\
Grey level variance                 &-0.83 &         &        &       &         &       &        \\
Run length variance                 & 0.66 &   0.50  &        &       &   0.35  &       &        \\
Run entropy                     	&-0.88 &         &        &       &         &       &        \\
Maximum 3D diameter        			&      &   0.82  &        &       &         &       &        \\
Surface area                   		&      &   0.89  &        &       &         &       &        \\
Surface to volume ratio             &      &  -0.70  &        &       &   0.31  &  0.42 &        \\
Radius of an equivolumetric sphere	&      &   0.91  &        &       &         &       &        \\
Total energy                  		&      &   0.72  &  0.31  &       &         &       &        \\
Fractal dimension (fitted)			&      &   0.55  &        &       &  -0.32  &       &        \\
Fractal abundance       			&      &   0.85  &        &       &         &       &        \\
Fractal lacunarity      			&      &  -0.67  &        &       &         &       &        \\
Grey level non-uniformity			& 0.31 &   0.87  &        &       &         &       &        \\
Run length non-uniformity			&      &   0.90  &        &       &         &       &        \\
Whole MRI median         			&      &         &  0.54  &       &         &       &        \\
Maximum grey level values $> .5$	&      &         &  0.77  &       &         &       &        \\
Mean grey level values $> .5$		&      &         &  0.85  &       &         &       &        \\
Tumor MRI minimum grey level		&      &  -0.32  &  0.62  &       &         &       &        \\
Tumor MRI range               		&      &         &  0.82  &       &         &       &        \\
Mean Laplacian          			&      &         &  0.83  &       &         &       &        \\
Variance                 			&      &         &  0.81  &       &         &       &        \\
Root-mean-square           			&      &         &  0.92  &       &         &       &        \\
Mean of the maximum and adjacent voxels	&      &         &  0.89  &       &         &       &        \\
Skewness               				&      &         &        &  0.59 &         &       &        \\
Area under intensity-volume histogram curve	&      &         &        & -0.74 &         &       &        \\
Short run low grey level emphasis	&      &         &        &  0.91 &         &       &        \\
Short run high grey level emphasis	&      &         &        & -0.92 &         &       &        \\
Long run low grey level emphasis	& 0.40 &   0.42  &        &  0.69 &         &       &        \\
Long run high grey level emphasis	&      &         &        & -0.88 &         &       &        \\
Moran's I               			&      &         &        &       &   0.71  &       &        \\
Geary's C               			&      &         &        &       &  -0.68  &       &        \\
Measure of information correlation 1&      &   0.37  &        &       &  -0.81  &       &        \\
Measure of information correlation 2&      &  -0.30  &        &       &   0.82  &       &        \\
Compactness B           			&      &         &        &       &         & -0.86 &        \\
Asphericity            				&      &         &        &       &         &  0.91 &        \\
Correlation                    		&      &   0.32  &        &       &   0.54  &       &   0.61 \\
Median grey level values $> .5$		&      &         &  0.49  &       &         &       &        \\
Fractal dimension (calculated)      &      &   0.39  &        &       &         &       &        \\
Inverse difference moment normalized& 0.44 &         &        &       &         &       &   0.48 \\
Cluster shade          				&      &         &        &  0.45 &         &       &        \\
\bottomrule
\multicolumn{8}{l}{}\\[-0.75\normalbaselineskip]%
\multicolumn{8}{l}{\footnotesize{Loadings $< |.3|$ omitted.}}%
\end{tabular}
}
\end{table}

In the third step, factor scores are obtained through the approach outlined in Section \ref{SMSSEC:FS} as well as Section 2.2.3.\ of the Main Text.
Using the estimates of the factor solution in conjunction with the training data will give the factor scores for the training data.
Using the estimates of the factor solution in conjunction with the validation data will give the factor scores for the validation data.
The factor scores are highly determinate with $\mathrm{SMC}_{k}^{\xi} > .9$ for all factors.

\subsection{Prediction results}
\label{SMSSEC:ORES}
\begin{figure}[h!]
\centering
  \includegraphics[width=\textwidth]{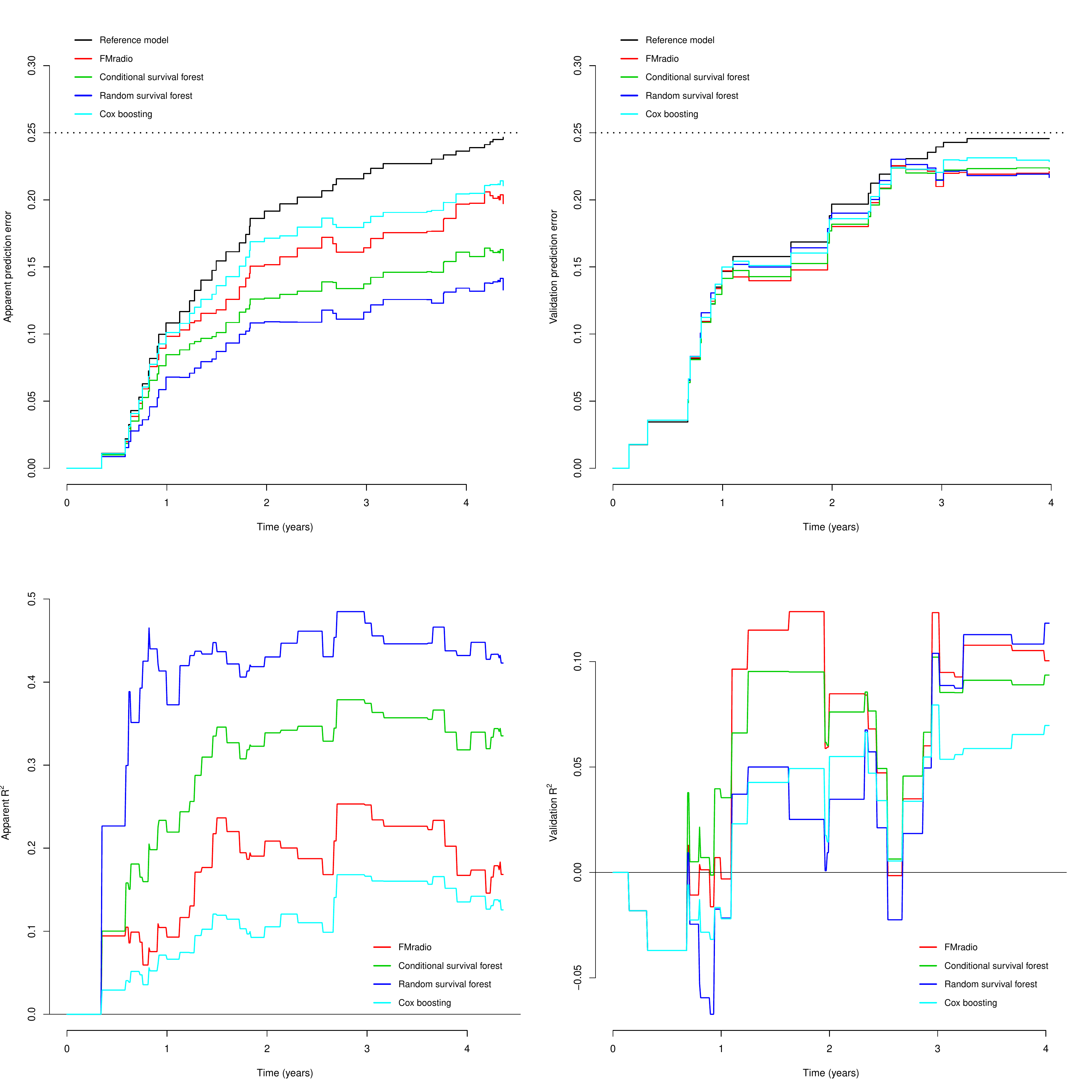}
    \caption{Visualizations for the external validation setting.
    The upper-panels contain prediction error curves.
    The bottom-panels contain $R^{2}$ plots.
    The left-hand panels pertain to apparent performance while the right-hand panels visualize the validation performance.
    }
  \label{SMFIG:ResultsEVD}
\end{figure}

Figure \ref{SMFIG:ResultsEVD} gives a visualization of the prediction results provided in Section 3.4.\ of the Main Text.

\section{Additional information head and neck squamous cell carcinoma data}
\label{SMSEC:ADIhnscc}
In this section we give additional information on the head and neck squamous cell carcinoma data as used in the internal validation setting.
Section \ref{SMSSEC:HNIP} contains an overview of data acquisition.
Section \ref{SMSSEC:HNFP} contains information on the factor solution obtained in the training projection.
Lastly, Section \ref{SMSSEC:HNFALL} gives the analysis results when redundancy-filtering is applied for all methods.

\subsection{Image processing and radiomic feature extraction}
\label{SMSSEC:HNIP}
A total of $175$ head and neck squamous cell carcinoma patients were available from the Amsterdam University medical centers.
On these patients low-dose $^{18}$F-FDG-PET/CT was performed according to the European Association of Nuclear Medicine (EANM) guidelines $2.0$ on a Gemini-TF or Ingenuity TF-64 PET/CT (Philips Heatlthcare, Best, The Netherlands) with EARL (EANM Research Ltd.) accreditation.
The PET images were reconstructed using iterative ordered subsets expectation maximization with the parameters optimized for the head and neck area (i.e., $4$ iterations, $16$ subsets, $5$ \si{\milli\metre} $3$-dimensional Gaussian filter) with photon attenuation correction.
Reconstructed images had an image matrix size of $144 \times 144$, voxel size of $4 \times 4 \times 4$ \si{\milli\metre}.
Low-dose-CT was collected using a beam current of $50$ \si{\mA\per\second} at $120$ \si{\kV} for anatomical correlation of $^{18}$F-FDG uptake and attenuation correction.
CT-scans were reconstructed using an image matrix size of $512 \times 512$ resulting in pixel sizes of $1.17 \times 1.17$ \si{\milli\metre} and a slice thickness of 5 \si{\milli\metre}.

Radiomics features were extracted from an in-house build Accurate tool \cite{AccurateBoel_SM}.
It provides a 3D implementation of feature extraction methods for shape, intensity, texture and wavelet-type features.
Wavelet-type features were obtained through a wavelet transform using a Coiflet scaling function.
For the texture and wavelet analysis, images were discretized to a fixed bin size of $64$ bins and $0.25$ \si{\gram\per\milli\litre} standardized uptake value (SUV) in the CT and PET, respectively \citep{MRIextract_SM}.
This choice of bin sizes resulted in similar numbers of bins in both modalities.
On average $64$ bins were analyzed.
For each patient a total of $p = 436$ radiomic features were extracted on the primary tumors.
Table \ref{SMTAB:FEatMart} gives a listing of all extracted features.

The ngldmFeatures3D\_Dependence count percentage, ngldmFeatures2Dmrg\_Dependence count percentage, gldzmFeatures2Davg\_Grey level variance GLDZM, and Intensity histogram\_minimum features were removed as they were behaving as de facto constants (no variation over the respective patients).
In addition, $1$ person with a follow-up time of $0$ was removed from the data.
Hence, the final data object concerned $p = 432$ radiomic features extracted from the primary tumors of $n = 174$ patients.

Additional information on the data acquisition and radiomic features can be found in \citet{Martens19_SM}.
The processed data object is available from the Authors upon reasonable request.

\begin{table}[ht]
\caption{Extracted radiomic features on the head and neck squamous cell carcinoma images.}
\label{SMTAB:FEatMart}
\scalebox{.34}{
\begin{tabular}{llll}
\toprule
Morphology\_Volume                                          & glcmFeatures2Dmrg\_inverse.difference.moment                      & GLRLMFeatures2DWmrg\_Grey.level.non.uniformity                    & 	 ngtdmFeatures2Dmrg\_coarseness                                        \\
Morphology\_approximate.volume                              & glcmFeatures2Dmrg\_inverse.difference.moment.normalised           & GLRLMFeatures2DWmrg\_Grey.level.non.uniformity.normalized         & 	 ngtdmFeatures2Dmrg\_contrast                                          \\
Morphology\_Surface                                         & glcmFeatures2Dmrg\_inverse.variance                               & GLRLMFeatures2DWmrg\_Run.length.non.uniformity                    & 	 ngtdmFeatures2Dmrg\_busyness                                          \\
Morphology\_Surface.to.volume.ratio                         & glcmFeatures2Dmrg\_correlation                                    & GLRLMFeatures2DWmrg\_Run.length.non.uniformity.normalized         & 	 ngtdmFeatures2Dmrg\_complexity                                        \\
Morphology\_Compactness1                                    & glcmFeatures2Dmrg\_autocorrelation                                & GLRLMFeatures2DWmrg\_Run.percentage                               & 	 ngtdmFeatures2Dmrg\_strength                                          \\
Morphology\_Compactness2                                    & glcmFeatures2Dmrg\_cluster.tendency                               & GLRLMFeatures2DWmrg\_Grey.level.variance                          & 	 ngtdmFeatures3D\_coarseness                                           \\
Morphology\_Spherical.disproportion                         & glcmFeatures2Dmrg\_cluster.shade                                  & GLRLMFeatures2DWmrg\_Run.length.variance                          & 	 ngtdmFeatures3D\_contrast                                             \\
\end{tabular}
}
\end{table}

\subsection{Factor pattern training data}
\label{SMSSEC:HNFP}
\begin{figure}[h!]
\centering
  \includegraphics[width=\textwidth]{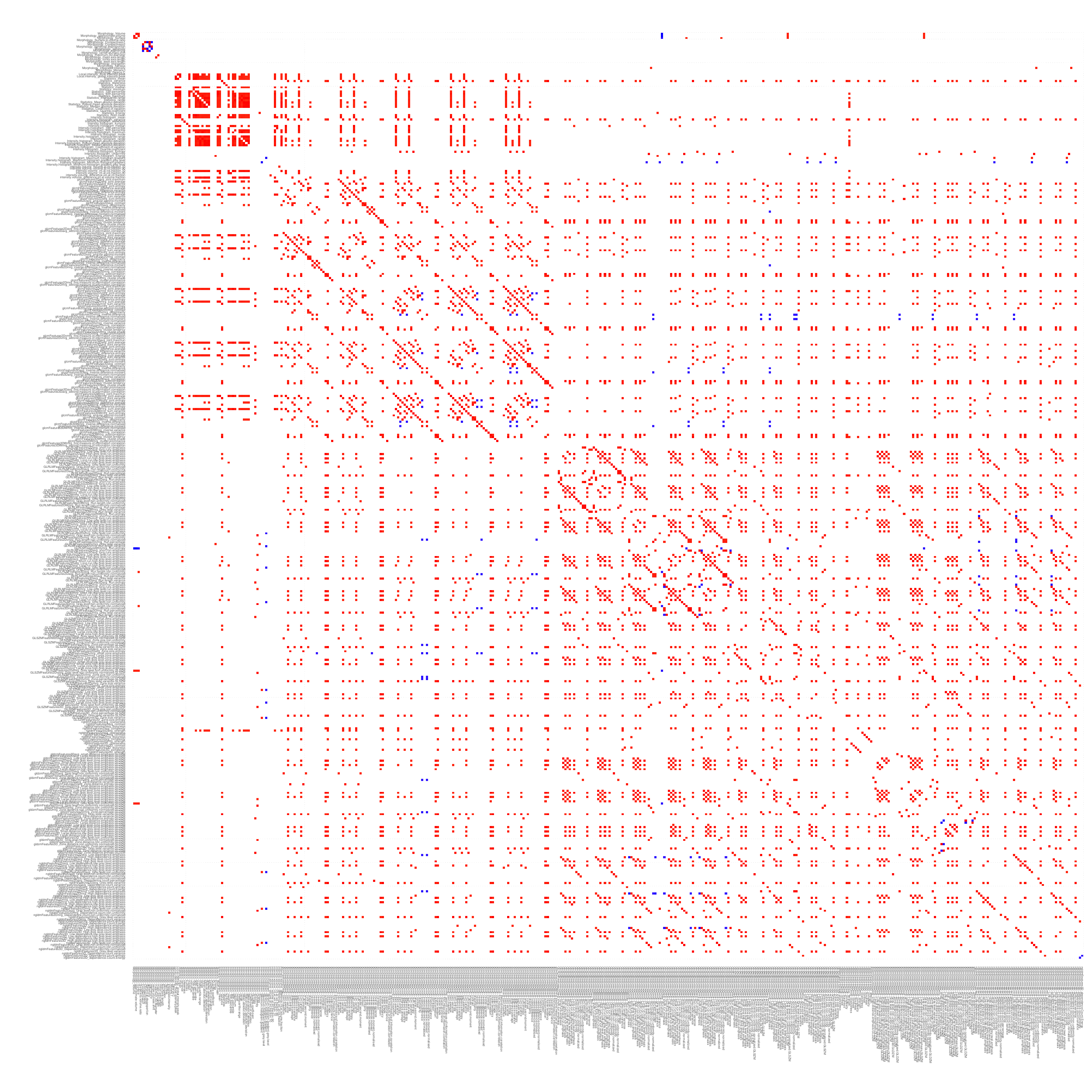}
    \caption{Thresholded correlation heatmap of the radiomic features for the head and neck squamous cell carcinoma data.
    Absolute correlations $< .95$ were set to zero such that only correlations equalling or exceeding an absolute marginal correlation threshold $\tau$ of $.95$ are visualized.
    Red then indicates a positive correlation $\geq .95$.
    Blue then indicates a negative correlation $\leq -.95$.
    Colored regions then visualize blocks of information redundancy.
    }
  \label{SMFIG:IVDredundant}
\end{figure}

Again, the first step in the factor modeling pipeline is to get a regularized estimate of the correlation matrix on the nonredundant radiomic features.
Figure \ref{SMFIG:IVDredundant} visualizes the redundancy pattern among the full set of $432$ radiomic features.
The redundancy filtering algorithm (Algorithm \ref{AlgRF}) was run with $\tau$ set to $.95$ and it retained $p^* = 124$ out of the original $p = 432$ features.
The remaining correlation matrix was then subjected to penalized ML estimation.
The optimal value for the penalty parameter was determined by 5-fold CV of the log-likelihood function and was found to be $.0591859$.
The resulting regularized correlation matrix was well-conditioned, as indicated by a relatively low condition number ($805.10$).

The second step consists of factor analytic data compression.
The GB indicates $13$ factors as an upper-bound to the optimal dimension of the latent vector.
As the proportion of additional variance explained flattens after 8 common factors and as solutions in excess of 8 factors produce weak factors, we set $\tilde{m} = 8$.
The Varimax-rotated maximum-likelihood estimate of the factor loadings matrix is given in Table \ref{SMTABLE:FacPatMart}.
The 8 common factors explain approximately 79\% of the variance in the observables.
The factor scores obtained in the third step are highly determinate with $\mathrm{SMC}_{k}^{\xi} > .94$ for all factors.

\begin{table}[ht]
\centering
\caption{Factor pattern for the head and neck squamous cell carcinoma training data.}
\label{SMTABLE:FacPatMart}
\scalebox{.37}{
\begin{tabular}{l r r r r r r r r}
\toprule
 & \multicolumn{8}{c}{Factor} \\
 \cline{2-9}
Feature             &1      &2      &3   &4      &5      &6    &7  &8 \\
\midrule
Morphology\_Surface.to.volume.ratio                              & -0.86 &        &       &        &          &         &        &         \\
Morphology\_minor.axis.length                                    &  0.80 &        &       &        &     0.32 &         &        &         \\
Morphology\_least.axis.length                                    &  0.81 &        &       &        &     0.42 &         &        &         \\
glcmFeatures2Dmrg\_inverse.difference.normalised                 &  0.62 &        &       &        &          &   0.32  &        &         \\
glcmFeatures2Dmrg\_correlation                                   &  0.83 &        &       &        &          &         &        &         \\
glcmFeatures2Dmrg\_second.measure.of.information.correlation     &  0.59 &   0.59 &       &        &          &         & -0.31  &         \\
glcmFeatures2Dvmrg\_inverse.difference.normalised                &  0.79 &        &       &        &          &         &        &         \\
glcmFeatures2Dvmrg\_inverse.difference.moment.normalised         &  0.84 &        &       &        &          &         &        &         \\
glcmFeatures2Dvmrg\_correlation                                  &  0.78 &        &       &        &          &         &        &  0.49   \\
glcmFeatures2Dvmrg\_second.measure.of.information.correlation    &  0.55 &   0.43 &       &        &          &         &        &  0.55   \\
glcmFeatures3DWmrg\_inverse.difference.normalised                &  0.80 &        &       &        &          &         &        &         \\
glcmFeatures3DWmrg\_inverse.difference.moment.normalised         &  0.84 &        &       &        &          &         &        &         \\
glcmFeatures3DWmrg\_correlation                                  &  0.84 &        &       &        &          &         &        &  0.34   \\
glcmFeatures3DWmrg\_second.measure.of.information.correlation    &  0.76 &        &       &        &          &         &        &  0.43   \\
GLRLMFeatures3Dmrg\_Run.length.non.uniformity                    &  0.75 &        &       &        &     0.49 &         &        &         \\
GLSZMFeatures2Davg\_Zone.size.non.uniformity                     &  0.78 &   0.32 &       &        &          &         &        &         \\
GLSZMFeatures2Dvmrg\_Small.zone.low.grey.level.emphasis          & -0.58 &  -0.52 &       &        &          &         &  0.40  &         \\
GLSZMFeatures2Dvmrg\_Large.zone.high.grey.level.emphasis         &  0.58 &        &       &    0.57&     0.44 &         &        &         \\
GLSZMFeatures3D\_Zone.size.non.uniformity                        &  0.65 &   0.34 &       &    0.43&          &         &        &         \\
GLSZMFeatures3D\_Zone.size.entropy                               &  0.63 &   0.46 &  -0.49&        &          &         &        &         \\
ngtdmFeatures2avg\_coarseness                                    & -0.64 &  -0.36 &       &        &          &         &        &         \\
ngtdmFeatures3D\_coarseness                                      & -0.68 &        &       &        &          &         &  0.50  &         \\
gldzmFeatures2Davg\_Grey.level.non.uniformity.GLDZM              &  0.69 &  -0.35 &       &        &     0.51 &         &        &         \\
gldzmFeatures2Davg\_Zone.distance.non.uniformity.GLDZM           &  0.82 &        &       &        &          &         &        &         \\
gldzmFeatures2Davg\_Zone.distance.non.uniformity.normalized.GLDZM& -0.72 &        &   0.33&        &          &   0.31  &        &         \\
gldzmFeatures2Davg\_Zone.distance.entropy.GLDZM                  &  0.74 &   0.45 &  -0.34&        &          &         &        &         \\
gldzmFeatures2Dmrg\_Zone.distance.non.uniformity.GLDZM           &  0.76 &        &       &        &     0.40 &   0.34  &        &         \\
gldzmFeatures2Dmrg\_Zone.distance.non.uniformity.normalized.GLDZM& -0.76 &        &   0.38&        &          &         &        &         \\
gldzmFeatures2Dmrg\_Zone.distance.variance.GLDZM                 &  0.83 &        &       &        &          &         &        &         \\
gldzmFeatures3D\_Zone.distance.non.uniformity.GLDZM              &  0.79 &        &       &        &          &         &        &         \\
gldzmFeatures3D\_Zone.distance.non.uniformity.normalized.GLDZM   & -0.56 &  -0.54 &       &        &          &         &        &         \\
gldzmFeatures3D\_Zone.distance.variance.GLDZM                    &  0.67 &   0.39 &       &        &          &         &        &         \\
ngldmFeatures2Davg\_Low.dependence.low.grey.level.emphasis       & -0.65 &  -0.44 &       &        &          &         &  0.33  &         \\
ngldmFeatures2Davg\_Dependence.count.non.uniformity              &  0.82 &        &       &        &          &         &        &         \\
ngldmFeatures2Davg\_Dependence.count.entropy                     &  0.86 &        &       &        &          &         &        &         \\
ngldmFeatures2Davg\_dependence.Count.Energy                      & -0.64 &  -0.31 &       &        &          &         &  0.37  &         \\
ngldmFeatures2Dmrg\_Dependence.count.entropy                     &  0.73 &   0.39 &  -0.30&        &          &         &        &         \\
ngldmFeatures2Dmrg\_dependence.Count.Energy                      & -0.62 &  -0.40 &   0.44&        &          &         &  0.31  &         \\
ngldmFeatures3D\_Low.dependence.low.grey.level.emphasis          & -0.73 &        &       &        &          &         &        &         \\
ngldmFeatures3D\_High.dependence.high.grey.level.emphasis        &  0.71 &        &       &    0.53&          &         &        &         \\
ngldmFeatures3D\_Dependence.count.non.uniformity                 &  0.78 &        &       &        &     0.33 &         &        &         \\
ngldmFeatures3D\_dependence.Count.Energy                         & -0.75 &  -0.39 &       &        &          &         &  0.30  &         \\
Statistics\_median                                               &  0.39 &   0.61 &       &    0.58&          &         &        &         \\
Statistics\_minimum                                              &       &   0.61 &       &    0.46&          &         &        &         \\
Statistics\_Coefficient.of.variation                             &       &   0.63 &       &        &          &         &        &         \\
Statistics\_Quartile.coefficient                                 &       &   0.61 &       &        &          &         &        &         \\
glcmFeatures2Davg\_difference.entropy                            &  0.52 &   0.65 &  -0.32&        &          &         &        &         \\
glcmFeatures2Davg\_first.measure.of.information.correlation      &       &  -0.84 &       &        &          &         &        &         \\
glcmFeatures2Dmrg\_dissimilarity                                 &       &   0.80 &       &    0.44&          &         &        &         \\
glcmFeatures2Dmrg\_inverse.difference.moment                     &       &  -0.74 &   0.52&        &          &         &        &         \\
glcmFeatures2Dmrg\_inverse.variance                              &       &  -0.82 &       &        &          &         &        &         \\
glcmFeatures2Dmrg\_first.measure.of.information.correlation      & -0.37 &  -0.73 &       &   -0.30&          &         &        &         \\
glcmFeatures3Davg\_first.measure.of.information.correlation      &  0.36 &  -0.59 &       &        &          &         &        & -0.36   \\
glcmFeatures3Davg\_second.measure.of.information.correlation     &       &   0.78 &       &        &          &         &        &         \\
glcmFeatures3DWmrg\_difference.entropy                           &       &   0.80 &  -0.41&        &          &         &        &         \\
glcmFeatures3DWmrg\_contrast                                     &       &   0.70 &       &    0.61&          &         &        &         \\
glcmFeatures3DWmrg\_inverse.variance                             &       &  -0.89 &       &        &          &         &        &         \\
GLRLMFeatures3Dmrg\_Run.entropy                                  &  0.54 &   0.69 &       &    0.30&          &         &        &         \\
GLSZMFeatures2Davg\_small.zone.emphasis                          &       &   0.68 &  -0.53&        &          &         &        &         \\
GLSZMFeatures2Davg\_Zone.size.non.uniformity.normalized          &       &   0.84 &       &        &          &         &        &         \\
GLSZMFeatures3D\_small.zone.emphasis                             &       &   0.74 &  -0.33&        &          &         &        &         \\
GLSZMFeatures3D\_Small.zone.low.grey.level.emphasis              & -0.51 &  -0.58 &       &        &          &         &        &         \\
GLSZMFeatures3D\_Zone.size.non.uniformity.normalized             &       &   0.80 &       &    0.35&          &         &        &         \\
ngtdmFeatures2avg\_strength                                      &       &   0.67 &       &    0.61&          &         &        &         \\
ngtdmFeatures3D\_contrast                                        &       &   0.80 &       &    0.38&          &         &        &         \\
ngtdmFeatures3D\_strength                                        &       &   0.70 &       &    0.48&          &         &        &         \\
ngldmFeatures2Dmrg\_Low.dependence.emphasis                      &       &   0.81 &  -0.45&        &          &         &        &         \\
ngldmFeatures2Dmrg\_Dependence.count.non.uniformity.normalized   &       &   0.88 &       &        &          &         &        &         \\
ngldmFeatures2Dmrg\_Dependence.count.variance                    &       &  -0.63 &   0.54&        &          &         &        &         \\
ngldmFeatures3D\_Low.dependence.emphasis                         &       &   0.88 &       &        &          &         &        &         \\
ngldmFeatures3D\_Dependence.count.non.uniformity.normalized      &       &   0.85 &       &        &          &         &        &         \\
Morphology\_center.of.mass.shift                                 &       &  -0.35 &   0.65&        &          &         &        &         \\
intensity.volume\_difference.vol.at.int.fraction                 &  0.45 &        &  -0.59&        &          &         & -0.48  &         \\
glcmFeatures2Davg\_angular.second.moment                         & -0.45 &  -0.31 &   0.53&        &          &         &  0.47  &         \\
glcmFeatures2Dmrg\_joint.maximum                                 & -0.37 &  -0.42 &   0.62&        &          &         &  0.39  &         \\
glcmFeatures2Dvmrg\_first.measure.of.information.correlation     &       &        &  -0.55&        &          &         &        & -0.55   \\
glcmFeatures3DWmrg\_angular.second.moment                        &       &        &   0.82&        &          &         &  0.31  &         \\
glcmFeatures3DWmrg\_first.measure.of.information.correlation     & -0.45 &        &  -0.69&        &          &         &        & -0.32   \\
GLRLMFeatures2DWmrg\_long.runs.emphasis                          &       &        &   0.88&        &          &         &        &         \\
GLRLMFeatures2DWmrg\_Run.percentage                              &       &   0.56 &  -0.60&        &          &         &        &         \\
GLRLMFeatures2Dvmrg\_Long.run.low.grey.level.emphasis            &       &        &   0.94&        &          &         &        &         \\
GLRLMFeatures3Davg\_Long.run.low.grey.level.emphasis             &       &        &   0.86&        &          &         &        &         \\
GLRLMFeatures3Davg\_Run.length.variance                          &       &        &   0.89&        &          &         &        &         \\
GLRLMFeatures3Dmrg\_Run.percentage                               &       &   0.53 &  -0.79&        &          &         &        &         \\
GLSZMFeatures2Dvmrg\_Zone.size.variance                          &       &        &   0.75&        &     0.46 &         &        &         \\
GLSZMFeatures3D\_Large.zone.low.grey.level.emphasis              &       &        &   0.88&        &          &         &        &         \\
ngtdmFeatures2avg\_busyness                                      &       &        &   0.83&        &          &         &        &         \\
ngtdmFeatures3D\_busyness                                        &       &        &   0.87&        &          &         &        &         \\
ngldmFeatures3D\_Low.grey.level.count.emphasis                   & -0.36 &  -0.32 &   0.72&        &          &         &  0.40  &         \\
ngldmFeatures3D\_Grey.level.non.uniformity.normalized            & -0.37 &  -0.50 &   0.68&        &          &         &        &         \\
ngldmFeatures3D\_Dependence.count.variance                       &       &  -0.42 &   0.77&        &          &         &        &         \\
Intensity.histogram\_mode                                        &  0.34 &        &       &    0.57&          &         &        &         \\
Intensity.histogram\_Maximum.histogram.gradient.grey.level       &  0.33 &   0.39 &       &    0.64&          &         &        &         \\
Intensity.histogram\_Minimum.histogram.gradient.grey.level       &  0.48 &   0.43 &       &    0.51&          &         &        &         \\
intensity.volume\_int.at.vol.fraction.90                         &  0.51 &   0.47 &       &    0.61&          &         &        &         \\
glcmFeatures2Dmrg\_cluster.shade                                 &       &        &       &   -0.76&          &         &        &         \\
glcmFeatures3DWmrg\_cluster.prominence                           &       &        &       &    0.77&          &         &        &         \\
GLRLMFeatures2Dvmrg\_Grey.level.variance                         &  0.55 &        &       &    0.61&          &         &        &         \\
gldzmFeatures2Dmrg\_Large.distance.high.grey.level.emphasis.GLDZM&  0.39 &   0.34 &       &    0.77&          &         &        &         \\
ngldmFeatures2Dmrg\_High.dependence.high.grey.level.emphasis     &  0.55 &        &       &    0.70&          &         &        &         \\
ngldmFeatures3D\_Low.dependence.high.grey.level.emphasis         &       &   0.51 &       &    0.79&          &         &        &         \\
GLRLMFeatures3Dmrg\_Grey.level.non.uniformity                    &  0.35 &        &       &        &     0.84 &         &        &         \\
GLSZMFeatures3D\_Large.zone.emphasis                             &       &        &   0.49&        &     0.76 &         &        &         \\
GLSZMFeatures3D\_Large.zone.high.grey.level.emphasis             &       &        &       &        &     0.88 &         &        &         \\
gldzmFeatures2Dmrg\_Grey.level.non.uniformity.GLDZM              &  0.59 &        &       &        &     0.66 &         &        &         \\
ngldmFeatures2Davg\_Grey.level.non.uniformity                    &  0.32 &  -0.32 &   0.38&        &     0.75 &         &        &         \\
Morphology\_Compactness2                                         & -0.35 &        &       &        &          &  -0.79  &        &         \\
Morphology\_asphericity                                          &  0.30 &        &       &        &          &   0.78  &        &         \\
Morphology\_major.axis.length                                    &  0.58 &        &       &        &          &   0.64  &        &         \\
Morphology\_elongation                                           &       &        &       &        &          &  -0.60  &        &         \\
Morphology\_flatness                                             &       &        &       &        &          &  -0.75  &        &         \\
Morphology\_Morans.I                                             & -0.32 &        &       &        &          &         &        &  0.62   \\
Morphology\_Gearys.C                                             &       &        &  -0.41&        &          &         & -0.34  &         \\
Intensity.histogram\_skewness                                    &       &        &       &        &          &   0.35  &        &  0.44   \\
Intensity.histogram\_kurtosis                                    &       &        &       &        &          &   0.47  &        &         \\
Intensity.histogram\_Coefficient.of.variation                    & -0.48 &        &       &   -0.32&          &         &        &  0.41   \\
Intensity.histogram\_Quartile.coefficient                        & -0.38 &        &       &        &          &         &        &         \\
intensity.volume\_volume.at.int.fraction.90                      & -0.33 &        &   0.37&        &          &  -0.34  &  0.34  &         \\
glcmFeatures2Davg\_second.measure.of.information.correlation     &  0.47 &   0.50 &  -0.40&        &          &         & -0.36  &         \\
glcmFeatures3DWmrg\_cluster.shade                                &       &        &       &        &          &         &        &         \\
GLRLMFeatures2DWmrg\_Short.run.low.grey.level.emphasis           & -0.46 &  -0.42 &   0.49&        &          &         &  0.48  &         \\
ngtdmFeatures2avg\_contrast                                      &       &   0.33 &       &    0.36&          &         &        &         \\
gldzmFeatures3D\_Large.distance.low.grey.level.emphasis.GLDZM    & -0.45 &  -0.50 &   0.47&        &          &         &  0.40  &         \\
ngldmFeatures2Davg\_Dependence.count.percentage                  &  0.45 &        &       &        &          &   0.33  &	     &	       \\
\bottomrule
\multicolumn{9}{l}{}\\[-0.75\normalbaselineskip]%
\multicolumn{9}{l}{\footnotesize{Loadings $< |.3|$ omitted.}}%
\end{tabular}
}
\end{table}

\subsection{Results under redundancy filtering for all methods}
\label{SMSSEC:HNFALL}
\begin{figure}[b!]
\centering
  \includegraphics[width=.94\textwidth]{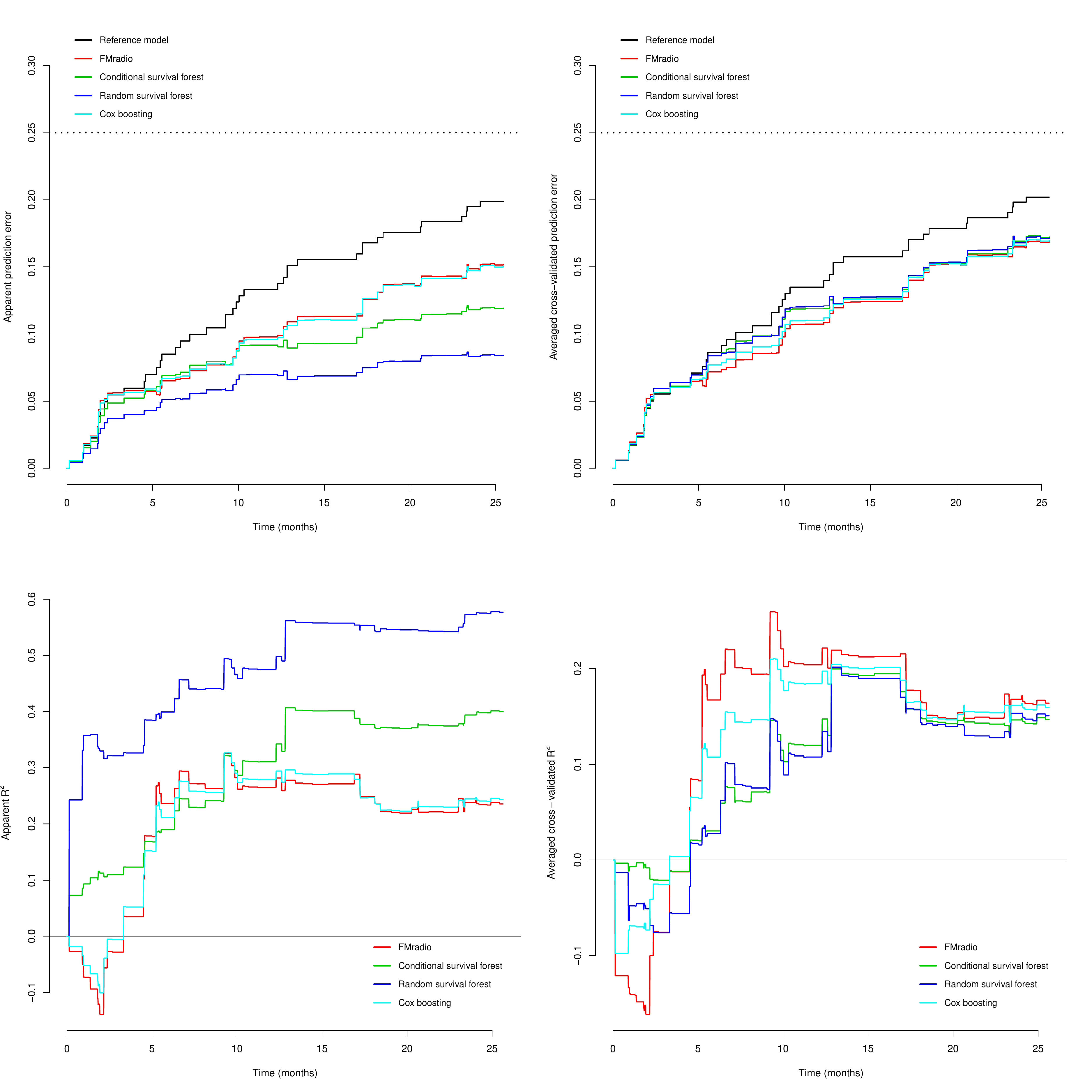}
    \caption{Visualizations for the internal validation setting under redundancy filtering for all methods.
    The upper-panels contain prediction error curves.
    The bottom-panels contain $R^{2}$ plots.
    The left-hand panels pertain to apparent performance while the right-hand panels visualize the averaged cross-validation performance.}
  \label{SMFIG:ResultsIVDfiltered}
\end{figure}

Figure \ref{SMFIG:ResultsIVDfiltered} and Table \ref{SMTABLE:ResultsIVDfiltered} contain the results when the RSF, CSF, and Cox boosting approaches are based on the same redundancy-filtered feature-set as the \texttt{FMradio} approach.
Table \ref{SMTABLE:ResultsIVDfiltered} contains the apparent and averaged cross-validated prediction errors (integrated Brier scores) as well as the overall time-dependent explained variation based on integrated Brier scores with the KM estimator as the reference model.
We see that the three alternative methods all fare somewhat better when redundancy filtering is applied.
Nonetheless, the overall results remain qualitatively the same as discussed in Section 3.5.\ of the Main Text.
The most stable approach resulting in the lowest prediction error is the proposed \texttt{FMradio} approach.

\begin{table}[h!]
\caption{Integrated apparent and averaged cross-validated Brier scores and explained residual variations.}
\label{SMTABLE:ResultsIVDfiltered}
\begin{tabular}{lcclcc}
\toprule
                            & \multicolumn{2}{c}{$\mathcal{B}^{I}$} &  & \multicolumn{2}{c}{$R^{2}$} \\ \cline{2-3} \cline{5-6}
                            & Apparent      & Cross-validated    &  & Apparent         &  Cross-validated      \\ \midrule
Reference model             & .128          & .130               &  & --               & --                    \\
\texttt{FMradio}            & .098          & \textbf{.108}      &  & .236             & \textbf{.169}         \\
Conditional survival forest & .084          & .113               &  & .340             & .132                  \\
Random survival forest      & .062          & .113               &  & .515             & .127                  \\
Cox boosting                & .097          & .109               &  & .242             & .157                  \\
\bottomrule
\end{tabular}
\end{table}

\clearpage

\begin{thebibliography}{51}
\providecommand{\natexlab}[1]{#1}
\providecommand{\url}[1]{\texttt{#1}}
\expandafter\ifx\csname urlstyle\endcsname\relax
  \providecommand{\doi}[1]{doi: #1}\else
  \providecommand{\doi}{doi: \begingroup \urlstyle{rm}\Url}\fi

\bibitem[Aerts(2018)]{Aerts18DataScience}
H.~J. W.~L. Aerts.
\newblock Data science in radiology: A path forward.
\newblock \emph{Clinical Cancer Research}, 24:\penalty0 532--534, 2018.

\bibitem[Aerts et~al.(2014)Aerts, {Rios-Velazquez}, Leijenaar, Parmar,
  Grossmann, Carvalho, Bussink, Monshouwer, {Haibe-Kains}, Rietveld, Hoebers,
  Rietbergen, Leemans, Dekker, Quackenbush, Gillies, and Lambin]{AertsLHN}
H.~J. W.~L. Aerts, E.~{Rios-Velazquez}, R.~T.~H. Leijenaar, C.~Parmar,
  P.~Grossmann, S.~Carvalho, J.~Bussink, R.~Monshouwer, B.~{Haibe-Kains},
  D.~Rietveld, F.~Hoebers, M.~M. Rietbergen, R.~Leemans, A.~Dekker,
  J.~Quackenbush, R.~J. Gillies, and P.~Lambin.
\newblock Decoding tumour phenotype by noninvasive imaging using a quantitative
  radiomics approach.
\newblock \emph{Nature Communications}, 5:\penalty0 4006, 2014.

\bibitem[Balagurunathan et~al.(2014)Balagurunathan, Kumar, Gu, Kim, Wang, Liu,
  Goldgof, Hall, Korn, Zhao, Schwartz, Basu, Eschrich, Gatenby, and
  Gillies]{Bala14CT}
Y.~Balagurunathan, V.~Kumar, Y.~Gu, J.~Kim, H.~Wang, Y.~Liu, D.~B. Goldgof,
  L.~O. Hall, R.~Korn, B.~Zhao, L.~H. Schwartz, S.~Basu, S.~A. Eschrich, R.~A.
  Gatenby, and R.~J. Gillies.
\newblock Test--retest reproducibility analysis of lung {CT} image features.
\newblock \emph{Journal of Digital Imaging}, 27:\penalty0 805--823, 2014.

\bibitem[Binder(2013)]{CoxBoost}
H.~Binder.
\newblock \emph{\texttt{CoxBoost}: {Cox} models by likelihood based boosting
  for a single survival endpoint or competing risks}, 2013.
\newblock R package version 1.4.

\bibitem[Binder and Schumacher(2008)]{Bind08}
H.~Binder and M.~Schumacher.
\newblock Allowing for mandatory covariates in boosting estimation of sparse
  high-dimensional survival models.
\newblock \emph{BMC Bioinformatics}, 9:\penalty0 14, 2008.

\bibitem[Binder et~al.(2009)Binder, Allignol, Schumacher, and
  Beyersmann]{Bind09}
H.~Binder, A.~Allignol, M.~Schumacher, and J.~Beyersmann.
\newblock Boosting for high-dimensional time-to-event data with competing
  risks.
\newblock \emph{Bioinformatics}, 25\penalty0 (7):\penalty0 890--896, 2009.

\bibitem[Brier(1950)]{Brier}
G.~W. Brier.
\newblock Verselection of forecasts expressed in terms of probability.
\newblock \emph{Monthly Weather Review}, 78:\penalty0 1--3, 1950.

\bibitem[Carvalho et~al.(2008)Carvalho, Chang, Lucas, Nevins, Wang, and
  West]{Carvalho08SFA}
C.~M. Carvalho, J.~Chang, J.~E. Lucas, J.~R. Nevins, Q.~Wang, and M.~West.
\newblock High-dimensional sparse factor modeling: Applications in gene
  expression genomics.
\newblock \emph{Journal of the American Statistical Association}, 103:\penalty0
  1438--1456, 2008.

\bibitem[Chen et~al.(2017)Chen, Zhang, Gan, Yang, and Li]{ChenAppLC}
B.~Chen, R.~Zhang, Y.~Gan, L.~Yang, and W.~Li.
\newblock Development and clinical application of radiomics in lung cancer.
\newblock \emph{Radiation Oncology}, 12:\penalty0 154, 2017.

\bibitem[Cook et~al.(2018)Cook, Azad, Owczarczyk, Siddique, and
  Goh]{CookLimitations}
G.~J.~R. Cook, G.~Azad, K.~Owczarczyk, M.~Siddique, and V.~Goh.
\newblock Challenges and promises of {PET} radiomics.
\newblock \emph{International Journal of Radiation Oncology, Biology, Physics},
  102:\penalty0 1083--1089, 2018.

\bibitem[Cox(1972)]{Cox72}
D.~R. Cox.
\newblock Regression models and life tables (with discussion).
\newblock \emph{Journal of the Royal Statistical Society, Series B},
  34:\penalty0 187--220, 1972.

\bibitem[{de Bin} et~al.(2014){de Bin}, Sauerbrei, and Boulesteix]{BSB14}
R.~{de Bin}, W.~Sauerbrei, and A.~L. Boulesteix.
\newblock Investigating the prediction ability of survival models based on both
  clinical and omics data: two case studies.
\newblock \emph{Statistics in Medicine}, 33:\penalty0 5310--5329, 2014.

\bibitem[Fletcher and Powell(1963)]{FPalgo63}
R.~Fletcher and M.~J.~D. Powell.
\newblock A rapidly convergent descent method for minimization.
\newblock \emph{Computer Journal}, 2:\penalty0 163--168, 1963.

\bibitem[Gillies et~al.(2016)Gillies, Kinahan, and Hricak]{RadiomicsOverview}
R.~J. Gillies, P.~E. Kinahan, and H.~Hricak.
\newblock Radiomics: Images are more than pictures, they are data.
\newblock \emph{Radiology}, 278:\penalty0 563--577, 2016.

\bibitem[Graf et~al.(1999)Graf, Schmoor, Sauerbrei, and
  Schumacher]{BrierModern}
E.~Graf, C.~Schmoor, W.~Sauerbrei, and M.~Schumacher.
\newblock Assessment and comparison of prognostic classifcation schemes for
  survival data.
\newblock \emph{Statistics in Medicine}, 18:\penalty0 2529--2545, 1999.

\bibitem[Hothorn et~al.(2004)Hothorn, Lausen, Benner, and
  {Radespiel-Tr\"{o}ger}]{CSF}
T.~Hothorn, B.~Lausen, A.~Benner, and M.~{Radespiel-Tr\"{o}ger}.
\newblock Bagging survival trees.
\newblock \emph{Statistics in Medicine}, 25\penalty0 (1):\penalty0 77--91,
  2004.

\bibitem[Hothorn et~al.(2017)Hothorn, Hornik, Strobl, and Zeileis]{party}
T.~Hothorn, K.~Hornik, C.~Strobl, and A.~Zeileis.
\newblock \emph{\texttt{party}: {A} laboratory for recursive partitioning},
  2017.
\newblock R package version 1.2-3.

\bibitem[Huang et~al.(2016)Huang, Liang, He, Tian, Liang, Chen, Ma, and
  Liu]{HuanAppCC}
Y.~Q. Huang, C.~H. Liang, L.~He, J.~Tian, C.~S. Liang, X.~Chen, Z.~L. Ma, and
  Z.~Y. Liu.
\newblock Development and validation of a radiomics nomogram for preoperative
  prediction of lymph node metastasis in colorectal cancer.
\newblock \emph{Journal of Clinical Oncology}, 18:\penalty0 2157--2165, 2016.

\bibitem[Ishawaran and Kogalur(2017)]{randomForestSRC}
H.~Ishawaran and U.~B. Kogalur.
\newblock \emph{\texttt{randomForestSRC}: {Random} forests for survival,
  regression, and classification}, 2017.
\newblock R package version 2.5.1.

\bibitem[Ishawaran et~al.(2008)Ishawaran, Kogalur, Blackstone, and Lauer]{RSF}
H.~Ishawaran, U.~B. Kogalur, E.~H. Blackstone, and M.~S. Lauer.
\newblock Random survival forests.
\newblock \emph{Annals of Applied Statistics}, 2\penalty0 (3):\penalty0
  841--860, 2008.

\bibitem[Jolliffe(1982)]{PCR}
I.~T. Jolliffe.
\newblock A note on the use of principal components in regression.
\newblock \emph{Journal of the Royal Statistical Society, Series C},
  31:\penalty0 300--303, 1982.

\bibitem[J\"{o}reskog(1967)]{Joreskog67}
K.~G. J\"{o}reskog.
\newblock Some contributions to maximum likelihood factor analysis.
\newblock \emph{Psychometrika}, 32:\penalty0 443--482, 1967.

\bibitem[Kaiser(1970)]{Kaiser70}
H.~F. Kaiser.
\newblock A second-generation little jiffy.
\newblock \emph{Psychometrika}, 35:\penalty0 401--415, 1970.

\bibitem[Kolossv\'{a}ry et~al.(2017)Kolossv\'{a}ry, Kar\'{a}dy, Szilveszter,
  Kitslaar, Hoffman, Merkely, and {Maurovich-Horvat}]{KolossvaryCP2017}
M.~Kolossv\'{a}ry, J.~Kar\'{a}dy, B.~Szilveszter, P.~Kitslaar, U.~Hoffman,
  B.~Merkely, and P.~{Maurovich-Horvat}.
\newblock Radiomic features are superior to conventional quantitative computed
  tomographic metrics to identify coronary plaques with napkin-ring sign.
\newblock \emph{Circulation: Cardiovascular Imaging}, 10:\penalty0 e006843,
  2017.

\bibitem[Kumar et~al.(2012)Kumar, Gua, Basu, Berglund, Eschrich, Schabath,
  Forster, Aerts, Dekker, Fenstermacher, Goldof, Hall, Lambin, Balagurunathan,
  Gatenby, and Gillies]{KumarOverview12}
V.~Kumar, Y.~Gua, S.~Basu, A.~Berglund, S.~A. Eschrich, M.~B. Schabath,
  K.~Forster, H.~J. W.~L. Aerts, A.~Dekker, D.~Fenstermacher, D.~B. Goldof,
  L.~O. Hall, P.~Lambin, Y.~Balagurunathan, R.~A. Gatenby, and R.~J. Gillies.
\newblock Radiomics: the process and the challenges.
\newblock \emph{Magnetic Resonance Imaging}, 30:\penalty0 1234--1248, 2012.

\bibitem[Lambin et~al.(2012)Lambin, {Rios-Velazquez}, Leijenaar, Carvalho, {van
  Stiphout}, Granton, Zegers, Gillies, Boellaard, Dekker, and
  Aerts]{RadiomicsOrigin}
P.~Lambin, E.~{Rios-Velazquez}, R.~Leijenaar, S.~Carvalho, R.~P. G.~M. {van
  Stiphout}, P.~Granton, C.~M.~L. Zegers, R.~Gillies, R.~Boellaard, A.~Dekker,
  and H.~J. W.~L. Aerts.
\newblock Radiomics: Extracting more information from medical images using
  advanced feature analysis.
\newblock \emph{European Journal of Cancer}, 48:\penalty0 441--446, 2012.

\bibitem[Lambin et~al.(2017)Lambin, Leijenaar, Deist, Peerlings, {de Jong},
  {van Timmeren}, Sanduleanu, Larue, Even, Jochems, {van Wijk}, Woodruff, {van
  Soest}, Lustberg, Roelofs, {van Elmpt}, Dekker, Mottaghy, Wildberger, and
  Walsh]{RadiomicsReviewLambin}
P.~Lambin, R.~T.~H. Leijenaar, T.~M. Deist, J.~Peerlings, E.~E.~C. {de Jong},
  J.~{van Timmeren}, S.~Sanduleanu, R.~T. H.~M. Larue, A.~J.~G. Even,
  A.~Jochems, Y.~{van Wijk}, H.~Woodruff, J.~{van Soest}, T.~Lustberg,
  E.~Roelofs, W.~{van Elmpt}, A.~Dekker, F.~M. Mottaghy, J.~E. Wildberger, and
  S.~and Walsh.
\newblock Radiomics: The bridge between medical imaging and personalized
  medicine.
\newblock \emph{Nature Reviews Clinical Oncology}, 14:\penalty0 749--762, 2017.

\bibitem[Larue et~al.(2017)Larue, Defraene, {de Ruysscher}, Lambin, and {van
  Elmpt}]{Larue17Methods}
R.~T. H.~M. Larue, G.~Defraene, D.~{de Ruysscher}, P.~Lambin, and W.~{van
  Elmpt}.
\newblock Quantitative radiomics studies for tissue characterization: a review
  of technology and methodological procedures.
\newblock \emph{British Journal of Radiology}, 90:\penalty0 20160665, 2017.

\bibitem[Liu et~al.(2009)Liu, Lafferty, and Wasserman]{NonPara}
H.~Liu, J.~Lafferty, and L.~Wasserman.
\newblock The nonparanormal: {S}emiparametric estimation of high dimensional
  undirected graphs.
\newblock \emph{Journal of Machine Learning Research}, 10:\penalty0 2295--2328,
  2009.

\bibitem[Martens et~al.(2019)Martens, Koopman, Noij, Pfaehler, \"{U}belh\"{o}r,
  Sharma, Vergeer, Leemans, Hoekstra, Maqsood, Zwezerijnen, Heymans, Peeters,
  {de Bree}, {de Graaf}, Castelijns, and Boellaard]{Martens19}
R.~M. Martens, T.~Koopman, D.~P. Noij, E.~Pfaehler, C.~\"{U}belh\"{o}r,
  S.~Sharma, M.~R. Vergeer, C.~R. Leemans, O.~S. Hoekstra, Y.~Maqsood,
  B.~Zwezerijnen, M.~W. Heymans, C.~F.~W. Peeters, R.~{de Bree}, P.~{de Graaf},
  J.~A. Castelijns, and R.~Boellaard.
\newblock Predictive value of quantitative $^{18}${F-FDG-PET} radiomics in
  patients with head and neck squamous cell carcinoma.
\newblock Technical report, Amsterdam University Medical Centers, 2019.

\bibitem[Mayr et~al.(2016)Mayr, Hofner, and Schmid]{MHS16}
A.~Mayr, B.~Hofner, and M.~Schmid.
\newblock Boosting the discriminatory power of sparse survival models via
  optimization of the concordance index and stability selection.
\newblock \emph{BMC Bioinformatics}, 17:\penalty0 288, 2016.

\bibitem[Mes et~al.(2019)Mes, {van Velden}, Peltenburg, Peeters, {te Beest},
  {van de Wiel}, Mekke, Mulder, Martens, Castelijns, Pameijer, {de Bree},
  Boellaard, Leemans, Brakenhoff, and {de Graaf}]{Mes19}
S.~W. Mes, F.~H.~P. {van Velden}, B.~Peltenburg, C.~F.~W. Peeters, D.~{te
  Beest}, M.~A. {van de Wiel}, J.~Mekke, D.~C. Mulder, R.~M. Martens, J.~A.
  Castelijns, F.~A. Pameijer, R.~{de Bree}, R.~Boellaard, C.~R. Leemans, R.~H.
  Brakenhoff, and P.~{de Graaf}.
\newblock Head and neck cancer outcome prediction by {MRI} radiomic signatures.
\newblock Technical report, Amsterdam University Medical Centers, 2019.

\bibitem[Mogensen et~al.(2012)Mogensen, Ishawaran, and Gerds]{PEcurves}
U.~B. Mogensen, H.~Ishawaran, and T.~A. Gerds.
\newblock Evaluating random forests for survival analysis using prediction
  error curves.
\newblock \emph{Journal of Statistical Software}, 50:\penalty0 11, 2012.

\bibitem[Mulaik(2010)]{Mulaik2010}
S.~A. Mulaik.
\newblock \emph{{Foundations of Factor Analysis}}.
\newblock {Boca Raton: Chapman \& Hall/CRC}, 2nd edition, 2010.

\bibitem[Parmar et~al.(2015)Parmar, Grossmann, Bussink, Lambin, and
  Aerts]{RadiomicsML}
C.~Parmar, P.~Grossmann, J.~Bussink, P.~Lambin, and H.~J. W.~L. Aerts.
\newblock Machine learning methods for quantitative radiomic biomarkers.
\newblock \emph{Scientific Reports}, 5:\penalty0 13087, 2015.

\bibitem[Peeters et~al.(2014)Peeters, Dziura, and {van Wesel}]{PeetersAEfa}
C.~F.~W. Peeters, J.~Dziura, and F.~{van Wesel}.
\newblock Pathophysiological domains underlying the metabolic syndrome: an
  alternative factor analytic strategy.
\newblock \emph{Annals of Epidemiology}, 24:\penalty0 762--770, 2014.

\bibitem[{R Development Core Team}(2011)]{Rman}
{R Development Core Team}.
\newblock \emph{R: A Language and Environment for Statistical Computing}.
\newblock R Foundation for Statistical Computing, Vienna, Austria, 2011.
\newblock URL \url{http://www.R-project.org/}.
\newblock {ISBN} 3-900051-07-0.

\bibitem[Schumacher et~al.(2007)Schumacher, Binder, and Gerds]{SBG07}
M.~Schumacher, H.~Binder, and T.~A. Gerds.
\newblock Assessment of survival prediction models based on microarray data.
\newblock \emph{Bioinformatics}, 23\penalty0 (14):\penalty0 1768--1774, 2007.

\bibitem[Segal et~al.(2007)Segal, Sirlin, Ooi, Adler, Gollub, Chen, Chan,
  Matcuk, Barry, Chang, and Kuo]{SegalBeginRadio}
E.~Segal, C.~B. Sirlin, C.~Ooi, A.~S. Adler, J.~Gollub, X.~Chen, B.~K. Chan,
  G.~R. Matcuk, C.~T. Barry, H.~Y. Chang, and M.~D. Kuo.
\newblock Decoding global gene expression programs in liver cancer by
  noninvasive imaging.
\newblock \emph{Nature Biotechnology}, 25:\penalty0 675--680, 2007.

\bibitem[Sun et~al.(1996)Sun, Shook, and Kay]{Sun96UniMulti}
G.~W. Sun, T.~L. Shook, and G.~L. Kay.
\newblock Inappropriate use of bivariable analysis to screen risk factors for
  use in multivariable analysis.
\newblock \emph{Journal of Clinical Epidemiology}, 49:\penalty0 907--916, 1996.

\bibitem[Thomson(1939)]{Thomson}
G.~Thomson.
\newblock \emph{{The Factorial Analysis of Human Ability}}.
\newblock {London: University of Londen Press}, 1939.

\bibitem[Thurstone(1947)]{Thurstone47}
L.~L. Thurstone.
\newblock \emph{{Multiple Factor Analysis}}.
\newblock {Chicago: University of Chicago Press}, 1947.

\bibitem[Thurstone(1954)]{Thurstone54}
L.~L. Thurstone.
\newblock An analytical method for simple structure.
\newblock \emph{Psychometrika}, 19:\penalty0 173--182, 1954.

\bibitem[Tibshirani(1996)]{lasso}
R~Tibshirani.
\newblock Regression shrinkage and selection via the lasso.
\newblock \emph{Journal of the Royal Statistical Society, Series B},
  58:\penalty0 267--288, 1996.

\bibitem[Trendafilov et~al.(2017)Trendafilov, Fontanella, and
  Adachi]{Trend17SFA}
N.~T. Trendafilov, S.~Fontanella, and K.~Adachi.
\newblock Sparse exploratory factor analysis.
\newblock \emph{Psychometrika}, 82:\penalty0 778--794, 2017.

\bibitem[Tutz and Binder(2007)]{TB07}
G.~Tutz and H.~Binder.
\newblock Boosting ridge regression.
\newblock \emph{Computational Statistics and Data Analysis}, 51\penalty0
  (12):\penalty0 6044--6059, 2007.

\bibitem[Wang et~al.(2011)Wang, Nan, Rosset, and Zhu]{randomLasso}
S.~Wang, B.~Nan, S.~Rosset, and J.~Zhu.
\newblock Random lasso.
\newblock \emph{The Annals of Applied Statistics}, 5:\penalty0 468--485, 2011.

\bibitem[Warton(2008)]{Warton08}
D.~I. Warton.
\newblock Penalized normal likelihood and ridge regularization of correlation
  and covariance matrices.
\newblock \emph{Journal of the American Statistical Association}, 103:\penalty0
  340--349, 2008.

\bibitem[Yip and Aerts(2016)]{Yip16Limitations}
S.~S.~F. Yip and H.~J. W.~L. Aerts.
\newblock Applications and limitations of radiomics.
\newblock \emph{Physics in Medicine \& Biology}, 61:\penalty0 R150--R166, 2016.

\bibitem[Yuan and Chan(2008)]{Yuan08SEMNS}
{K.-H.} Yuan and W.~Chan.
\newblock Structural equation modeling with near singular covariance matrices.
\newblock \emph{Computational Statistics and Data Analysis}, 52:\penalty0
  4842--4858, 2008.

\bibitem[Zou and Hastie(2005)]{enet}
H.~Zou and T.~Hastie.
\newblock Regularization and variable selection via the elastic net.
\newblock \emph{Journal of the Royal Statistical Society, Series B},
  67:\penalty0 301--320, 2005.

\end{thebibliography}


\begin{thebibliography}{40}
\providecommand{\natexlab}[1]{#1}
\providecommand{\url}[1]{\texttt{#1}}
\expandafter\ifx\csname urlstyle\endcsname\relax
  \providecommand{\doi}[1]{doi: #1}\else
  \providecommand{\doi}{doi: \begingroup \urlstyle{rm}\Url}\fi

\bibitem[Akaike(1973)]{Aik73_SM}
H.~Akaike.
\newblock Information theory and an extension of the maximum likelihood
  principle.
\newblock In B.~N. Petrov and F.~Csaki, editors, \emph{Second International
  Symposium on Information Theory}, pages 267--281. Budapest: Akademiai Kaido,
  1973.

\bibitem[Amemiya and Anderson(1990)]{AA_LRT_SM}
Y.~Amemiya and T.~W. Anderson.
\newblock Asymptotic chi-square tests for a large class of factor analysis
  models.
\newblock \emph{The Annals of Statistics}, 18:\penalty0 1453--1463, 1990.

\bibitem[Anderson(2003, 3rd ed.)]{AndersonBible_SM}
T.~W. Anderson.
\newblock \emph{An Introduction to Multivariate Statistical Analysis}.
\newblock Hoboken, {NJ}: {J}ohn {W}iley \& {S}ons, {I}nc., 2003, 3rd ed.

\bibitem[Anderson and Rubin(1956)]{AndersonRubinClassic_SM}
T.~W. Anderson and H.~Rubin.
\newblock Statistical inference in factor analysis.
\newblock In \emph{Proceedings of the Third Berkeley Symposium on Mathematical
  Statistics and Probability}, volume {5: Contributions to Econometrics,
  Industrial Research, and Psychometry}, pages 111--150. {Berkeley, CA:
  University of California Press}, 1956.

\bibitem[Bartlett(1937)]{BartlettScores_SM}
M.~S. Bartlett.
\newblock The statistical conception of mental factors.
\newblock \emph{British Journal of Psychology}, 28:\penalty0 97--104, 1937.

\bibitem[Boellaard et~al.(2015)Boellaard, {Delgado-Bolton}, Oyen, Giammarile,
  Tatsch, Eschner, Verzijlbergen, Barrington, Pike, Weber, Stroobants, Delbeke,
  Donohoe, Holbrook, Graham, Testanera, Hoekstra, Zijlstra, Visser, Hoekstra,
  Pruim, Willemsen, Arends, Kotzerke, Bockisch, Beyer, Chiti, and
  Krause]{AccurateBoel_SM}
R.~Boellaard, R.~{Delgado-Bolton}, W.~J. Oyen, F.~Giammarile, K.~Tatsch,
  W.~Eschner, F.~J. Verzijlbergen, S.~F. Barrington, L.~C. Pike, W.~A. Weber,
  S.~Stroobants, D.~Delbeke, K.~J. Donohoe, S.~Holbrook, M.~M. Graham,
  G.~Testanera, O.~S. Hoekstra, J.~Zijlstra, E.~Visser, C.~J. Hoekstra,
  J.~Pruim, A.~Willemsen, B.~Arends, J.~Kotzerke, A.~Bockisch, T.~Beyer,
  A.~Chiti, and B.~J. Krause.
\newblock {FDG PET/CT: EANM} procedure guidelines for tumour imaging: version
  2.0.
\newblock \emph{European Journal of Nuclear Medicine and Molecular Imaging},
  42:\penalty0 328--354, 2015.

\bibitem[Brent(1971)]{Brent_SM}
R.~P. Brent.
\newblock An algorithm with guaranteed convergence for finding a zero of a
  function.
\newblock \emph{The Computer Journal}, 14:\penalty0 422--425, 1971.

\bibitem[Brier(1950)]{Brier_SM}
G.~W. Brier.
\newblock Verselection of forecasts expressed in terms of probability.
\newblock \emph{Monthly Weather Review}, 78:\penalty0 1--3, 1950.

\bibitem[D. and Horii(1992)]{DICOM_SM}
Bidgood~W. D. and S.~C. Horii.
\newblock Introduction to the {ACRNEMA DICOM} standard.
\newblock \emph{Radiographics}, 12:\penalty0 345--355, 1992.

\bibitem[Fabrigar et~al.(1999)Fabrigar, Wegener, {MacCallum}, and
  Strahan]{FWMS99_SM}
L.~Fabrigar, D.~T. Wegener, R.~C. {MacCallum}, and E.~J. Strahan.
\newblock Evaluating the use of exploratory factor analysis is psychological
  research.
\newblock \emph{Psychological Methods}, 4:\penalty0 272--299, 1999.

\bibitem[Fava and Velicer(1992)]{FAV92_SM}
J.~L. Fava and W.~F. Velicer.
\newblock The effects of overextraction on factor and component analysis.
\newblock \emph{Multivariate Behavioral Research}, 27:\penalty0 387--415, 1992.

\bibitem[Fletcher and Powell(1963)]{FPalgo63_SM}
R.~Fletcher and M.~J.~D. Powell.
\newblock A rapidly convergent descent method for minimization.
\newblock \emph{Computer Journal}, 2:\penalty0 163--168, 1963.

\bibitem[Gerds(2017)]{pec_SM}
T.~A. Gerds.
\newblock \emph{\texttt{pec}: {Prediction} error curves for risk prediction
  models in survival analysis}, 2017.
\newblock R package version 2.5.4.

\bibitem[Graf et~al.(1999)Graf, Schmoor, Sauerbrei, and
  Schumacher]{BrierModern_SM}
E.~Graf, C.~Schmoor, W.~Sauerbrei, and M.~Schumacher.
\newblock Assessment and comparison of prognostic classifcation schemes for
  survival data.
\newblock \emph{Statistics in Medicine}, 18:\penalty0 2529--2545, 1999.

\bibitem[Guttman(1955)]{Gutt55_SM}
L.~Guttman.
\newblock The determinacy of factor score matrices with implications for five
  other basic problems of common-factor theory.
\newblock \emph{The British Journal of Statistical Psychology}, 8:\penalty0
  65--81, 1955.

\bibitem[Guttman(1956)]{Guttman56_SM}
L.~Guttman.
\newblock Best possible systematic estimates of communalities.
\newblock \emph{Psychometrika}, 21:\penalty0 273--285, 1956.

\bibitem[Hayashi et~al.(2007)Hayashi, Bentler, and Yuan]{HBY07_SM}
K.~Hayashi, P.~M. Bentler, and {K.-H.} Yuan.
\newblock On the likelihood ratio test for the number of factors in exploratory
  factor analysis.
\newblock \emph{Structural Equation Modeling}, 14:\penalty0 505--526, 2007.

\bibitem[Horst(1965)]{Horst65_SM}
P.~Horst.
\newblock \emph{{Factor Analysis of Data Matrices}}.
\newblock {New York: Holt, Rinehart and Winston}, 1965.

\bibitem[J\"{o}reskog(1967)]{Joreskog67_SM}
K.~G. J\"{o}reskog.
\newblock Some contributions to maximum likelihood factor analysis.
\newblock \emph{Psychometrika}, 32:\penalty0 443--482, 1967.

\bibitem[Kaiser(1958)]{Kaiser58_SM}
H.~F. Kaiser.
\newblock The varimax criterion for analytic rotation in factor analysis.
\newblock \emph{Psychometrika}, 23:\penalty0 187--200, 1958.

\bibitem[Kaiser(1970)]{Kaiser70_SM}
H.~F. Kaiser.
\newblock A second-generation little jiffy.
\newblock \emph{Psychometrika}, 35:\penalty0 401--415, 1970.

\bibitem[Kaiser and Rice(1974)]{KaiserRice74_SM}
H.~F. Kaiser and J.~Rice.
\newblock Little jiffy, mark {IV}.
\newblock \emph{Educational and Pscyhological Measurement}, 34:\penalty0
  111--117, 1974.

\bibitem[Kollo and Ruul(2003)]{KR03_SM}
T.~Kollo and K.~Ruul.
\newblock Approximations to the distribution of the sample correlation matrix.
\newblock \emph{Journal of Multivariate Analysis}, 85:\penalty0 318--334, 2003.

\bibitem[Lopes and West(2004)]{LW04_SM}
H.~Lopes and M.~West.
\newblock Bayesian model assessment in factor analysis.
\newblock \emph{Statistica Sinica}, 14:\penalty0 41--67, 2004.

\bibitem[Martens et~al.(2019)Martens, Koopman, Noij, Pfaehler, \"{U}belh\"{o}r,
  Sharma, Vergeer, Leemans, Hoekstra, Maqsood, Zwezerijnen, Heymans, Peeters,
  {de Bree}, {de Graaf}, Castelijns, and Boellaard]{Martens19_SM}
R.~M. Martens, T.~Koopman, D.~P. Noij, E.~Pfaehler, C.~\"{U}belh\"{o}r,
  S.~Sharma, M.~R. Vergeer, C.~R. Leemans, O.~S. Hoekstra, Y.~Maqsood,
  B.~Zwezerijnen, M.~W. Heymans, C.~F.~W. Peeters, R.~{de Bree}, P.~{de Graaf},
  J.~A. Castelijns, and R.~Boellaard.
\newblock Predictive value of quantitative $^{18}${F-FDG-PET} radiomics in
  patients with head and neck squamous cell carcinoma.
\newblock Technical report, Amsterdam University Medical Centers, 2019.

\bibitem[Mes et~al.(2019)Mes, {van Velden}, Peltenburg, Peeters, {te Beest},
  {van de Wiel}, Mekke, Mulder, Martens, Castelijns, Pameijer, {de Bree},
  Boellaard, Leemans, Brakenhoff, and {de Graaf}]{Mes19_SM}
S.~W. Mes, F.~H.~P. {van Velden}, B.~Peltenburg, C.~F.~W. Peeters, D.~{te
  Beest}, M.~A. {van de Wiel}, J.~Mekke, D.~C. Mulder, R.~M. Martens, J.~A.
  Castelijns, F.~A. Pameijer, R.~{de Bree}, R.~Boellaard, C.~R. Leemans, R.~H.
  Brakenhoff, and P.~{de Graaf}.
\newblock Head and neck cancer outcome prediction by {MRI} radiomic signatures.
\newblock Technical report, Amsterdam University Medical Centers, 2019.

\bibitem[Mogensen et~al.(2012)Mogensen, Ishawaran, and Gerds]{PEcurves_SM}
U.~B. Mogensen, H.~Ishawaran, and T.~A. Gerds.
\newblock Evaluating random forests for survival analysis using prediction
  error curves.
\newblock \emph{Journal of Statistical Software}, 50:\penalty0 11, 2012.

\bibitem[Mulaik(2010)]{Mulaik2010_SM}
S.~A. Mulaik.
\newblock \emph{{Foundations of Factor Analysis}}.
\newblock {Boca Raton: Chapman \& Hall/CRC}, 2nd edition, 2010.

\bibitem[Peeters(2012)]{PeetersThesis_SM}
C.~F.~W. Peeters.
\newblock \emph{{Bayesian Exploratory and Confirmatory Factor Analysis:
  Perspectives on Constrained-Model Selection}}.
\newblock PhD thesis, {Dept.\ of Methodology \& Statistics, Utrecht
  University}, 2012.

\bibitem[Peeters et~al.(2014)Peeters, Dziura, and {van Wesel}]{PeetersAEfa_SM}
C.~F.~W. Peeters, J.~Dziura, and F.~{van Wesel}.
\newblock Pathophysiological domains underlying the metabolic syndrome: an
  alternative factor analytic strategy.
\newblock \emph{Annals of Epidemiology}, 24:\penalty0 762--770, 2014.

\bibitem[Peeters et~al.(2016)Peeters, {van de Wiel}, and {van
  Wieringen}]{PeetersCNplot_SM}
C.~F.~W. Peeters, M.~A. {van de Wiel}, and W.~N. {van Wieringen}.
\newblock The spectral condition number plot for regularization parameter
  determination.
\newblock arXiv:1608.04123 [stat.CO], 2016.

\bibitem[Preacher et~al.(2013)Preacher, Zhang, Kim, and Mels]{PreachChoose_SM}
K.~J. Preacher, G.~Zhang, C.~Kim, and G.~Mels.
\newblock Choosing the optimal number of factors in exploratory factor
  analysis: {A} model selection perspective.
\newblock \emph{Multivariate Behavioral Research}, 48:\penalty0 28--56, 2013.

\bibitem[Reyment and J\"{o}reskog(1996)]{RJbookFANA}
R.~Reyment and K.~G. J\"{o}reskog.
\newblock \emph{Applied factor analysis in the natural sciences}.
\newblock Cambridge: Cambridge University Press, 1996.

\bibitem[Schwarz(1978)]{BIC_SM}
G.~E. Schwarz.
\newblock Estimating the dimension of a model.
\newblock \emph{Annals of Statistics}, 6:\penalty0 461--464, 1978.

\bibitem[Thomson(1939)]{Thomson_SM}
G.~Thomson.
\newblock \emph{{The Factorial Analysis of Human Ability}}.
\newblock {London: University of Londen Press}, 1939.

\bibitem[{van Velden} et~al.(2016){van Velden}, Kramer, Frings, Nissen, Mulder,
  {de Langen}, Hoekstra, Smit, and Boellaard]{MRIextract_SM}
F.~H.~P. {van Velden}, G.~M. Kramer, V.~Frings, I.~A. Nissen, E.~R. Mulder,
  A.~J. {de Langen}, O.~S. Hoekstra, E.~F. Smit, and R.~Boellaard.
\newblock Repeatability of radiomic features in non-small-cell lung cancer
  {[$^{18}$F]FDG-PET/CT} studies: {Impact} of reconstruction and delineation.
\newblock \emph{Molecular Imaging and Biology}, 18:\penalty0 788--795, 2016.

\bibitem[Warton(2008)]{Warton08_SM}
D.~I. Warton.
\newblock Penalized normal likelihood and ridge regularization of correlation
  and covariance matrices.
\newblock \emph{Journal of the American Statistical Association}, 103:\penalty0
  340--349, 2008.

\bibitem[Wood et~al.(1996)Wood, Tataryn, and Gorsuch]{WTG96_SM}
J.~M. Wood, D.~J. Tataryn, and R.~L. Gorsuch.
\newblock Effects of under- and overextraction on principal axis factor
  analysis with varimax rotation.
\newblock \emph{Psychological Methods}, 1:\penalty0 354--365, 1996.

\bibitem[Woodbury(1950)]{Woodbury_SM}
M.~A. Woodbury.
\newblock Inverting modified matrices.
\newblock {Statistical Research Group Memorandum Report}~42, {Princeton
  University}, 1950.

\bibitem[Yuan and Chan(2008)]{Yuan08SEMNS_SM}
{K.-H.} Yuan and W.~Chan.
\newblock Structural equation modeling with near singular covariance matrices.
\newblock \emph{Computational Statistics and Data Analysis}, 52:\penalty0
  4842--4858, 2008.

\end{thebibliography}
\putbib[Radiomics_SM]
\end{bibunit}

\end{document}